\PassOptionsToPackage{fleqn}{amsmath}
\documentclass[twoside]{new-aiaa}
\usepackage{style}
\title{Neural-Rendezvous: Provably Robust Guidance and Control\\to Encounter Interstellar Objects\thanks{YouTube video: {\color{caltechgreen}{\href{https://youtu.be/O4cAyemZgdA}{https://youtu.be/O4cAyemZgdA}}}. 
Part of the project was carried out at the Jet Propulsion Laboratory, California Institute of Technology, under a contract with the National Aeronautics and Space Administration.}}
\author{Hiroyasu Tsukamoto\thanks{Ph.D. Student, Student Member AIAA (at the date of submission); Affiliate, Maritime and Multi-Agent Autonomy Group, Jet Propulsion Laboratory (JPL) (2023 -- ); Assistant Professor, Department of Aerospace Engineering, University of Illinois at Urbana-Champaign (2024 -- ), Young Professional AIAA; \href{mailto:hiroyasu.tsukamoto@jpl.nasa.gov}{hiroyasu.tsukamoto@jpl.nasa.gov} (corresponding author).} and Soon-Jo Chung\thanks{Bren Professor of Control and Dynamical Systems and JPL Senior Research Scientist, Associate Fellow AIAA; \href{mailto:sjchung@caltech.edu}{sjchung@caltech.edu}.}}
\affil{Division of Engineering and Applied Science, California Institute of Technology, Pasadena, California 91125}
\author{Yashwanth Kumar Nakka\thanks{Robotics Technologist, Maritime and Multi-Agent Autonomy Group, Member AIAA; \href{mailto:yashwanth.kumar.nakka@jpl.nasa.gov}{yashwanth.kumar.nakka@jpl.nasa.gov}.}, Benjamin Donitz\thanks{Systems Engineer, Advanced Systems Design Engineering Group, Member AIAA; \href{mailto:benjamin.p.donitz@jpl.nasa.gov}{benjamin.p.donitz@jpl.nasa.gov}.}, Declan Mages\thanks{Navigation Engineer, Outer Planet Navigation Group; \href{mailto:declan.m.mages@jpl.nasa.gov}{declan.m.mages@jpl.nasa.gov}.}, and Michel Ingham\thanks{Chief Technologist, Flight Software Systems Engineering and Architectures Group, Associate Fellow AIAA; \href{mailto:michel.d.ingham@jpl.nasa.gov}{michel.d.ingham@jpl.nasa.gov}.}}
\affil{Jet Propulsion Laboratory, California Institute of Technology, Pasadena, California 91109}

\begin{document}
\twocolumn[{\maketitle\thispagestyle{titlepagefancy}\begin{adjustwidth}{0.3in}{0.3in}\vspace{0.1in}\begin{abstract}
Interstellar objects (ISOs) are likely representatives of primitive materials invaluable in understanding exoplanetary star systems. Due to their poorly constrained orbits with generally high inclinations and relative velocities, however, exploring ISOs with conventional human-in-the-loop approaches is significantly challenging. This paper presents Neural-Rendezvous -- a deep learning-based guidance and control framework for encountering fast-moving objects, including ISOs, robustly, accurately, and autonomously in real time. It uses pointwise minimum norm tracking control on top of a guidance policy modeled by a spectrally-normalized deep neural network, where its hyperparameters are tuned with a loss function directly penalizing the MPC state trajectory tracking error. We show that Neural-Rendezvous provides a high probability exponential bound on the expected spacecraft delivery error, the proof of which leverages stochastic incremental stability analysis. In particular, it is used to construct a non-negative function with a supermartingale property, explicitly accounting for the ISO state uncertainty and the local nature of nonlinear state estimation guarantees. In numerical simulations, Neural-Rendezvous is demonstrated to satisfy the expected error bound for 100 ISO candidates. This performance is also empirically validated using our spacecraft simulator and in high-conflict and distributed UAV swarm reconfiguration with up to 20 UAVs.
\end{abstract}
\end{adjustwidth}\vspace{0.2in}}]
\pagestyle{myfancy}
\saythanks
\section*{Nomenclature}
{\renewcommand\arraystretch{1.0}
\begin{supertabular}{@{}p{1.171in} @{\quad=\quad} p{2in}@{}}
$\mathscr{A}$ & weak infinitesimal operator of  stochastic processes \\
$\mathop{\mathbb{E}}$ & expected value operator \\
$\mathop{\mathbb{E}}_Z$ & conditional expected value operator given $Z$ \\
$\mathrm{I}_{m\times n}$ & $m\times n$ identity matrix \\
$m_{\mathrm{sc}}(t)$ & mass of spacecraft at time $t$ \\
$\mathrm{O}_{m\times n}$ & $m\times n$ zero matrix \\
$\textit{\oe}(t),~\hat{\textit{\oe}}(t),~\textit{\oe}_d(t)$ & true, estimated, and desired orbital elements of ISO at time $t$ \\
$\mathop{\mathbb{P}}$ & probability measure \\
$p(t),~\hat{p}(t),~p_d(t)$ & true, estimated, and desired position of spacecraft relative to ISO in LVLH frame at time $t$ \\
\end{supertabular}
\begin{supertabular}{@{}p{1.171in} @{\quad=\quad} p{2in}@{}}
$x(t),~\hat{x}(t),~x_d(t)$ & true, estimated, and desired state of spacecraft relative to ISO in LVLH frame at time $t$ \\
$u$ & control policy \\
$y(t)$ & state measurement at time $t$ \\
$\rho$ & desired terminal position of spacecraft relative to ISO \\
\end{supertabular}
}
\section{Introduction}
\label{sec_introduction}
\begin{figure}[htbp]
    \centering
    \href{https://youtu.be/O4cAyemZgdA}{\includegraphics[width=83mm]{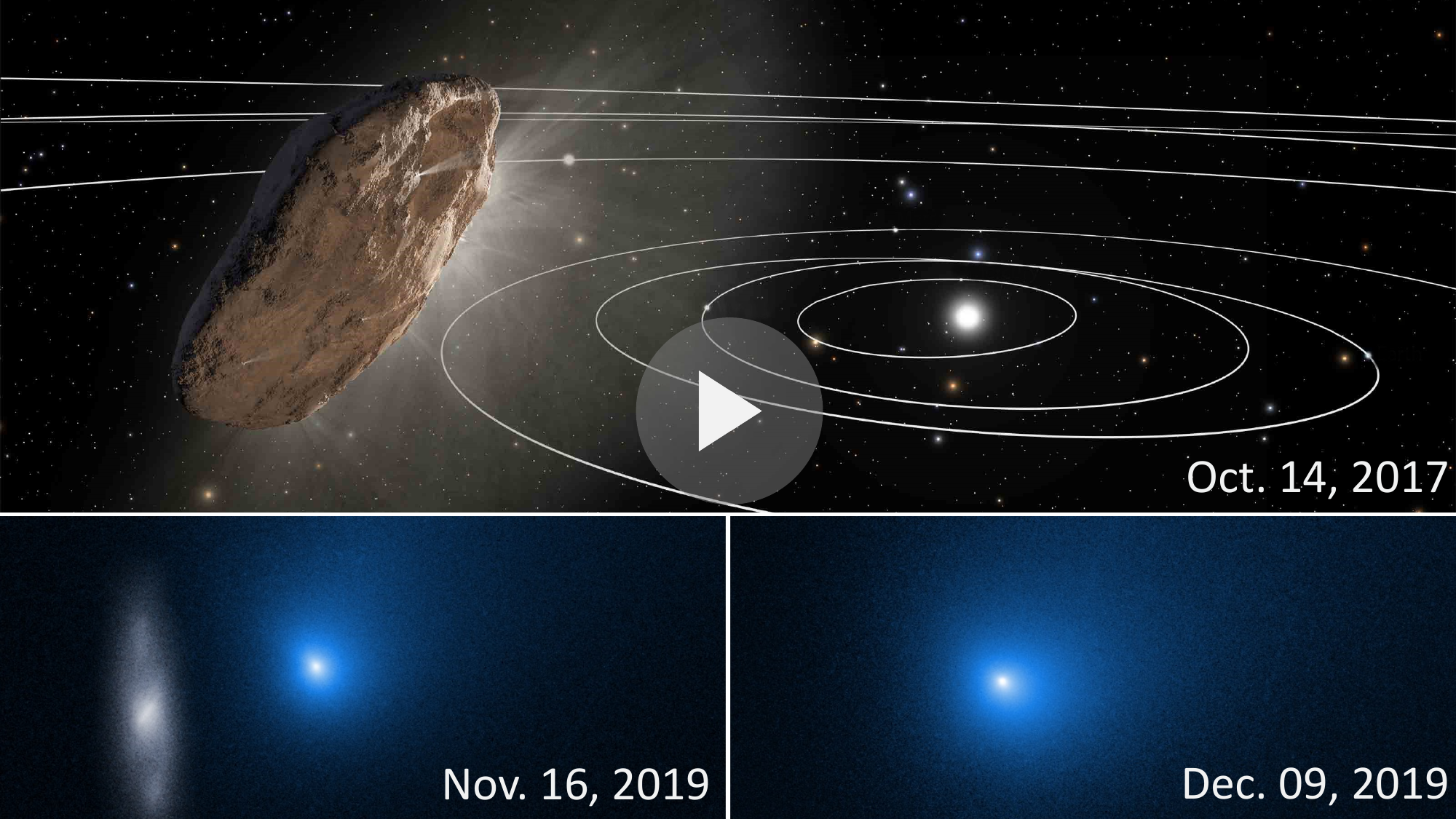}}
    \caption{Above: Artist's illustration 1I/`Oumuamua (\href{https://photojournal.jpl.nasa.gov/catalog/PIA22357}{NASA, ESA, and STScI}). Below: 2I/Borisov near and at perihelion (\href{https://apod.nasa.gov/apod/ap220305.html}{NASA, ESA, and D. Jewitt}).}
    \label{fig_oumuamua}
\end{figure}
\begin{figure}[htbp]
    \centering
    \includegraphics[width=83mm]{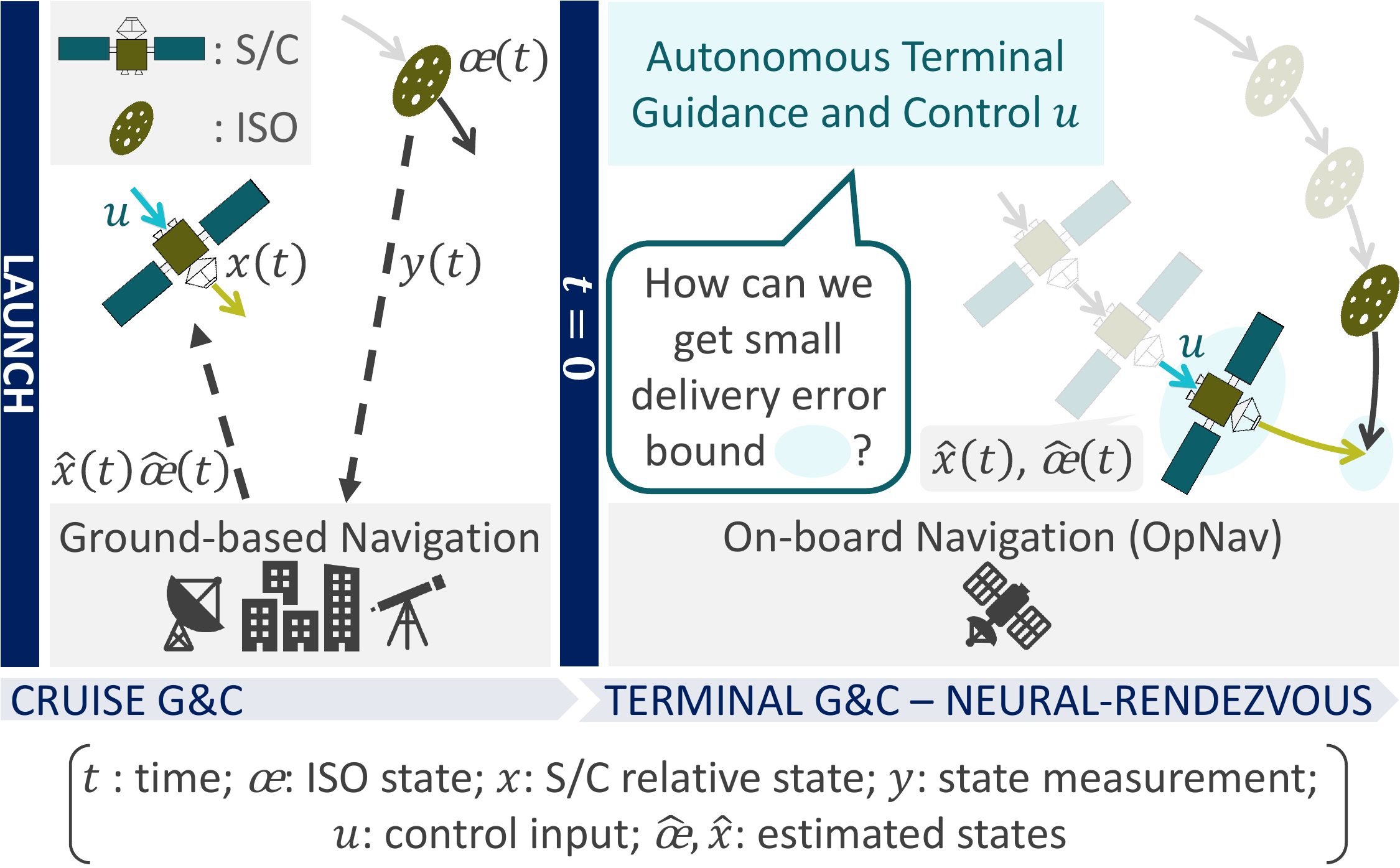}
    \caption{Illustration of cruise and terminal GNC. Neural-Rendezvous provides a verifiable delivery error bound under the large ISO state uncertainty and high-velocity challenges.}
    \label{fig_nr_outline}
\end{figure}

\lettrine{I}{nterstellar} objects (ISOs) represent one of the last unexplored classes of solar system objects. They are active or inert objects passing through our solar system on an unbounded hyperbolic trajectory about the Sun, which could sample planetesimals and primitive materials that provide vectors to compare our solar system with neighboring exoplanetary star systems. To date, two such objects have been identified and observed: 1I/`Oumuamua~\cite{oumuamua} discovered in 2017 and 2I/Borisov~\cite{guzick-2020} discovered in 2019 (Fig.~\ref{fig_oumuamua}). In 2022, the US Department of Defense confirmed that a third ISO impacted Earth in 2014. ISOs are physical laboratories that can enable the study of exosolar systems in situ rather than remotely using telescopes such as the Hubble or James Webb Space Telescopes. Using a dedicated spacecraft to flyby an ISO opens the doors to high-resolution imaging, mass or dust spectroscopy, and a larger number of vantage points than Earth observation. It could also resolve the target's nucleus shape and spin, characterize the volatiles being shed from an active body, reveal fresh surface material using an impactor, and more~\cite{castillo-wp}.
The discovery and exploration of ISOs are challenging for three main reasons: (i) they are not discovered until they are close to Earth, meaning that launches to encounter them often require high launch energy; (ii) their orbital properties are poorly constrained at launch, generally leading to significant on-board resources to encounter; and (iii) the encounter speeds are typically high ($>10$s~of~\si{\kilo\metre\per\second}) requiring fast response autonomous operations. As outlined in Fig.~\ref{fig_nr_outline}, the guidance, navigation, and control (GNC) of spacecraft for the ISO encounter are split into two segments: (i) the cruise phase, where the spacecraft utilizes state estimation obtained by ground-based telescopes and navigates via ground-in-the-loop operations, and; (ii) the terminal phase, where it switches to fully autonomous operation with existing onboard navigation frameworks. Our proposed approach, Neural-Rendezvous, is for performing the second phase of the autonomous terminal guidance and control (G\&C) with the on-board state estimates and is built upon the spectrally-normalized deep neural network (SN-DNN)~\cite{miyato2018spectral}. The overall mission outline and motivation are visually summarized at {\color{caltechgreen}\href{https://youtu.be/AhDPE-R5GZ4}{https://youtu.be/AhDPE-R5GZ4}} (see Fig.~\ref{fig_oumuamua}).
\subsection*{Outline of Neural-Rendezvous}
In this paper, we utilize a dynamical system-based SN-DNN for designing a real-time guidance policy that approximates nonlinear model predictive control (MPC), the linear and discretized version of which is known to be near-optimal in terms of dynamic regret (\ie{}, the MPC performance minus the optimal performance in hindsight~\cite{NEURIPS2020_155fa095}). This is to avoid solving nonlinear MPC optimization at each time instant and compute a spacecraft's control input autonomously, even with its limited online computational capacity. Consistent with our objective of encountering ISOs, our SN-DNN uses a loss function that directly imitates the MPC state trajectory performing dynamics integration~\cite{neuralode,meta_trj_loss}, as well as indirectly imitating the MPC control input as in existing methods~\cite{learningmpc}. We then introduce learning-based min-norm feedback control to be used on top of this guidance policy. This provides an optimal and robust control input that minimizes its instantaneous deviation from that of the SN-DNN guidance policy, under the incremental stability condition as in the one of contraction theory~\cite{Ref:contraction1}. Our contributions here are summarized as follows.
\subsection*{Contribution}
If the SN-DNN guidance policy is equipped with the learning-based min-norm control, the state tracking error bound with respect to the desired state trajectory decreases exponentially in expectation with a finite probability, robustly against the state uncertainty. This indicates that the terminal spacecraft deliver error at the ISO encounter (\ie{}, the shaded blue region in Fig.~\ref{fig_nr_outline}) is probabilistically bounded in expectation, where its size can be modified accordingly to the mission requirement by tuning its feedback control parameters. Our main contribution is to provide a rigorous proof of this fact by leveraging stochastic incremental stability analysis. It is based on constructing a non-negative function with a supermartingale property for finite-time tracking performance, explicitly accounting for the ISO state uncertainty and the local nature of nonlinear state estimation guarantees. Also, the pointwise min-norm controller design is addressed and solved in analytical form for Lagrangian dynamical systems~\cite[p. 392]{Ref_Slotine}, which describe a variety of general robotic motions, not just the one used in our paper for ISO exploration. We further show that the SN-DNN guidance policy possesses a verifiable optimality gap with respect to the optimization-based G\&C, under the assumption that the nonlinear MPC policy is Lipschitz. 

Although there are numerous studies that use neural networks for learning-based control, the strength of our proposed technique lies in formally deriving their mathematical guarantees, with the assumptions behind them clearly spelled out. It is natural that these guarantees could be conservative when satisfying the assumptions is challenging. Even in these situations, our approach is at least mathematically interpretable without treating the learning part like a black box, thereby providing its users with a clear understanding of how each part of the design leads to its superior/inferior performance in the real world.

\begin{figure*}[htbp]
\centering
\begin{subfigure}[b]{0.33\textwidth}
\centering
\href{https://youtu.be/AhDPE-R5GZ4?t=210}{\includegraphics[width=\textwidth]{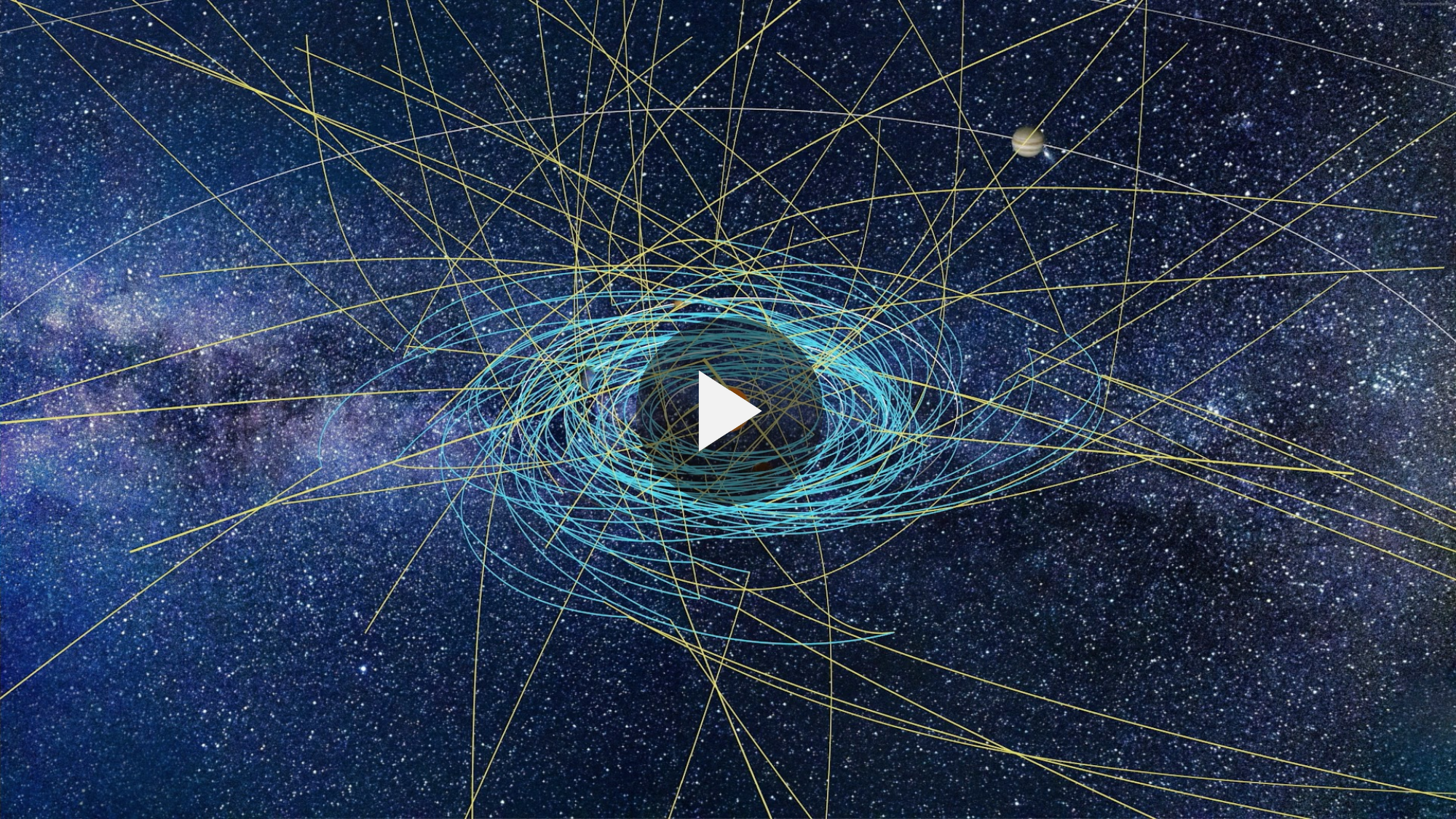}}
\caption{Numerical simulation}
\label{fig_isoplay}
\end{subfigure}
\hfill
\begin{subfigure}[b]{0.33\textwidth}
\centering
\href{https://youtu.be/D0HxsNh_rHM}{\includegraphics[width=\textwidth]{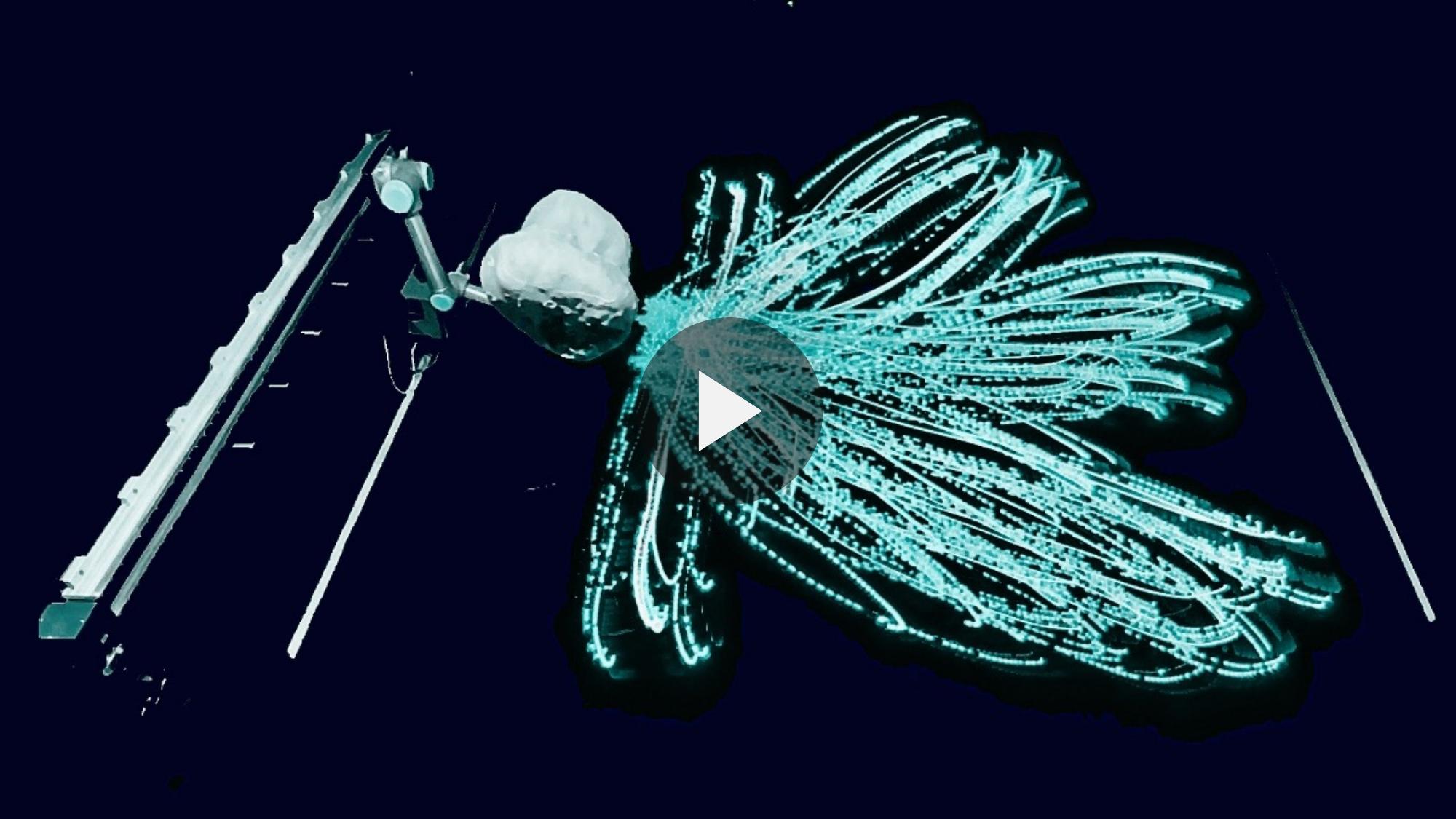}}
\caption{Spacecraft simulator rendezvous}
\label{fig_scplay}
\end{subfigure}
\hfill
\begin{subfigure}[b]{0.33\textwidth}
\centering
\href{https://youtu.be/u8ASO_r8rEA}{\includegraphics[width=\textwidth]{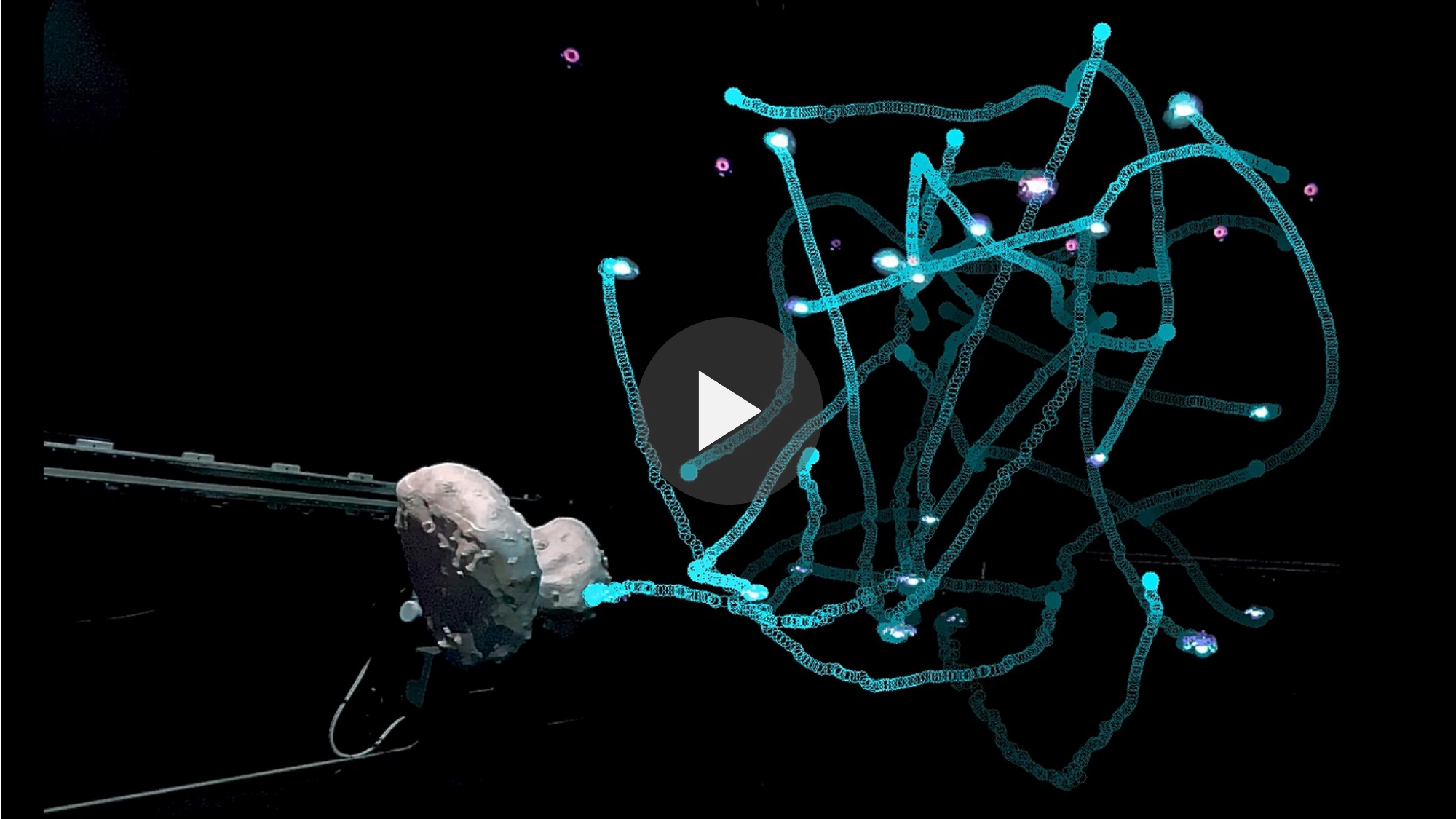}}
\caption{UAV swarm reconfiguration}
\label{fig_uavplay}
\end{subfigure}
\caption{Numerical and experimental validation of Neural-Rendezvous to be discussed in Sec.~\ref{sec_simulation} and Sec.~\ref{sec_experiment}.}
\label{fig_allplays}
\end{figure*}

In numerical simulations with $100$ ISO candidates for possible exploration obtained as in~\cite{our_iso_mission}, Neural-Rendezvous outperforms other G\&C algorithms including (i) the SN-DNN guidance policy; (ii) proportional-derivative (PD) control with a pre-computed desired trajectory; (iii) robust nonlinear tracking control of~\cite[pp. 397-402]{Ref_Slotine} with a precomputed desired trajectory; and (iv) MPC with linearized dynamics~\cite{mpc1}, in terms of the terminal spacecraft delivery error. In our simulation setup, Neural-Rendezvous indeed satisfies the probabilistic state tracking error bound for all the ISO candidates. It achieves a delivery error less than \SI{0.2}{\kilo\metre} for \SI{99}{\percent} of the candidates, with $8.0\times10^{-4}$~\si{\second} computational time for its one-step evaluation and with control effort less than \SI{0.6}{\kilo\metre\per\second}. It is also demonstrated that the Neural-Rendezvous's delivery error and control effort are only $5.84\times10^{-3}$\% and $5.48$\% larger than that of the robust nonlinear MPC, respectively. Note that the robust nonlinear MPC is the optimization-based G\&C scheme our methods attempt to imitate. A YouTube video that visualizes these simulation results as in (a) of~Fig.~\ref{fig_allplays} can be found at {\color{caltechgreen}\href{https://youtu.be/AhDPE-R5GZ4?t=210}{https://youtu.be/AhDPE-R5GZ4?t=210}}.

It is worth noting that Neural-Rendezvous and its guarantees are general enough to be used not only for encountering ISOs, but also for solving various nonlinear autonomous rendezvous problems accurately in real time under external disturbances and various sources of uncertainty (\eg{}, state measurement noise, process noise, control execution error, unknown parts of dynamics, or parametric/non-parametric variations of dynamics and environments). To further corroborate these arguments, the performance of Neural-Rendezvous is also empirically validated using our thruster-based spacecraft simulator called M-STAR~\cite{SCsimulator} and in high-conflict and distributed UAV swarm reconfiguration with up to 20 UAVs (see (b) and (c) of~Fig.~\ref{fig_allplays}).
\subsubsection*{Contribution in the context of learning-based control}
Let us make a few important remarks about our approach, which is based on \textit{offline} imitation learning of nonlinear optimization. Firstly, our stability and robustness results are independent of the methods used for learning as implied earlier. This means that the proofs also work for learned motion planning policies with indirect methods as in adaptive dynamic programming~\cite{ADP_review} (see~\cite{ddp_constraint} for its potential in including nonlinear constraints). Secondly, the role of our offline learning here is to replace the heavy online computation, especially in solving nonlinear optimization for motion planning, with a single neural network evaluation. It provides an approximate online solution even for general complex and large-scale nonlinear motion planning and control problems (\eg{}, distributed and intelligent motion planning and control of high-conflict UAV swarm reconfiguration, the problem of which is highly nonlinear and thus solving even a single optimization online could become unrealistic). We have also performed experimental validation on these scenarios to corroborate this argument. Thirdly, online learning and adaptive control methods~\cite{ADP_review,doi:10.1126/scirobotics.abm6597} can be used on top of our work, exploiting real-time information for actively dealing with uncertainty in our dynamics. In our paper, the estimation scheme for the online state uncertainty quantification is assumed to be given externally by the AutoNav system~\cite{autonav}.


\subsection*{Remarks on Related Work}
The state of practice in realizing asteroid and comet rendezvous missions is to pre-construct an accurate spacecraft state trajectory to the target before launch, and then perform a few trajectory correction maneuvers (TCMs) along the way based on the state measurements obtained by ground-based and onboard autonomous navigation schemes as discussed in~\cite{our_iso_mission,declan_nav}. Such a G\&C approach is only feasible for targets with sufficient information on their orbital properties in advance, which is not realistic for ISOs visiting our solar system from interstellar space with large state uncertainty.

The ISO rendezvous problem can be cast as a robust motion planning and control problem that has been investigated in numerous studies in the field of robotics. The most well-developed and commercialized of these G\&C methods is the robust nonlinear MPC~\cite{10.1007/BFb0109870}, which extensively utilizes knowledge about the underlying nonlinear dynamical system to design an optimal control input at each time instant, thereby allowing a spacecraft to use the most updated ISO state information. If the MPC is augmented with feedback control, robustness against the state uncertainty and various external disturbances can be shown using, \eg{}, Lyapunov and contraction theory~(see, \eg{},~\cite{Ref:contraction1,tutorial} and references therein). However, as mentioned earlier, the spacecraft's onboard computational power is not necessarily sufficient to solve the MPC optimization at each time instant of its TCM, which could lead to failure in accounting for the ISO state that changes dramatically in a few seconds due to the body's high velocity. Even when we solve the MPC in a discrete-time manner, the computational load of solving it online is not negligible, especially if the computational resource of the agent is limited and not sufficient to perform nonlinear optimization in real time.

Learning-based control designs have been considered a promising solution to this problem, as they allow replacing these optimization-based G\&C algorithms with computationally-cheap mathematical models, \eg{}, neural networks~\cite{learningmpc}. Neural-Rendezvous can be viewed as a novel variant of a learning-based control design with formal optimality, stability, and robustness guarantees, which are obtained by extending the results of~\cite{tutorial} for general nonlinear systems to the case of the ISO rendezvous problem and more. 
\subsection*{Notation}
\label{notation}
We let $\|x\|$ denote the Euclidean norm for a vector $x \in \mathbb{R}^n$, and let $A \succ 0$, $A \succeq 0$, $A \prec 0$, and $A \preceq 0$ denote the symmetric positive definite, positive semi-definite, negative definite, negative semi-definite matrices, respectively, for a square matrix $A \in \mathbb{R}^{n \times n}$. Also, the target celestial bodies will be considered to be ISOs in the subsequent sections for the sake of consistency, but our proposed method should work for long-period comets (LPCs), UAVs, and other fast-moving objects in robotics and aerospace applications, with the same arguments to be presented. In this paper, the main result will be stated in Theorem~\ref{Thm:control_robustness}, and the key concepts for understanding our contributions will be illustrated in Fig.~\ref{fig_node_analogy}~--~\ref{fig_nr_diagram}.
\section{Technical Challenges in ISO Exploration}
\label{sec_problem}
In this paper, we consider the following translational dynamical system of a spacecraft relative to an Interstellar Object (ISO), equipped with feedback control $u:\mathbb{R}^n\times\mathbb{R}^n\times\mathbb{R}_{\geq 0}\rightarrow\mathbb{R}^m$:
\begin{align}
    \label{original_dyn}
    \dot{x}(t) = f(x(t),\textit{\oe}(t),t)+B(x(t),\textit{\oe}(t),t)u(\hat{x}(t),\hat{\textit{\oe}}(t),t)
\end{align}
where $t\in\mathbb{R}_{\geq 0}$ is time, $\textit{\oe}:\mathbb{R}_{\geq 0}\rightarrow\mathbb{R}^n$ are the ISO orbital elements evolving by a separate equation of motion (see~\cite{doi:10.2514/1.55705} for details), $x:\mathbb{R}_{\geq 0}\rightarrow\mathbb{R}^n$ is the state of the spacecraft relative to $\textit{\oe}$ in a local-vertical local-horizontal (LVLH) frame centered on the ISO~\cite[pp. 710-712]{lvlhbook}, ${f}:
\mathbb{R}^n\times\mathbb{R}^n\times\mathbb{R}_{\geq 0}\rightarrow\mathbb{R}^n$ and $B:\mathbb{R}^n\times\mathbb{R}^n\times\mathbb{R}_{\geq 0}\rightarrow\mathbb{R}^{n\times m}$ are known smooth functions~\cite{doi:10.2514/1.37261} (see~\eqref{actualfnB}), and $\hat{\textit{\oe}}:\mathbb{R}_{\geq 0}\rightarrow\mathbb{R}^n$ and $\hat{x}:\mathbb{R}_{\geq 0}\rightarrow\mathbb{R}^n$ are the estimated ISO and spacecraft relative state in the LVLH frame given by an on-board navigation scheme, respectively. Particularly when we select $x$ as $x = [p^{\top},\dot{p}^{\top}]^{\top}$ as its state, where $p \in \mathbb{R}^3$ is the position of the spacecraft relative to the ISO, we have that
\begin{align}
    f &= \left[\begin{smallmatrix}\dot{p}\\-m_{\mathrm{sc}}(t)^{-1
    }\left(C(\textit{\oe})\dot{p}+G(p,\textit{\oe})\right)\end{smallmatrix}\right],~B = \left[\begin{smallmatrix}\mathrm{O}_{3\times 3}\\m_{\mathrm{sc}}(t)^{-1
    }\mathrm{I}_{3\times 3}\end{smallmatrix}\right]
    \label{actualfnB}
\end{align}
where $m_{\mathrm{sc}}(t)$ is the mass of the spacecraft described by the Tsiolkovsky rocket equation, the functions $G$ and $C$ are as given in~\cite{doi:10.2514/1.37261,doi:10.2514/1.55705}, and the arguments of $f$ and $B$ are omitted.
\begin{remark}
\label{remark_nav}
In general, the estimation errors $\|\hat{x}(t)-x(t)\|$ and $\|\hat{\textit{\oe}}(t)-\textit{\oe}(t)\|$ are expected to decrease locally in its state space with the help of the state-of-the-art onboard navigation schemes as the spacecraft gets closer to the ISO, utilizing more accurate ISO state measurements obtained from an onboard sensor as detailed in~\cite{declan_nav}. For example, if the extended Kalman filter or contraction theory-based estimator (see~\cite{6849943,Ref:contraction1,tutorial} and references therein) is used for navigation, their expected values can be shown to be locally bounded and exponentially decreasing in $t$. Although G\&C are the focus of our study and thus developing such a navigation technique is beyond our scope, those interested in this field can also refer to, \eg{},~\cite{our_iso_mission,declan_nav}, to tighten the estimation error bound using practical knowledge specific to the ISO dynamics. Online learning and adaptive control methods~\cite{ADP_review,doi:10.1126/scirobotics.abm6597} can also be used to actively deal with such uncertainty with some formal guarantees.
\end{remark}
Since the ISO state and its onboard estimate in~\eqref{original_dyn} change dramatically in time due to their poorly constrained orbits with high inclinations and high relative velocities, using a fixed desired trajectory computed at some point earlier in time could fail to utilize the radically changing ISO state information as much as possible. Even with the use of MPC with receding horizons, the online computational load of solving it is not negligible, especially if the computational resource of the agent is limited and not sufficient to perform nonlinear optimization in real time. Also, when we consider more challenging and highly nonlinear G\&C scenarios (see Sec.~\ref{sec_experiment}), solving even a single optimization online could become unrealistic. We, therefore, construct a guidance policy to optimally achieve the smallest spacecraft delivery error for given estimated states $\hat{x}(t)$ and $\hat{\textit{\oe}}(t)$ in~\eqref{original_dyn} at time $t$, and then design a learning-based guidance algorithm that approximates it with a verifiable optimality gap, so the spacecraft can update its desired trajectory autonomously in real time using the most recent state estimates $\hat{x}(t)$ and $\hat{\textit{\oe}}(t)$, which become more accurate as the spacecraft gets closer to the ISO as discussed in Remark~\ref{remark_nav}. Note that the size of the neural network for offline learning is selected to be small enough to significantly reduce the online computational load required in solving optimization.

The major design challenge lies in providing the learning approach with a formal guarantee for successful ISO encounters, \ie{}, achieving a sufficiently small spacecraft delivery error with respect to a given desired relative position to flyby or impact the ISO, which leads to the problem formulated as follows.
\paragraph{Autonomous terminal G\&C for ISO encounter with formal guarantees}
\label{Problem1}\mbox{}\\
We aim to design an autonomous nonlinear G\&C algorithm that (i) robustly tracks the desired trajectory with zero spacecraft delivery error, computed by utilizing the above learning-based guidance, and (ii) guarantees a finite tracking error bound even under the presence of the state uncertainty and the learning error.

\section{Autonomous Guidance via Dynamical System-Based Deep Learning}
\label{sec_guidance}
Before proceeding to solve~\ref{Problem1} in Sec.~\ref{sec_problem}, this section describes dynamical system-based deep learning to design the autonomous terminal guidance algorithm to be used in the subsequent sections. It utilizes the known dynamics of~\eqref{original_dyn} and~\eqref{actualfnB} in approximating an optimization-based guidance policy, thereby directly minimizing the deviation of the learned state trajectory from the optimal state trajectory with the smallest spacecraft delivery error at the ISO encounter, possessing a verifiable optimality gap with respect to the optimal guidance policy.
\subsection{Model Predictive Control Problem}
\label{sec_mpc_formulation}
Let us first introduce the following definition of an ISO state flow, which maps the ISO state at any given time to the one at time $t$, so we can account for the rapidly changing ISO state estimate of~\eqref{original_dyn} in our proposed framework.
\begin{definition}
\label{def_iso_flow}
A flow $\varphi^t(\textit{\oe}_{0})$, where $\textit{\oe}_{0}$ is some given ISO state, defines the solution trajectory of the autonomous ISO dynamical system~\cite{doi:10.2514/1.55705} at time $t$, which satisfies $\varphi^0(\textit{\oe}_{0})=\textit{\oe}_{0}$ at $t=0$.
\end{definition}
Utilizing the ISO flow given in Definition~\ref{def_iso_flow}, we consider the following optimal guidance problem for the ISO encounter, given estimated states $\hat{x}(\tau)$ and $\hat{\textit{\oe}}(\tau)$ in~\eqref{original_dyn} at $t=\tau$:
\begin{align}
    &u^*(\hat{x}(\tau),\hat{\textit{\oe}}(\tau),t,\rho)  \label{mpc_eq}  \\
    &\text{~}= \mathrm{arg}\min_{u(t)\in\mathcal{U}(t)}\left(c_0\|p_{\xi}(t_f)-\rho\|^2+c_1\int_{\tau}^{t_f}P\left(u(t),\xi(t)\right)dt\right) \nonumber \\
    &\text{s.t.~}\dot{\xi}(t) = f(\xi(t),\varphi^{t-\tau}(\hat{\textit{\oe}}(\tau)),t)+B(\xi(t),\varphi^{t-\tau}(\hat{\textit{\oe}}(\tau)),t)u(t) \nonumber \\
    &\text{~for }t\in(\tau,t_f],~\xi(\tau)=\hat{x}(\tau)
    \label{mpc_dynamics}
\end{align}
where $\tau \in [0,t_f)$ is the current time at which the spacecraft solves~\eqref{mpc_eq}, $\xi$ is the fictitious spacecraft relative state of the dynamics~\eqref{mpc_dynamics}, $\int_{\tau}^{t_f}P(u(t),\xi(t))dt$ is some performance-based cost function, such as $L^2$ control effort and $L^2$ trajectory tracking error, and information-based cost~\cite{doi:10.2514/6.2021-1103}, $p_{\xi}(t_f)$ is the terminal relative position of the spacecraft satisfying $\xi(t_f)=[p_{\xi}(t_f)^{\top},\dot{p}_{\xi}(t_f)^{\top}]^{\top}$, $\rho$ is a mission-specific predefined terminal position relative to the ISO at given terminal time $t_f$ ($\rho=0$ for impacting the ISO), $\mathcal{U}(t)$ is a set containing admissible control inputs, and $c_0\in\mathbb{R}_{> 0}$ and $c_1\in\mathbb{R}_{\geq 0}$ are the weights on each objective function. This paper assumes that the terminal time $t_f$ is not a decision variable but a fixed constant, as varying it is demonstrated to have a small impact on the objective value of~\eqref{mpc_eq} in our simulation setup in Sec.~\ref{sec_simulation}. Note that $\rho$ is explicitly considered as one of the inputs to $u^*$ to account for the fact that it could change depending on the target ISO. Note that the policy $u^*$ only depends on the information available at the current time $t$ without using the information of future time steps. The future estimated states are computed by integrating the nominal dynamics as mentioned in definition~\ref{def_iso_flow}.
\begin{remark}
\label{Remark_mpc_prob}
Since it is not realistic to match the spacecraft velocity with that of the ISOs due to their high inclination nature, the terminal velocity tracking error is intentionally not included in the cost function, although it could be with an appropriate choice of $P$ in~\eqref{mpc_eq}. Also, we can set $c_0=0$ and augment the problem with a constraint $\|p_{\xi}(t_f)-\rho\| = 0$ if the problem is feasible with this constraint.
\end{remark}
Since the spacecraft relative state changes dramatically due to the ISO's high relative velocity, and the actual dynamics are perturbed by the ISO and spacecraft relative state estimation uncertainty, which decreases as $t$ gets closer to $t_f$ (see Remark~\ref{remark_nav}), it is expected that the delivery error at the ISO encounter (\ie{},~$\|p_{\xi}(t_f)-\rho\|$) becomes smaller as the spacecraft solves~\eqref{mpc_eq} more frequently onboard using the updated state estimates in the initial condition~\eqref{mpc_dynamics} as in~\eqref{original_dyn}. More specifically, it is desirable to apply the optimal guidance policy solving~\eqref{mpc_eq} at each time instant $t$ as follows as in model predictive control (MPC)~\cite{mpc1}:
\begin{align}
    \label{mpc_control}
    u_{\mathrm{mpc}}(\hat{x}(t),\hat{\textit{\oe}}(t),t,\rho) = u^*(\hat{x}(t),\hat{\textit{\oe}}(t),t,\rho)
\end{align}
where $u$ is the control input of~\eqref{original_dyn}. Note that $\tau$ of $u^*$ in~\eqref{mpc_control} is now changed to $t$ unlike~\eqref{mpc_eq}, implying we only utilize the solution of~\eqref{mpc_eq} at the initial time $t =\tau$. Due to the predictive nature of the MPC, which leverages future predictions of the states $x(t)$ and $\textit{\oe}(t)$ for $t\in[\tau,t_f]$ obtained by integrating their dynamics given $\hat{x}(\tau)$ and $\hat{\textit{\oe}}(\tau)$ as in~\eqref{mpc_dynamics}, the solution of its linearized and discretized version can be shown to be near-optimal~\cite{NEURIPS2020_155fa095} in terms of dynamic regret, \ie{}, the MPC performance minus the optimal performance in hindsight. This fact could be utilized to provide local optimality of the nonlinear MPC, where proving it globally is still an open problem.

However, solving the nonlinear optimization problem~\eqref{mpc_eq} at each time instant to obtain~\eqref{mpc_control} is not realistic for a spacecraft with limited computational power. Again, even when we use MPC with receding horizons, the online computational load of solving it is not negligible, especially if the computational resource of the agent is limited and not sufficient to perform nonlinear optimization in real time. Using offline learning to replace online optimization could enable our approach applicable also to more challenging and highly nonlinear G\&C scenarios (see Sec.~\ref{sec_experiment}), where solving even a single optimization online could become unrealistic.
\subsection{Imitation Learning of MPC State and Control Trajectories}
\label{sec_node_algorithm}
In order to compute the MPC of~\eqref{mpc_control} in real time with a verifiable optimality guarantee, our proposed learning-based terminal guidance policy models it using an SN-DNN~\cite{miyato2018spectral}, which is a deep neural network constructed to be Lipschitz continuous and robust to input perturbation by design (see, \eg{}, Lemma~6.2 of~\cite{tutorial}). Note that the size of the neural network for offline learning is selected to be small enough to significantly reduce the online computational load required in solving optimization.

\begin{figure}[htbp]
    \centering
    \includegraphics[width=83mm]{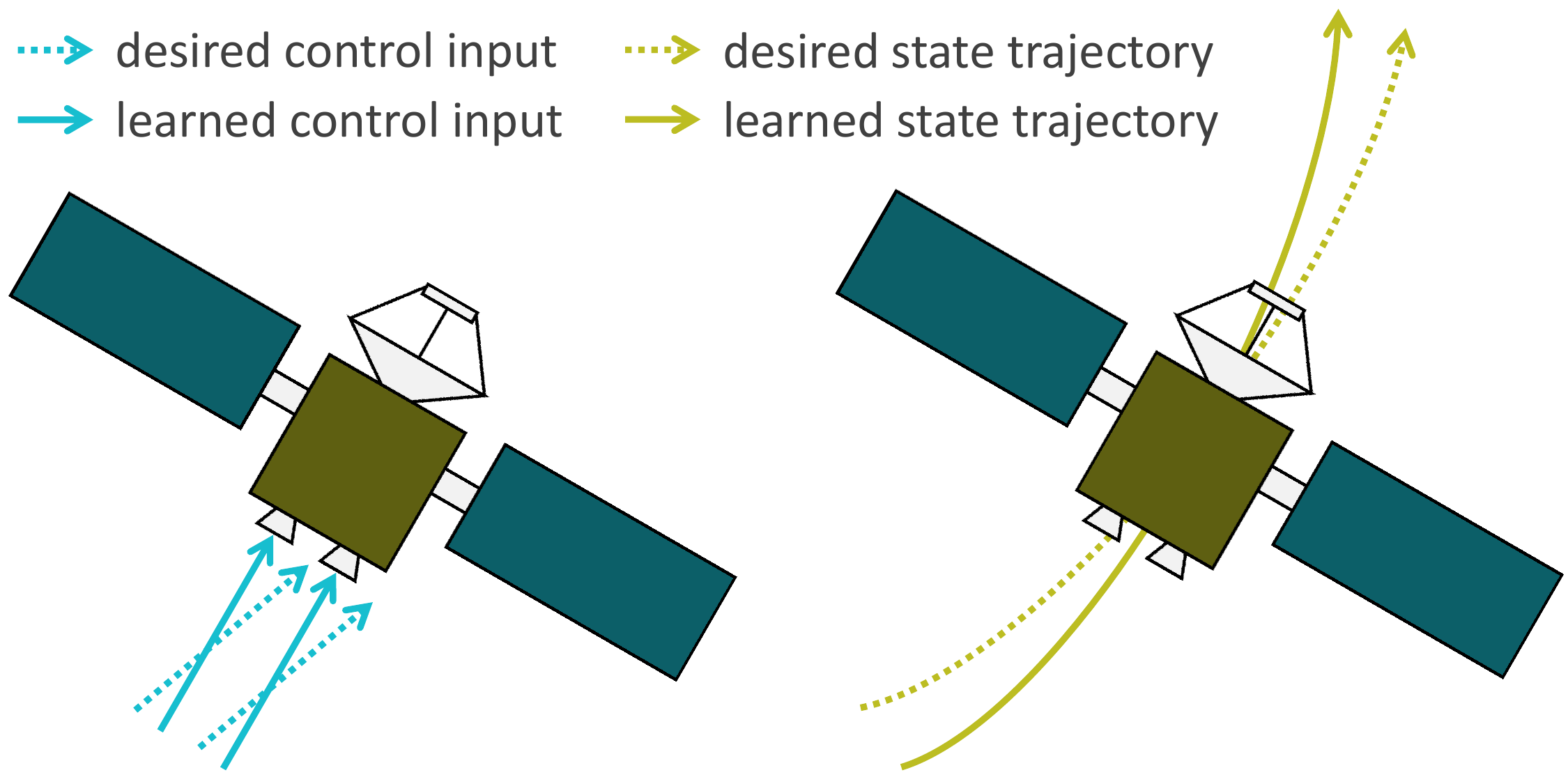}
    \caption{Left: SN-DNN imitating a desired control input (first term of~(\ref{node_loss})). Right: SN-DNN imitating a desired state trajectory (second term of~(\ref{node_loss})).}
    \label{fig_node_analogy}
\end{figure}

Let us denote the proposed learning-based terminal guidance policy as $u_{\ell}(\hat{x}(t),\hat{\textit{\oe}}(t),t,\rho;\theta_{\mathrm{nn}})$, which models the MPC policy $u_{\mathrm{mpc}}(\hat{x}(t),\hat{\textit{\oe}}(t),t,\rho)$ of~\eqref{mpc_control} using the SN-DNN, where $\theta_{\mathrm{nn}}$ is its hyperparameter. The following definition of the process induced by the spacecraft dynamics with $u_{\mathrm{mpc}}$ and $u_{\ell}$, which map the ISO and spacecraft relative state at any given time to their respective spacecraft relative state at time $t$, is useful for simplifying notation in our framework.
\begin{definition}
\label{process_definition}
Mappings denoted as $\varphi_{\ell}^t(x_{\tau},\textit{\oe}_{\tau},\tau,\rho;\theta_{\mathrm{nn}})$ and $\varphi_{\mathrm{mpc}}^t(x_{\tau},\textit{\oe}_{\tau},\tau,\rho)$ (called \textit{processes}~\cite[p. 24]{nonautonomousbook}) define the solution trajectories of the following non-autonomous dynamical systems at time $t$, controlled by the SN-DNN and MPC policy, respectively:
\begin{align}
    \dot{\xi}(t) &= f(\xi(t),\varphi^{t-\tau}(\textit{\oe}_{\tau}),t)+B(\xi(t),\varphi^{t-\tau}(\textit{\oe}_{\tau}),t) \\
    &\times u_{\ell}(\xi(t),\varphi^{t-\tau}(\textit{\oe}_{\tau}),t,\rho;\theta_{\mathrm{nn}}),~\xi(\tau) = x_{\tau} \label{whole_net_node}\\
    \dot{\xi}(t) &= f(\xi(t),\varphi^{t-\tau}(\textit{\oe}_{\tau}),t)+B(\xi(t),\varphi^{t-\tau}(\textit{\oe}_{\tau}),t)\\
    &\times u_{\mathrm{mpc}}(\xi(t),\varphi^{t-\tau}(\textit{\oe}_{\tau}),t,\rho),~\xi(\tau) = x_{\tau}
    \label{target_integration}
\end{align}
where $\tau\in[0,t_f]$, $t_f$ and $\rho$ are the given terminal time and relative position at the ISO encounter as in~\eqref{mpc_dynamics}, $\textit{\oe}_{\tau}$ and $x_{\tau}$ are some given ISO and spacecraft relative state at time $t=\tau$, respectively, $f$ and $B$ are given in~\eqref{original_dyn} and~\eqref{actualfnB}, and $\varphi^{t-\tau}(\textit{\oe}_{\tau})$ is the ISO state trajectory with $\varphi^{0}(\textit{\oe}_{\tau})=\textit{\oe}_{\tau}$ at $t=\tau$ as given in Definition~\ref{def_iso_flow}.
\end{definition}
Let $(\bar{x},\bar{\textit{\oe}},\bar{t},\bar{\rho},\Delta \bar{t})$ denote a sampled data point for the spacecraft state, ISO state, current time, desired terminal relative position, and time of integration to be used in~\eqref{node_loss}, respectively. Also, let $\mathrm{Unif}(\mathcal{S})$ be the uniform distribution over a compact set $\mathcal{S}$, which produces $(\bar{x},\bar{\textit{\oe}},\bar{t},\bar{\rho},\Delta \bar{t}) \sim \mathrm{Unif}(\mathcal{S})$. Using Definition~\ref{process_definition}, we introduce the following new loss function to be minimized by optimizing the hyperparameter $\theta_{\mathrm{nn}}$ of the SN-DNN guidance policy $u_{\ell}(\hat{x}(t),\hat{\textit{\oe}}(t),t,\rho;\theta_{\mathrm{nn}})$:
\begin{align}
    &\mathcal{L}_{\mathrm{nn}}(\theta_{\mathrm{nn}}) = \mathbb{E}\left[\|u_{\ell}(\bar{x},\bar{\textit{\oe}},\bar{t},\bar{\rho};\theta_{\mathrm{nn}})-u_{\mathrm{mpc}}(\bar{x},\bar{\textit{\oe}},\bar{t},\bar{\rho})\|_{C_u}^2\right. \\
    &\left.+\|\varphi_{\ell}^{\bar{t}+\Delta\bar{t}}(\bar{x},\bar{\textit{\oe}},\bar{t},\bar{\rho};\theta_{\mathrm{nn}})-\varphi_{\mathrm{mpc}}^{\bar{t}+\Delta\bar{t}}(\bar{x},\bar{\textit{\oe}},\bar{t},\bar{\rho})\|_{C_x}^2\right]
    \label{node_loss}
\end{align}
where $\|(\cdot)\|_{C_x}$ and $\|(\cdot)\|_{C_u}$ are the weighted Euclidean $2$-norm given as $\|(\cdot)\|_{C_u}^2=(\cdot)^{\top}C_u(\cdot)$ and $\|(\cdot)\|_{C_x}^2=(\cdot)^{\top}C_x(\cdot)$ for symmetric positive definite weight matrices $C_u,C_x\succ 0$, and $\varphi_{\ell}^t(\bar{x},\bar{\textit{\oe}},\bar{t},\bar{\rho};\theta_{\mathrm{nn}})$ and $\varphi_{\mathrm{mpc}}^t(\bar{x},\bar{\textit{\oe}},\bar{t},\bar{\rho})$ are the solution trajectories with the SN-DNN and MPC guidance policy given in Definition~\ref{process_definition}, respectively. As illustrated in Fig.~\ref{fig_node_analogy}, we train the SN-DNN to also minimize the deviation of the state trajectory with the SN-DNN guidance policy from the desired MPC state trajectory~\cite{neuralode,meta_trj_loss}. The learning objective is thus not just to minimize $\|u_{\ell}-u_{\mathrm{mpc}}\|$, but to imitate the optimal trajectory with the smallest spacecraft delivery error at the ISO encounter. We will see how the selection of the weights $C_u$ and $C_x$ affects the performance of $u_{\ell}$ in Sec.~\ref{sec_simulation}.
\begin{remark}
\label{remark_empirical_loss}
As in standard learning algorithms for neural networks, including stochastic gradient descent (SGD), the expectation of the loss function~\eqref{node_loss} can be approximated using sampled data points as follows:
\begin{align}
    &\mathcal{L}_{\mathrm{emp}}(\theta_{\mathrm{nn}}) = \sum_{i=1}^{N_{d}}\|u_{\ell}(\bar{x}_{i},\bar{\textit{\oe}}_{i},\bar{t}_i,\bar{\rho}_i;\theta_{\mathrm{nn}})-u_{\mathrm{mpc}}(\bar{x}_{i},\bar{\textit{\oe}}_{i},\bar{t}_i,\bar{\rho}_i)\|^2_{C_u} \nonumber\\
    &+\|\varphi_{\ell}^{\bar{t}_i+\Delta\bar{t}_i}(\bar{x}_{i},\bar{\textit{\oe}}_{i},\bar{t}_i,\bar{\rho}_i;\theta_{\mathrm{nn}})-\varphi_{\mathrm{mpc}}^{\bar{t}_i+\Delta\bar{t}_i}(\bar{x}_{i},\bar{\textit{\oe}}_{i},\bar{t}_i,\bar{\rho}_i)\|^2_{C_x}
    \label{node_empirical_loss}
\end{align}
where the training data points $\{(\bar{x}_{i},\bar{\textit{\oe}}_{i},\bar{t}_i,\bar{\rho}_i,\Delta \bar{t}_i)\}_{i=1}^{N_{d}}$ are drawn independently from $\mathrm{Unif}(\mathcal{S})$. 
\end{remark}
\subsection{Optimality Gap of Deep Learning-Based Guidance}
\begin{figure}[htbp]
    \centering
    \includegraphics[width=83mm]{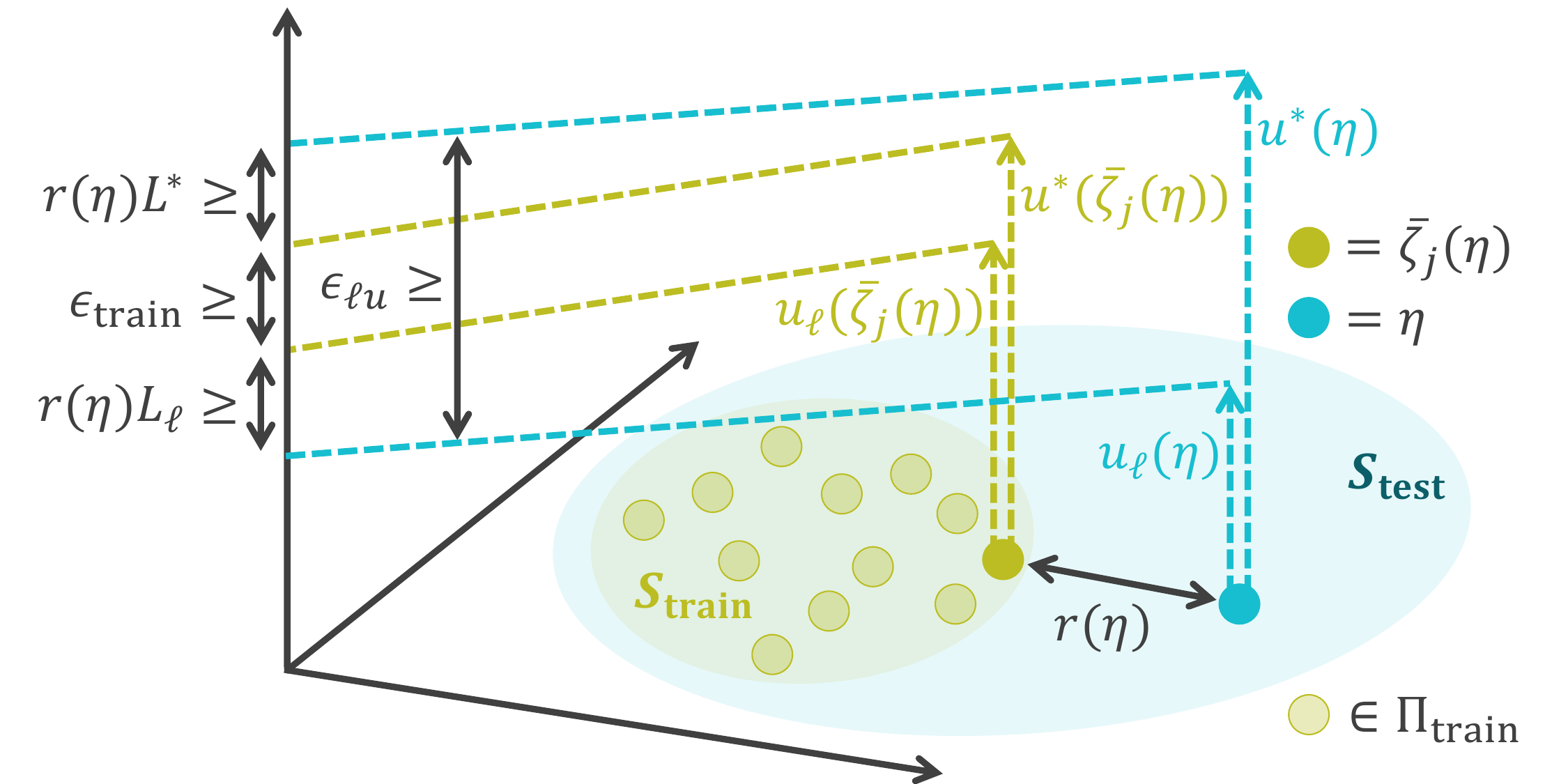}
    \caption{Illustration of the optimality gap~(\ref{sn_learning_error}), where $\bm{\mathcal{S}_{\mathrm{train}}}$ denotes the training set in which training data $\bm{(\bar{x}_i,\bar{\textit{\ae}}_i,\bar{t}_i,\bar{\rho}_i) \in \Pi_{\mathrm{train}}}$ is sampled.}
    \label{fig_exp_sndnn_bound}
\end{figure}

Since an SN-DNN is Lipschitz bounded by design and robust to perturbation, the optimality gap of the guidance framework $u_{\ell}$ introduced in Sec.~\ref{sec_node_algorithm} can be bounded as in the following theorem, where the notations are summarized in Table~\ref{tab:notations_thm1} and the proof concept is illustrated in Fig.~\ref{fig_exp_sndnn_bound}.
\begin{table*}[htbp]
\caption{Notations in Lemma~\ref{Lemma:SNDNNlearning}. \label{tab:notations_thm1}}
\vspace{-1em}
\footnotesize
\begin{center}
\renewcommand{\arraystretch}{1.4}
\begin{tabular}{ m{2.5cm} m{12cm} } 
\hline
\hline
Notation & Description \\ \hline
$L_{\ell}$ & $2$-norm Lipschitz constant of $u_{\ell}$ in $\mathbb{R}_{>0}$, guaranteed to exist by design \\
$N_{d}$ & Number of training data points \\
$\mathcal{S}_{\mathrm{test}}$ & Any compact test set containing $(x,\textit{\oe},t,\rho)$, not necessarily the training set $\mathcal{S}_\mathrm{train}$ itself \\
$u_{\ell}$ & Learning-based terminal guidance policy that models $u_{\mathrm{mpc}}$ by the SN-DNN using the loss function given in~\eqref{node_loss} \\
$u_{\mathrm{mpc}}$ & Optimal MPC terminal guidance policy given in~\eqref{mpc_control} \\
$(x,\textit{\oe},t,\rho)$ & Test data point for the S/C relative state, ISO state, current time, and desired terminal relative position of~\eqref{mpc_dynamics}, respectively \\
$(\bar{x}_i,\bar{\textit{\oe}}_i,\bar{t}_i,\bar{\rho}_i)$ & Training data point for the S/C relative state, ISO state, current time, and desired terminal relative position of~\eqref{mpc_dynamics}, respectively, where $i\in\mathbb{N}\cup[1,N_{d}]$ \\
$\Pi_{\mathrm{train}}$ & Training dataset containing a finite number of training data points, \ie{}, $\Pi_{\mathrm{train}}=\{(\bar{x}_i,\bar{\textit{\oe}}_i,\bar{t}_i,\bar{\rho}_i)\}_{i=1}^{N_{d}}$ \\
\hline
\hline
\end{tabular}
\end{center}
\vspace{-1em}
\end{table*}
\begin{lemma}
\label{Lemma:SNDNNlearning}
Suppose that $u_{\mathrm{mpc}}$ is Lipschitz with its $2$-norm Lipschitz constant $L_{\mathrm{mpc}}\in\mathbb{R}_{> 0}$. If $u_{\ell}$ is trained using the empirical loss function~\eqref{node_empirical_loss} of Remark~\ref{remark_empirical_loss} to have $\exists\epsilon_{\mathrm{train}}\in\mathbb{R}_{\geq 0}$ \st{}
\begin{align}
\label{training_error}
\sup_{i\in\mathbb{N}\cup[1,N_{d}]}\|u_{\ell}(\bar{x}_{i},\bar{\textit{\oe}}_{i},\bar{t}_i,\bar{\rho}_i;\theta_{\mathrm{nn}})-u_{\mathrm{mpc}}(\bar{x}_{i},\bar{\textit{\oe}}_{i},\bar{t}_i,\bar{\rho}_i)\| \leq \epsilon_{\mathrm{train}}~~~~~~~~~~~~~~~
\end{align}
then we have the following bound:
\begin{align}
\label{sn_learning_error}
&\|u_{\ell}(x,\textit{\oe},t,\rho;\theta_{\mathrm{nn}})-u_{\mathrm{mpc}}(x,\textit{\oe},t,\rho)\| \\
&\leq \epsilon_{\mathrm{train}}+r(x,\textit{\oe},t,\rho)(L_{\ell}+L_{\mathrm{mpc}}) = \epsilon_{\ell u},~\forall (x,\textit{\oe},t,\rho) \in \mathcal{S}_{\mathrm{test}}
\end{align}
where $r(x,\textit{\oe},t,\rho)$ is given by
\begin{align}
    &r(x,\textit{\oe},t,\rho) \\
    &= \inf_{i\in\mathbb{N}\cup[1,N_{d}]}\sqrt{\|\bar{x}_{i}-x\|^2+\|\bar{\textit{\oe}}_{i}-\textit{\oe}\|^2+(\bar{t}_i-t)^2+\|\bar{\rho}_i-\rho\|^2}.
\end{align}
\end{lemma}
\begin{proof}
Let $\eta=(x,\textit{\oe},t,\rho)$ be a test element in $\mathcal{S}_\mathrm{test}$ (\ie{}, $\eta\in\mathcal{S}_\mathrm{test}$) and let $\bar{\zeta}_{j}(\eta)$ be the training data point in $\Pi_{\mathrm{train}}$ (\ie{}, $\bar{\zeta}_{j}(\eta)\in\Pi_{\mathrm{train}}$) that achieves the infimum of $r(\eta)$ with $i=j$ for a given $\eta \in \mathcal{S}_\mathrm{test}$. Since $u_{\ell}$ and $u_{\mathrm{mpc}}$ are Lipschitz by design and by assumption, respectively, we have for any $\eta \in \mathcal{S}_\mathrm{test}$ that
\begin{align}
&\|u_{\ell}(\eta;\theta_{\mathrm{nn}})-u_{\mathrm{mpc}}(\eta)\| \leq \|u_{\ell}(\bar{\zeta}_{j}(\eta);\theta_{\mathrm{nn}})-u_{\mathrm{mpc}}(\bar{\zeta}_{j}(\eta))\| \\
&+\|u_{\ell}(\eta;\theta_{\mathrm{nn}})-u_{\ell}(\bar{\zeta}_{j}(\eta);\theta_{\mathrm{nn}})\|+\|u_{\mathrm{mpc}}(\eta)-u_{\mathrm{mpc}}(\bar{\zeta}_{j}(\eta))\| \nonumber \\
&\leq \|u_{\ell}(\bar{\zeta}_{j}(\eta);\theta_{\mathrm{nn}})-u_{\mathrm{mpc}}(\bar{\zeta}_{j}(\eta))\|+r(\eta)(L_{\ell}+L_{\mathrm{mpc}}) \\
&\leq \epsilon_{\mathrm{train}}+r(\eta)(L_{\ell}+L_{\mathrm{mpc}})
\end{align}
where the second inequality follows from the definition of $\bar{\zeta}_{j}(\eta)$ and the third inequality follows from~\eqref{training_error} and the fact that $\bar{\zeta}_{j}(\eta)\in\Pi_{\mathrm{train}}$. This relation leads to the desired result~\eqref{sn_learning_error} as it holds for any $\eta \in \mathcal{S}_\mathrm{test}$.
\end{proof}
The SN-DNN provides a verifiable optimality gap even for data points not in its training set, which indicates that it still benefits from the near-optimal guarantee of the MPC in terms of dynamic regret as discussed below~\eqref{mpc_control}. As illustrated in Fig.~\ref{fig_exp_sndnn_bound}, each term in the optimality gap~\eqref{sn_learning_error} can be interpreted as follows:
\begin{enumerate}
    \item $\epsilon_{\mathrm{train}}$ of~\eqref{training_error} is the training error of the SN-DNN, which is expected to decrease as the learning proceeds using SGD. In general, the value of $\epsilon_{\mathrm{train}}$ is affected by the choice of Lipschitz constant $L_{\ell}$ (see~\cite{marginbounds1} for more details).
    \item $r(\eta)$ is the closest distance from the test element $\eta \in \mathcal{S}_{\mathrm{test}}$ to a training data point in $\Pi_{\mathrm{train}}$, expected to decrease as the number of data points $N_{d}$ in $\Pi_{\mathrm{train}}$ increases and as the training set $\mathcal{S}_{\mathrm{train}}$ gets larger.
    \item The Lipschitz constant $L_{\ell}$ is a design parameter we could arbitrarily choose when constructing the SN-DNN. Note that the SN-DNN is Lipschitz \textit{by design} by spectrally normalizing the weights of the neural network~\cite{miyato2018spectral}, and thus we can freely choose $L_{\ell}$ independently of the choice of the neural net parameters.
    \item We could also treat $L_{\mathrm{mpc}}$ as a design parameter by adding a Lipschitz constraint, $\|\partial u_{\mathrm{mpc}}/\partial \eta\| \leq L_{\mathrm{mpc}}$, in solving~\eqref{mpc_eq}.
\end{enumerate}
Note that $\epsilon_{\mathrm{train}}$ and $r(\eta)$ can always be computed numerically for a given $\eta$ as the dataset $\Pi_{\mathrm{train}}$ only has a finite number of data points. 

The obvious downsides of the optimality gap~(\ref{sn_learning_error}) are that (i) it still suffers the generalization issue, which is common in the field of machine learning, and (ii) we may have to empirically find the region that the MPC is locally Lipschitz. Also, in practice, although the SN-DNN is originally designed for a better generalization property, a smaller Lipschitz constant $L_{\ell}$ could lead to a larger training error $\epsilon_{\mathrm{train}}$, requiring some tuning by trading-off the training error and test error in real-world simulations (see Sec.~\ref{sec_simulation} and Sec.~\ref{sec_experiment})~\cite{marginbounds1}. The benefit of having Lemma~\ref{Lemma:SNDNNlearning} is to explicitly describe all the assumptions we need to mathematically obtain the bound on the learning error, without treating it like a black box.
\begin{remark}
For controllers whose optimal solutions are not necessarily Lipschitz (\eg{}, impulsive-type controllers), the SN-DNN is not a suitable representation due to its Lipschitz assumption, which is one of the limitations of the bound in Lemma~\ref{Lemma:SNDNNlearning}. Although we have demonstrated the practicality of this assumption for our cases in Sec.~\ref{sec_experiment} (and in~\cite{our_iso_mission} with the comparison of our method with impulsive maneuvers), improving the expressibility of our method is required for its more general applications with more general mathematical guarantees. This is left as one of our future works.
\end{remark}
\begin{remark}
\label{remark_guidance_state_error}
Obtaining a tighter and more general optimality gap of the learned control and state trajectories has been an active field of research. Particularly for a neural network equipped with a dynamical structure as in our proposed approach, we could refer to generalization bounds for the neural ordinary differential equations~\cite{neuralode}, or use systems and control theoretical methods to augment it with stability and robustness properties~\cite{contractive_node,lyanet}. We could also consider combining the SN-DNN with neural networks that have enhanced structural guarantees of robustness, including robust implicit networks~\cite{implicit1} and robust equilibrium networks~\cite{eqnet1}.
\end{remark}
The optimality gap discussed in Lemma~\ref{Lemma:SNDNNlearning}~and~in Remark~\ref{remark_guidance_state_error} is useful in that they provide mathematical guarantees for learning-based frameworks only with the Lipschitz assumption (\eg{}, it could prove safety in the learning-based MPC framework~\cite{learningmpc}). However, when the system is perturbed by the state uncertainty as in~\eqref{original_dyn}, we could only guarantee that the distance between the state trajectories controlled by $u_{\mathrm{mpc}}$ of~\eqref{mpc_control} and $u_{\ell}$ of Lemma~\ref{Lemma:SNDNNlearning} is bounded by a function that increases exponentially with time~\cite{tutorial}. In Sec.~\ref{sec_control}~and~\ref{sec_control_proof}, we will see how such a conservative bound can be replaced by a decreasing counterpart.
\section{Deep learning-Based Optimal Tracking Control}
\label{sec_control}
This section proposes a feedback control policy to be used on top of the SN-DNN terminal guidance policy $u_{\ell}$ of Lemma~\ref{Lemma:SNDNNlearning} in Sec.~\ref{sec_guidance}. In particular, we utilize $u_{\ell}$ to design the desired trajectory with zero spacecraft delivery error and then construct pointwise optimization-based tracking control with a Lyapunov stability condition, which can be shown to have an analytical solution for real-time implementation. In Sec.~\ref{sec_control_proof}, we will see that this framework plays an essential part in solving our main problem~\ref{Problem1} of Sec.~\ref{sec_problem}.
\begin{remark}
\label{remark_rho_dependence}
For notational convenience, we drop the dependence on the desired terminal relative position $\rho$ of~\eqref{mpc_dynamics} in $u_{\ell}$, $u_{\mathrm{mpc}}$, $\varphi_{\ell}^t$, and $\varphi_{\mathrm{mpc}}^t$ in the subsequent sections, as it is time-independent and thus does not affect the arguments to be made in the following. Note that the SN-DNN is still to be trained regarding $\rho$ as one of its inputs, so we do not have to retrain SN-DNNs every time we change $\rho$. 
\end{remark}
\subsection{Desired State Trajectory with Zero Delivery Error}
\label{sec_target}
\begin{figure}[htbp]
    \centering
    \includegraphics[width=83mm]{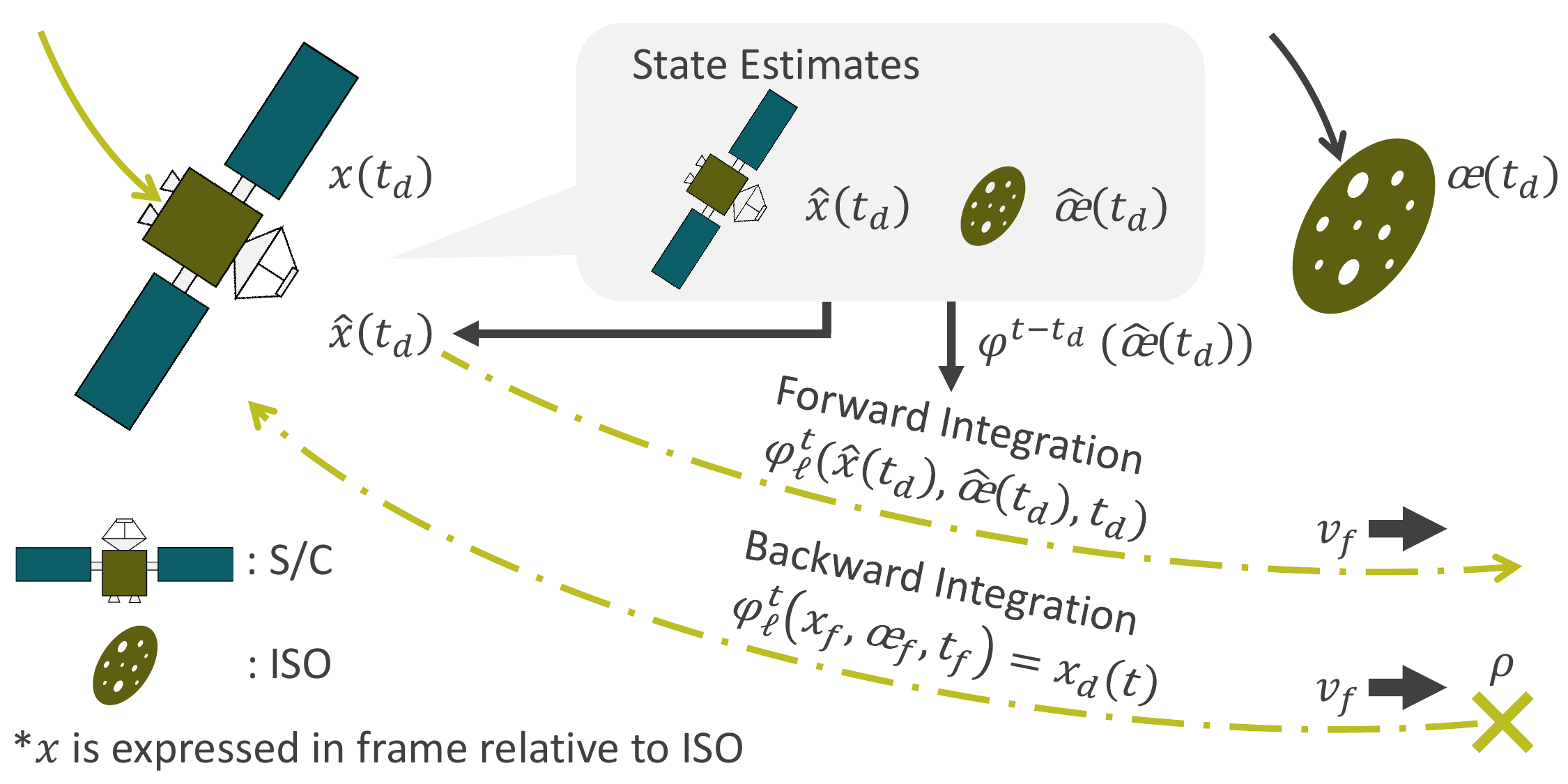}
    \caption{Illustration of the desired trajectory in Sec.~\ref{sec_target} for $\bm{v_f = C_{\mathrm{s2v}}\varphi_{\ell}^{t_f}(\hat{x}(t_d),\hat{\textit{\oe}}(t_d),t_d)}$ in~(\ref{xf_verifiable}). Spacecraft obtains $\bm{x_d}$ of~(\ref{target_trajectory}) via backward integration (see Remark~\ref{remark_ts}).}
    \label{fig_target_trajectory}
\end{figure}

Let us recall the definitions of the ISO state flow and spacecraft state processes of Definitions~\ref{def_iso_flow}~and~\ref{process_definition} summarized in Table~\ref{tab_flow_recap}. Ideally at some given time $t=t_d\in[0,t_f)$ when the spacecraft possesses enough information on the ISO, we would like to construct a desired trajectory using the estimated states $\hat{\textit{\oe}}(t_d)$ and $\hat{x}(t_d)$ \st{} it ensures zero spacecraft delivery error at the ISO encounter assuming zero state estimation errors for $t\in [t_d,t_f]$ in~\eqref{original_dyn}. As in trajectory generation in space and aerial robotics~\cite{backward1,backward2}, this can be achieved by obtaining a desired state trajectory as $\varphi_{\ell}^t(x_f,\textit{\oe}_{f},t_f)$ of Table~\ref{tab_flow_recap} with $\textit{\oe}_{f} = \varphi^{t_f-t_d}(\hat{\textit{\oe}}(t_d))$, \ie{}, by solving the following dynamics with the SN-DNN terminal guidance policy $u_{\ell}$~\eqref{whole_net_node} backward in time:
\begin{align}
    \dot{\xi}(t) &= f(\xi(t),\varphi^{t-t_d}(\hat{\textit{\oe}}(t_d)),t)+B(\xi(t),\varphi^{t-t_d}(\hat{\textit{\oe}}(t_d)),t) \\
    &\times u_{\ell}(\xi(t),\varphi^{t-t_d}(\hat{\textit{\oe}}(t_d)),t;\theta_{\mathrm{nn}}),~\xi(t_f) = x_{f}
    \label{backward_target}
\end{align}
where the property of the ISO solution flow in Table~\ref{tab_flow_recap} introduced in Definition~\ref{def_iso_flow}, $\varphi^{t-t_f}(\textit{\oe}_{f}) = \varphi^{t-t_f}(\varphi^{t_f-t_d}(\textit{\oe}(t_d))) = \varphi^{t-t_d}(\textit{\oe}(t_d))$, is used to get~\eqref{backward_target}, $t_f$ is the given terminal time at the ISO encounter as in~\eqref{mpc_dynamics}, and the ideal spacecraft terminal relative state $x_f$ is defined as
\begin{equation}
    \label{xf_verifiable}
    x_f = \left[\begin{smallmatrix}\rho \\ C_{\mathrm{s2v}}\varphi_{\ell}^{t_f}(\hat{x}(t_d),\hat{\textit{\oe}}(t_d),t_d)\end{smallmatrix}\right]
\end{equation}
where $\rho$ is the desired terminal relative position given in~\eqref{mpc_dynamics}, $\varphi_{\ell}^{t_f}(\hat{x}(t_d),\hat{\textit{\oe}}(t_d),t_d)$ is the spacecraft relative state at $t=t_f$ obtained by integrating the dynamics forward as in Table~\ref{tab_flow_recap}, and $C_{\mathrm{s2v}} = [\mathrm{O}_{3\times3}~\mathrm{I}_{3\times3}]\in\mathbb{R}^{3\times 6}$ is a matrix that maps the spacecraft relative state to its velocity vector. Figure~\ref{fig_target_trajectory} illustrates the construction of such a desired trajectory. 
\begin{table*}[htbp]
\caption{Summary of the flow and processes in Definitions~\ref{def_iso_flow}~and~\ref{process_definition}, where $\bm{u_{\ell}}$ and $\bm{u_{\mathrm{mpc}}}$ are the SN-DNN and MPC guidance policies, respectively. The dependence on $\bm{\rho}$ is omitted (see Remark~\ref{remark_rho_dependence}). \label{tab_flow_recap}}
\vspace{-1em}
\footnotesize
\begin{center}
\renewcommand{\arraystretch}{1.4}
\begin{tabular}{ m{2.5cm} m{12cm} } 
\hline
\hline
Notation & Description \\ \hline
$\varphi^{t-\tau}(\textit{\oe}_{\tau})$ & Solution trajectory of the ISO dynamics at time $t$ which satisfies $\varphi^{0}(\textit{\oe}_{\tau})=\textit{\oe}_{\tau}$ \\
$\varphi_{\ell}^t(x_{\tau},\textit{\oe}_{\tau},\tau;\theta_{\mathrm{nn}})$ & Solution trajectory of the S/C relative dynamics at time $t$, controlled by $u_{\ell}$ with no state estimation error, which satisfies $\varphi_{\ell}^\tau(x_{\tau},\textit{\oe}_{\tau},\tau;\theta_{\mathrm{nn}})=x_{\tau}$ at $t=\tau$ and $\textit{\oe}(t)=\varphi^{t-\tau}(\textit{\oe}_{\tau})$ \\
$\varphi_{\mathrm{mpc}}^t(x_{\tau},\textit{\oe}_{\tau},\tau)$ & Solution trajectory of the S/C relative dynamics at time $t$, controlled by $u_{\mathrm{mpc}}$ with no state estimation error, which satisfies $\varphi_{\mathrm{mpc}}^\tau(x_{\tau},\textit{\oe}_{\tau},\tau)=x_{\tau}$ at $t=\tau$ and $\textit{\oe}(t)=\varphi^{t-\tau}(\textit{\oe}_{\tau})$ \\
\hline
\hline
\end{tabular}
\end{center}
\vspace{-1em}
\end{table*}
\begin{remark}
\label{remark_ts}
Although the desired state trajectory design depicted in Fig.~\ref{fig_target_trajectory} involves backward and forward integration~\eqref{backward_target}~and~\eqref{xf_verifiable} as in Definition~\ref{process_definition}, which is common in trajectory generation in space and aerial robotics~\cite{backward1,backward2}, the SN-DNN approximation of the MPC policy allows performing it within a short period. In fact, when measured using the Mid 2015 MacBook Pro laptop, it takes only about $3.0\times10^{-4}$~\si{\second} for numerically integrating the dynamics for one step using the fourth-order Runge–Kutta method (\eg{}, it takes $\sim 3$~\si{\second} to get the desired trajectory for $t_d=0$~{\rm(\si{\second})} and $t_f=10000$~{\rm(\si{\second})} with the discretization time step~\SI{1}{\second}). The computational time should decrease as $t_d$ becomes larger. Section~\ref{sec_mpc_and_min_norm} delineates how we utilize such a desired trajectory in real time with the feedback control to be introduced in Sec.~\ref{sec_tracking_control}.
\end{remark}
\subsection{Deep Learning-Based Pointwise Min-Norm Control}
\label{sec_tracking_control}
Using the results given in Sec.~\ref{sec_target}, we design pointwise optimal tracking control resulting in verifiable spacecraft delivery error, even under the presence of state uncertainty. For notational simplicity, let us denote the desired spacecraft relative state and ISO state trajectories of~\eqref{backward_target}, constructed using $\hat{x}(t_d)$ and $\hat{\textit{\oe}}(t_d)$ at $t=t_d$ with the terminal state~\eqref{xf_verifiable} as illustrated in Fig.~\ref{fig_target_trajectory}, as follows:
\begin{align}
    \label{target_trajectory}
    x_d(t) = \varphi_{\ell}^t(x_f,\textit{\oe}_{f},t_f),~\textit{\oe}_d(t) = \varphi^{t-t_d}(\hat{\textit{\oe}}(t_d)).
\end{align}
Also, assuming that we select $x$ as $x = [p^{\top},\dot{p}^{\top}]^{\top}$ as the state of~\eqref{original_dyn} as in~\eqref{actualfnB}, where $p \in \mathbb{R}^3$ is the position of the spacecraft relative to the ISO, let $\mathscr{f}(x,\textit{\oe},t)$ and $\mathscr{B}(x,\textit{\oe},t)$ be defined as follows:
\begin{align}
    \label{fnB_pdotdynamics}
    \mathscr{f}(x,\textit{\oe},t) = C_{\mathrm{s2v}}f(x,\textit{\oe},t),~\mathscr{B}(x,\textit{\oe},t) = C_{\mathrm{s2v}}B(x,\textit{\oe},t)
\end{align}
where $f$ and $B$ are given in~\eqref{original_dyn} and $C_{\mathrm{s2v}} = [\mathrm{O}_{3\times3}~\mathrm{I}_{3\times3}]\in\mathbb{R}^{3\times 6}$ as in~\eqref{xf_verifiable}. Note that the relations~\eqref{fnB_pdotdynamics} imply
\begin{align}
    \label{pdotdynamics}
    \ddot{p}=\mathscr{f}(x,\textit{\oe},t)+\mathscr{B}(x,\textit{\oe},t)u(\hat{x},\hat{\textit{\oe}},t).
\end{align}

Given the desired trajectory~\eqref{target_trajectory}, $x_d = [p_d^{\top},\dot{p}_d^{\top}]^{\top}$, which achieves zero spacecraft delivery error at the ISO encounter when the state estimation errors are zero for $t\in [t_d,t_f]$ in~\eqref{original_dyn}, we design a controller $u$ of~\eqref{original_dyn}~and~\eqref{pdotdynamics} as follows, as considered in~\cite{6709752,tutorial} for general nonlinear systems:
\begin{align}
    \label{min_norm_controller}
    u^*_{\ell}(\hat{x},\hat{\textit{\oe}},t;\theta_{\mathrm{nn}}) &= u_{\ell}(x_d(t),\textit{\oe}_d(t),t;\theta_{\mathrm{nn}})+k(\hat{x},\hat{\textit{\oe}},t) \\
    \label{min_norm_controller_k}
    k(\hat{x},\hat{\textit{\oe}},t) &= \begin{cases} 0 &\text{if }\Upsilon(\hat{x},\hat{\textit{\oe}},t) \leq 0 \\ -\frac{\Upsilon(\hat{x},\hat{\textit{\oe}},t)(\dot{\hat{p}}-\varrho_d(\hat{p},t))}{\|\dot{\hat{p}}-\varrho_d(\hat{p},t)\|^2} &\text{otherwise} \end{cases}
\end{align}
with $\Upsilon$ and $\varrho_d$ defined as 
\begin{align}
    \label{def_upsilon}
    \Upsilon &= (\dot{\hat{p}}-\varrho_d(\hat{p},t))^{\top}M(t)(\mathscr{f}(\hat{x},\hat{\textit{\oe}},t)-\mathscr{f}(x_d(t),\textit{\oe}_d(t),t) \\
    &+\Lambda(\dot{\hat{p}}-\dot{p}_d(t))+\alpha(\dot{\hat{p}}-\varrho_d(\hat{p},t))) \\
    \label{def_varrho}
    \varrho_d &= -\Lambda(\hat{p}-p_d(t))+\dot{p}_d(t)
\end{align}
where $\hat{p} \in \mathbb{R}^3$ is the estimated position of the spacecraft relative to the ISO \st{} $\hat{x} = [\hat{p}^{\top},\dot{\hat{p}}^{\top}]^{\top}$, $u_{\ell}$ is the SN-DNN terminal guidance policy of Lemma~\ref{Lemma:SNDNNlearning}, $\textit{\oe}_d$, $\mathscr{f}$, and $\mathscr{B}$ are given in~\eqref{target_trajectory}~and~\eqref{fnB_pdotdynamics}, $\Lambda\succ0$ is a given symmetric positive definite matrix, $\alpha\in\mathbb{R}_{> 0}$ is a given positive constant, and $M(t) = m_{\mathrm{sc}}(t)\mathrm{I}_{3\times3}$ for the spacecraft mass $m_{\mathrm{sc}}(t)$ given in~\eqref{actualfnB}. This control policy can be shown to possess the following pointwise optimality property, in addition to the robustness and stability guarantees to be seen in Sec.~\ref{sec_control_proof}. Note that \eqref{min_norm_controller} is well-defined even with the division by $\|\dot{\hat{p}}-\varrho_d(\hat{p},t)\|$ because we have $\Upsilon(\hat{x},\hat{\textit{\oe}},t) = 0$ when $\|\dot{\hat{p}}-\varrho_d(\hat{p},t)\| = 0$ and thus $k(\hat{x},\hat{\textit{\oe}},t) = 0$ in this case.

Consider the following optimization problem, which computes an optimal control input that minimizes its instantaneous deviation from that of the SN-DNN guidance policy at each time instant, under an incremental Lyapunov stability condition:
\begin{align}
    \label{min_norm_controller_prob}
    &u^*_{\ell}(\hat{x},\hat{\textit{\oe}},t;\theta_{\mathrm{nn}}) = \mathrm{arg}\min_{u\in\mathbb{R}^{m}}\|u-u_{\ell}(x_d(t),\textit{\oe}_d(t),t;\theta_{\mathrm{nn}})\|^2 
\end{align}
\begin{align}
    &\text{\st{}~}\frac{\partial V}{\partial \dot{\hat{p}}}(\dot{\hat{p}},\varrho_d(\hat{p},t),t)(\mathscr{f}(\hat{x},\hat{\textit{\oe}},t)+\mathscr{B}(\hat{x},\hat{\textit{\oe}},t)u)\\
    &\text{\color{white}\st{}~}+\frac{\partial V}{\partial \varrho_d}(\dot{\hat{p}},\varrho_d(\hat{p},t),t)(\ddot{p}_d(t)-\Lambda(\dot{\hat{p}}-\dot{p}_d(t))) \\
    &\text{\color{white}\st{}~}\leq -2\alpha V(\dot{\hat{p}},\varrho_{d}(\hat{p},t),t)
    \label{Vstability}
\end{align}
where $V$ is a non-negative function defined as
\begin{align}
    \label{def_lyapunov}
    V(\dot{\hat{p}},\varrho_d,t) &= (\dot{\hat{p}}-\varrho_d)^{\top}M(t)(\dot{\hat{p}}-\varrho_d)
\end{align}
and the other notation is as given in~\eqref{min_norm_controller}.
\begin{lemma}
\label{lemma_feasible}
The optimization problem~\eqref{min_norm_controller_prob} is always feasible, and the controller~\eqref{min_norm_controller} defines its analytical optimal solution for the spacecraft relative dynamical system~\eqref{actualfnB}. Furthermore, substituting $u=u_{\ell}(x_d(t),\textit{\oe}_d(t),t;\theta_{\mathrm{nn}})$ into~\eqref{Vstability} yields $\Upsilon(\hat{x},\hat{\textit{\oe}},t) \leq 0$, which implies the controller~\eqref{min_norm_controller} modulates the desired input, $u_{\ell}(x_d(t),\textit{\oe}_d(t),t;\theta_{\mathrm{nn}})$, only when necessary to ensure the stability condition~\eqref{Vstability}.
\end{lemma}
\begin{proof}
Let $u_n(\hat{x},\hat{\textit{\oe}},t)$ be defined as follows:
\begin{align}
    u_{n} &= M(t)\dot{\varrho}_{d}(\hat{p},t)+C(\hat{\textit{\oe}})\varrho_{d}(\hat{p},t)+G(\hat{p},\hat{\textit{\oe}})-\alpha M(t) \nonumber \\
    &\times (\dot{\hat{p}}-\varrho_{d}(\hat{p},t))
\end{align}
where $\varrho_{d}$ is as given in~\eqref{def_varrho}. Substituting this into the left-hand side of~\eqref{Vstability} as $u=u_{n}(\hat{x},\hat{\textit{\oe}},t)$ gives
\begin{align}
    \label{skew_sym_lyap}
    &2(\dot{\hat{p}}-\varrho_d(\hat{p},t))^{\top}(-C(\hat{\textit{\oe}})-\alpha M(t))(\dot{\hat{p}}-\varrho_d(\hat{p},t)) \\
    &= -2\alpha V(\dot{\hat{p}},\varrho_{d}(\hat{p},t),t)
\end{align}
where $\mathscr{f}$ of~\eqref{fnB_pdotdynamics} is computed with~\eqref{actualfnB}, and the equality follows from the skew-symmetric property of $C$, \ie{}, $C(\hat{\textit{\oe}})+C(\hat{\textit{\oe}})^{\top}=\mathrm{O}_{3\times 3}$~\cite{doi:10.2514/1.37261}. The relation indeed indicates that $u=u_n(\hat{x},\hat{\textit{\oe}},t)$ is a feasible solution to the optimization problem~\eqref{min_norm_controller_prob}. Furthermore, applying the KKT condition to~\eqref{min_norm_controller_prob} yields~\eqref{min_norm_controller}. Also, we get the condition $\Upsilon(\hat{x},\hat{\textit{\oe}},t) \leq 0$ by substituting $V$ of~\eqref{def_lyapunov} and $\ddot{p}$ into the stability condition~\eqref{Vstability} with $u=u_{\ell}(x_d(t),\textit{\oe}_d(t),t;\theta_{\mathrm{nn}})$.
\end{proof}
Let us emphasize again that, as proven in Lemma~\ref{lemma_feasible}, the deviation term $k(\hat{x},\hat{\textit{\oe}},t)$ of the controller~\eqref{min_norm_controller} is non-zero only when the stability condition~\eqref{Vstability} cannot be satisfied with the SN-DNN terminal guidance policy $u_{\ell}$ of Lemma~\ref{Lemma:SNDNNlearning}. This result can be viewed as an extension of the min-norm control methodology of~\cite{6709752} for control-affine nonlinear systems. In Lemma~\ref{lemma_feasible}, the Lagrangian system-type structure of the spacecraft relative dynamics is extensively used to obtain the analytical solution~\eqref{min_norm_controller} of the quadratic optimization problem~\eqref{min_norm_controller_prob}, for the sake of its real-time implementation. Since there is no standard, systematic way to construct a Lyapunov function for a general nonlinear system~\cite{tutorial}, having at least one explicit form of a Lyapunov function as in Lemma~\ref{lemma_feasible} is useful for the practical application of our approach. See Sec.~\ref{sec_extension_lyap} for more details.
\begin{remark}
Note that the target learned policy $u_{\ell}$ could be designed to satisfy a given input constraint through the optimization and the appropriate choice of the network activation function. However, for general cases, $u^*_{\ell}$ of~\eqref{min_norm_controller_prob} could violate the given constraint even though we minimize its deviation from $u_{\ell}$. We could still efficiently handle (i) linear input constraints, \ie{}, 
$u_{\min} \leq u_j \leq u_{\max}$ for each $j$, which leads to a quadratic program if used in~\eqref{min_norm_controller_prob}, or (ii) norm-based input upper bounds (\ie{}, $\|u\| \leq u_{\max}$), which preserves convexity of~\eqref{min_norm_controller_prob}. Combining this with more general input constraints is left as a future work.
\end{remark}
\subsection{Assumptions for Robustness and Stability}
\label{sec_assumption}
Before proceeding to the next section on proving the robustness and stability properties of the SN-DNN min-norm control of~\eqref{min_norm_controller} of Lemma~\ref{lemma_feasible}, let us make a few assumptions with the notations given in Table~\ref{tab:notations_assump}, the first of which is that the spacecraft has access to an on-board navigation scheme that satisfies the following conditions. Due to the system nonlinearity, we can only assume the following only \textit{locally} (see~\eqref{event_local_est}), which is one of the reasons why we need to construct a non-negative function with a supermartingale property later in Theorem~\ref{Thm:control_robustness}.
\begin{table*}[htbp]
\caption{Notations in Sec.~\ref{sec_assumption}. \label{tab:notations_assump}}
\vspace{-1em}
\footnotesize
\begin{center}
\renewcommand{\arraystretch}{1.4}
\begin{tabular}{ m{2.5cm} m{12cm} }
\hline
\hline
Notation & Description \\
\hline
$\mathcal{C}_{\mathrm{iso}}(r),~\mathcal{C}_{\mathrm{sc}}(r)$ & Tubes of radius $r\in\mathbb{R}_{>0}$ centered around the desired trajectories $x_d$ and $\textit{\oe}_d$ given in~\eqref{target_trajectory}, \ie{}, $\mathcal{C}_{\mathrm{iso}}(r) = \bigcup_{t\in[t_s,t_f]}\{\xi\in\mathbb{R}^n|\|\xi-\textit{\oe}_d(t)\| < r\} \subset \mathbb{R}^n$ and $\mathcal{C}_{\mathrm{sc}}(r) = \bigcup_{t\in[t_s,t_f]}\{\xi\in\mathbb{R}^n|\|\xi-x_d(t)\| < r\} \subset \mathbb{R}^n$ \\
$\mathcal{E}_{\mathrm{iso}},~\mathcal{E}_{\mathrm{sc}}$ & Subsets of $\mathbb{R}^n$ that have the ISO and S/C state estimation error vectors $\|\hat{\textit{\oe}}(t)-\textit{\oe}(t)\|$ and $\|\hat{x}(t)-x(t)\|$, respectively, where a given on-board navigation scheme is valid (\eg{}, region of attraction) \\
$\mathbb{E}_{Z_1}\left[\cdot\right]$ & Conditional expected value operator \st{} $\mathbb{E}\left[\left.\cdot\right|x(t_1)=x_1,\hat{x}(t_1)=\hat{x}_1,\textit{\oe}(t_1)=\textit{\oe}_1,\hat{\textit{\oe}}(t_1)=\hat{\textit{\oe}}_1\right]$ \\
$t_f$ & Given terminal time at the ISO encounter as in~\eqref{mpc_dynamics} \\
$t_s$ & Time when the S/C activates the SN-DNN min-norm control policy~\eqref{min_norm_controller} of Lemma~\ref{lemma_feasible} \\
$Z_1$ & Tuple of the true and estimated ISO and S/C state at time $t=t_1$, \ie{}, $Z_1 = (\textit{\oe}_1,\hat{\textit{\oe}}_1,x_1,\hat{x}_1)$ \\
$\varsigma^t(Z_1,t_1)$ & Expected estimation error upper bound at time $t=t_1$ given $Z_1$ at time $t=t_1$, which can be determined based on the choice of navigation techniques (see Remark~\ref{remark_nav}) \\
\hline
\hline
\end{tabular}
\end{center}
\vspace{-1em}
\end{table*}
\begin{assumption}
\label{assumption_navigation}
Let the probability of the error vectors remaining in $\mathcal{E}_{\mathrm{iso}}$ and $\mathcal{E}_{\mathrm{sc}}$ be bounded as follows for $\exists\varepsilon_{\mathrm{est}}\in\mathbb{R}_{\geq 0}$:
\begin{align}
    &\mathbb{P}\left[\bigcap_{t\in[0,t_f]}\left(\hat{\textit{\oe}}(t)-\textit{\oe}(t) \right)\in\mathcal{E}_{\mathrm{iso}}\cap \left(\hat{x}(t)-x(t)\right)\in\mathcal{E}_{\mathrm{sc}}\right] \\
    &\geq 1-\varepsilon_{\mathrm{est}}.
    \label{event_local_est}
\end{align}
We assume that the process noise and sensor noise in our navigation scheme are represented by the Weiner process, and that if the event of~\eqref{event_local_est} has occurred, then we have the following bound for any $t_1,t_2\in[0,t_f]$ \st{} $t_1\leq t_2$:
\begin{align}
    &\mathbb{E}_{Z_1}\left[\sqrt{\|\hat{\textit{\oe}}(t)-\textit{\oe}(t) \|^2+\|\hat{x}(t)-x(t)\|^2}\right] \leq \varsigma^t(Z_1,t_1),\\
    &\forall t\in[t_1,t_2].
    \label{navigation_exp_bound}
\end{align}
We also assume that given the $2$-norm estimation error satisfies $\sqrt{\|\hat{\textit{\oe}}(t)-\textit{\oe}(t) \|^2+\|\hat{x}(t)-x(t)\|^2} < c_e$ for $c_e\in\mathbb{R}_{\geq0}$ at time $t=t_s$, then it satisfies this bound for $\forall t\in[t_s,t_f]$ with probability at least $1-\varepsilon_{\mathrm{err}}$, where $\exists\varepsilon_{\mathrm{err}}\in\mathbb{R}_{\geq 0}$. 
\end{assumption}
If the extended Kalman filter or contraction theory-based estimator (see~\cite{6849943,Ref:contraction1,tutorial,Ref:Stochastic} and references therein) is used for navigation with disturbances expressed as the Gaussian white noise processes, then we have 
\begin{align}
    \varsigma^t(Z_1,t_1) = e^{-\beta(t-t_1)}\sqrt{\|\hat{\textit{\oe}}_1-\textit{\oe}_1 \|^2+\|\hat{x}_1-x_1\|^2}+c
\end{align}
where $Z_1 = (\textit{\oe}_1,\hat{\textit{\oe}}_1,x_1,\hat{x}_1)$ and $\beta$ and $c$ are some given positive constants (see Example~\ref{example_exponential}). The last statement of the boundedness is expected for navigation schemes resulting in a decreasing estimation error, and can be shown formally using Ville's maximal inequality for supermartingales~\cite[pp. 79-83]{sto_stability_book} with an appropriate Lyapunov-like function for navigation synthesis~\cite[pp. 79-83]{sto_stability_book} (see Lemma~\ref{lemma_supermartingale} for similar computation in control synthesis). Note that if $\mathcal{E}_{\mathrm{iso}}=\mathcal{E}_{\mathrm{sc}}=\mathbb{R}^n$, \ie{}, the bound~\eqref{navigation_exp_bound} holds globally, then we have $\varepsilon_{\mathrm{est}} = 0$. Let us further make the following assumption on the SN-DNN min-norm control policy~\eqref{min_norm_controller}.
\begin{assumption}
\label{assumption_control1}
We assume that $k$ of~\eqref{min_norm_controller_k} is locally Lipschitz in its first two arguments, \ie{}, $\exists r_{\mathrm{sc}},r_{\mathrm{iso}},L_k\in\mathbb{R}_{> 0}$ \st{}
\begin{align}
    &\|k(x_1,\textit{\oe}_1,t)-k(x_2,\textit{\oe}_2,t)\| \leq L_k\sqrt{\|\textit{\oe}_1-\textit{\oe}_2\|^2+\|x_1-x_2\|^2},\\
    &\forall \textit{\oe}_1,\textit{\oe}_2\in\mathcal{C}_{\mathrm{iso}}(r_{\mathrm{iso}}),~x_1,x_2\in\mathcal{C}_{\mathrm{sc}}(r_{\mathrm{sc}}),~t\in[t_s,t_f].
    \label{Lipschitzk}
\end{align}
We also assume that $r_{\mathrm{iso}}$ and $r_{\mathrm{sc}}$ are large enough to have $r_{\mathrm{iso}}-2c_e \geq 0$ and $r_{\mathrm{sc}}-2c_e \geq 0$ for $c_e$ of Assumption~\ref{assumption_navigation}, and that the set $\mathcal{C}_{\mathrm{iso}}(r_{\mathrm{iso}}-c_e)$ is forward invariant, \ie{},
\begin{align}
    \textit{\oe}(t_s)\in\mathcal{C}_{\mathrm{iso}}(r_{\mathrm{iso}}-c_e) \Rightarrow~& \varphi^{t-t_s}(\textit{\oe}(t_s))\in\mathcal{C}_{\mathrm{iso}}\left(r_{\mathrm{iso}}-c_e\right), \\
    &\forall t\in[t_s,t_f]
    \label{assump_deterministic_isoflow}
\end{align}
where $\varphi^{t-t_s}(\textit{\oe}(t_s))$ is the ISO state trajectory with $\varphi^{0}(\textit{\oe}(t_s))=\textit{\oe}(t_s)$ at $t=t_s$ as given in Table~\ref{tab_flow_recap} and defined in Definition~\ref{def_iso_flow}.
\end{assumption}
If $\mathscr{f}$ of~\eqref{fnB_pdotdynamics} is locally Lipschitz in $x$ and $\textit{\oe}$, $k$ of~\eqref{min_norm_controller_k} can be expressed as a composition of locally Lipschitz functions in $x$ and $\textit{\oe}$, which implies that the Lipschitz assumption~\eqref{Lipschitzk} always holds for finite $r_{\mathrm{iso}}$ and $r_{\mathrm{sc}}$ (see, \eg{}, Theorem~2 of~\cite{7040372} and the references therein for further explanation). Since the estimation error is expected to decrease in general, $c_e$ of Assumption~\ref{assumption_navigation} can be made smaller as $t_s$ gets larger, which renders the second condition of Assumption~\ref{assumption_control1} less strict.
\section{Neural-Rendezvous: Learning-based Robust Guidance and Control to Encounter ISOs}
\label{sec_control_proof}
This section finally presents Neural-Rendezvous, a deep learning-based terminal G\&C approach to autonomously encounter ISOs, thereby solving the problem~\ref{Problem1} of Sec.~\ref{sec_problem} that arise from the large state uncertainty and high-velocity challenges. It will be shown that the SN-DNN min-norm control~\eqref{min_norm_controller} of Lemma~\ref{lemma_feasible} verifies a formal exponential bound on expected spacecraft delivery error, which provides valuable information in determining whether we should use the SN-DNN terminal guidance policy or enhance it with the SN-DNN min-norm control, depending on the size of the state uncertainty.
\subsection{Robustness and Stability Guarantee}
The assumptions introduced in Sec.~\ref{sec_assumption} allow bounding the mean squared distance between the spacecraft relative position of~\eqref{original_dyn} controlled by~\eqref{min_norm_controller} and the desired position $p_d(t)$ given in~\eqref{target_trajectory}, even under the presence of the state uncertainty (see Fig.~\ref{fig_exp_sndnn_bound}). We remark that the additional notations in the following theorem are summarized in Table~\ref{tab:notations_thm2}, where the others are consistent with the ones in Table~\ref{tab:notations_assump}. Let us remark that the use of our incremental Lyapunov-like function, along with the supermartingale analysis, allows for easy handling of the aforementioned local assumptions in estimation and control.
\begin{table*}
\caption{Notations in Theorem~\ref{Thm:control_robustness}. \label{tab:notations_thm2}}
\vspace{-1em}
\footnotesize
\begin{center}
\renewcommand{\arraystretch}{1.4}
\begin{tabular}{ m{2.5cm} m{12cm} }
\hline
\hline
Notation & Description \\
\hline
$\mathcal{D}_{\mathrm{est}}$ & Set defined as $\mathcal{D}_{\mathrm{est}} = \{(\textit{\oe}_s,x_s)\in\mathbb{R}^n\times\mathbb{R}^n|\sqrt{\|\textit{\oe}_s-\hat{\textit{\oe}}_s\|^2+\|x_s-\hat{x}_s\|^2}\leq c_e$ \\
$L_k$ & Lipschitz constant of $k$ in Assumption~\eqref{assumption_navigation} \\
$m_{\mathrm{sc}}(t)$ & Spacecraft mass of~\eqref{actualfnB} satisfying $m_{\mathrm{sc}}(t)\in[m_{\mathrm{sc}}(t_s),m_{\mathrm{sc}}(t_f)]$ due to the Tsiolkovsky rocket equation \\
$\textit{\oe}_s,\hat{\textit{\oe}}_s,x_s,\hat{x}_s$ & True and estimated ISO and S/C state at time $t=t_s$, respectively  \\
$p_d(t)$ & Desired S/C relative position trajectory, \ie{}, $x_d(t) = [p_d(t)^{\top},\dot{p}_d(t)^{\top}]^{\top}$ for $x_d$ of~\eqref{target_trajectory} \\
$p_s,~\hat{p}_s$ & True and estimated S/C relative position at time $t=t_s$, \ie{}, $x_s = [p_s^{\top},\dot{p}_s^{\top}]^{\top}$ and $\hat{x}_s = [\hat{p}_s^{\top},\dot{\hat{p}}_s^{\top}]^{\top}$ \\
$\bar{R}_{\mathrm{iso}},~\bar{R}_{\mathrm{sc}}$ & Constants defined as $\bar{R}_{\mathrm{iso}}=r_{\mathrm{iso}}-2c_e$ and $\bar{R}_{\mathrm{sc}}=r_{\mathrm{sc}}-2c_e$ for $c_e$ of Assumption~\ref{assumption_navigation} and $r_{\mathrm{iso}}$ and $r_{\mathrm{sc}}$ of Assumption~\ref{assumption_control1} \\
$u^*_{\ell}$ & SN-DNN min-norm control policy~\eqref{min_norm_controller} of Lemma~\ref{lemma_feasible} \\
$v(x,t)$ & Non-negative function given as $v(x,t) = \sqrt{V(\dot{p},\varrho_d(p,t),t)} = \sqrt{m_{\mathrm{sc}}(t)}\|\Lambda(p-p_d(t))+\dot{p}-\dot{p}_d(t)\|$ for $V$ of~\eqref{def_lyapunov} \\
$Z_s$ & Tuple of the true and estimated ISO and S/C state at time $t=t_s$, \ie{}, $Z_s = (\textit{\oe}_s,\hat{\textit{\oe}}_s,x_s,\hat{x}_s)$ \\
$\alpha$ & Positive constant of~\eqref{Vstability} \\ 
$\underline{\lambda}$ & Minimum eigenvalue of $\Lambda$ defined in~\eqref{Vstability} \\
$\rho$ & Desired terminal S/C relative position of~\eqref{mpc_dynamics} \\
\hline
\hline
\end{tabular}
\end{center}
\vspace{-1em}
\end{table*}
\begin{theorem}
\label{Thm:control_robustness}
Suppose that Assumptions~\ref{assumption_navigation}~and~\ref{assumption_control1} hold, and that the spacecraft relative dynamics with respect to the ISO, given in~\eqref{original_dyn}, is controlled by $u=u^*_{\ell}$. If the estimated states at time $t=t_s$ satisfy $\hat{\textit{\oe}}_s \in \mathcal{C}_{\mathrm{iso}}(\bar{R}_{\mathrm{iso}})$ and $\hat{x}_s \in \mathcal{C}_{\mathrm{sc}}(\bar{R}_{\mathrm{sc}})$, and if the expected norm of the state tracking error is smooth in $t$, then the spacecraft delivery error is explicitly bounded as follows with probability at least $1-\varepsilon_{\mathrm{ctrl}}$:
\begin{align}
    &\mathbb{E}\left[\left.\|p(t_f)-\rho\|\right|\hat{\textit{\oe}}(t_s)=\hat{\textit{\oe}}_s\cap\hat{x}(t_s)=\hat{x}_s\right] \\
    &\leq \sup_{(\textit{\oe}_s,x_s)\in\mathcal{D}_{\mathrm{est}}}e^{-\underline{\lambda} (t_f-t_s)} \|p_s-p_d(t_s)\|+\text{$\tfrac{e^{-\underline{\lambda} t_f}L_k\int_{t_s}^{t_f}\varpi^{\tau}(Z_s,t_s)d\tau}{m_{\mathrm{sc}}(t_f)}$} \nonumber \\
    &+\begin{cases}\frac{e^{-\underline{\lambda} (t_f-t_s)}-e^{-\alpha (t_f-t_s)}}{\alpha-\underline{\lambda}}\frac{v(x_s,t_s)}{\sqrt{m_{\mathrm{sc}}(t_f)}} &\text{if }\alpha\neq\underline{\lambda} \\
    (t_f-t_s)e^{-\underline{\lambda} (t_f-t_s)}\frac{v(x_s,t_s)}{\sqrt{m_{\mathrm{sc}}(t_f)}}&\text{if }\alpha=\underline{\lambda}
    \end{cases}
    \label{position_error_bound_terminal}
\end{align}
where $\varepsilon_{\mathrm{ctrl}}\in\mathbb{R}_{\geq 0}$ is the probability to be given in~\eqref{probability_last}, $\varpi^t(Z_s,t_s)$ is a time-varying function defined as $\varpi^t(Z_s,t_s) = e^{(\underline{\lambda}-\alpha) t}\int_{t_s}^{t}e^{\alpha \tau}\allowbreak\varsigma^{\tau}(Z_s,t_s)d\tau$.
Note that~\eqref{position_error_bound_terminal} yields a bound on
\begin{align}
    \mathbb{P}\left[\left.\|p(t_f)-\rho\| \leq d\right|\hat{\textit{\oe}}(t_s)=\hat{\textit{\oe}}_s\cap\hat{x}(t_s)=\hat{x}_s\right]
\end{align}
as in Theorem~2.5 of~\cite{tutorial}, where $d\in\mathbb{R}_{\geq0}$ is any given distance of interest, using Markov's inequality.
\end{theorem}
\begin{proof}
The proof is structured as follows.
\begin{enumerate}
    \item The weak infinitesimal operator $\mathscr{A}$~\cite[p. 9]{sto_stability_book} is applied to analyze the time evolution of the system using Lemma~\ref{lemma_feasible} under Assumption~\ref{assumption_control1}.
    \item A finite-time exponential bound on the expected position error is computed along with its probability under Assumption~\ref{assumption_navigation}.
    \item The bound in Step 2 is proved to hold for $\forall t\in[t_s,t_f]$ by constructing a non-negative supermartingale function out of the Lyapunov-like function, leading to the bound~\eqref{position_error_bound_terminal}.
\end{enumerate}
\vspace{1em}
\begin{proof}[\rm \textbf{Step 1: Analyze the system's time evolution}]\renewcommand{\qedsymbol}{}
The dynamical system \eqref{pdotdynamics} controlled by~\eqref{min_norm_controller} can be rewritten as
\begin{align}
    \label{stochastic_decomp}
    \ddot{p} &= \mathscr{f}(x,\textit{\oe},t)+\mathscr{B}(x,\textit{\oe},t)u^*_{\ell}(x,\textit{\oe},t;\theta_{\mathrm{nn}})+\mathscr{B}(x,\hat{\textit{\oe}},t)\tilde{u}
\end{align}
where $\tilde{u} = u^*_{\ell}(\hat{x},\hat{\textit{\oe}},t;\theta_{\mathrm{nn}})-u^*_{\ell}(x,\textit{\oe},t;\theta_{\mathrm{nn}})$. Since we are overloading the dot notation for the time derivative as the third term of~\eqref{stochastic_decomp} is subject to stochastic disturbance due to the state uncertainty of Assumption~\ref{assumption_navigation}, we consider the weak infinitesimal operator $\mathscr{A}$ given in~\cite[p. 9]{sto_stability_book}, instead of taking the time derivative, for analyzing the time evolution of the non-negative function $v$ defined as $v(x,t) = \sqrt{V(\dot{p},\varrho_d(p,t),t)} = \sqrt{m_{\mathrm{sc}}(t)}\|\Lambda(p-p_d(t))+\dot{p}-\dot{p}_d(t)\|$ for $V$ of~\eqref{def_lyapunov}. To this end, let us compute the following time increment of $v$ evaluated at the true state and time $(x,\textit{\oe},t)$:
\begin{align}
    \label{Vincrement_def}
    \Delta v &= v(x(t+\Delta t),t+\Delta t)-v(x,t) \\
    &= \frac{1}{2v(x,t)}\left(\frac{\partial V}{\partial \dot{p}}\ddot{p}(t)+\frac{\partial V}{\partial \varrho_d}\ddot{\varrho}_d(p(t),t)+\frac{\partial V}{\partial t}\right)\Delta t+\mathcal{O}\left(\Delta t^2\right)\nonumber
\end{align}
where $\Delta t\in\mathbb{R}_{\geq 0}$ and the arguments $(\dot{p}(t),\varrho_d(p,t),t)$ of the partial derivatives of $V$ are omitted for notational convenience. Dropping the argument $t$ for the state variables for simplicity, the dynamics decomposition~\eqref{stochastic_decomp} gives
\begin{align}
    &\frac{\partial V}{\partial \dot{p}}\ddot{p}+\frac{\partial V}{\partial \varrho_d}\ddot{\varrho}_d(p,t)+\frac{\partial V}{\partial t} \leq \frac{\partial V}{\partial \dot{p}}(\mathscr{f}(x,\textit{\oe},t)+\mathscr{B}(x,\textit{\oe},t) \\
    &\times u^*(\hat{x},\hat{\textit{\oe}},t;\theta_{\mathrm{nn}}))+\frac{\partial V}{\partial \varrho_d}(\ddot{p}_d-\Lambda(\dot{p}-\dot{p}_d)) \\
    &\leq -2\alpha V(\dot{p},\varrho_d(p,t),t)+2\frac{\partial V}{\partial \dot{p}}\mathscr{B}(x,\textit{\oe},t)\tilde{u} \\
    &= -2\alpha v(x,t)^2+2v(x,t)\frac{\|\tilde{u}\|}{\sqrt{m_{\mathrm{sc}}(t)}}
    \label{Vdot}
\end{align}
where the first inequality follows from the fact that the spacecraft mass $m_{\mathrm{sc}}(t)$, described by the Tsiolkovsky rocket equation, is a decreasing function and thus $\partial V/\partial t \leq 0$, and the second inequality follows from the stability condition~\eqref{Vstability} evaluated at $(x,\textit{\oe},t)$, which is guaranteed to be feasible due to Lemma~\ref{lemma_feasible}. Since $u^*$ is assumed to be Lipschitz as in~\eqref{Lipschitzk} of Assumption~\ref{assumption_control1}, we get the following relation for any $x_s,\hat{x}_s\in\mathcal{C}_{\mathrm{sc}}(r_{\mathrm{sc}})$, $\textit{\oe}_s,\hat{\textit{\oe}}_s\in\mathcal{C}_{\mathrm{iso}}(r_{\mathrm{iso}})$, and $t_s\in[0,t_f]$, by substituting~\eqref{Vdot} into~\eqref{Vincrement_def}:
\begin{align}
    &\mathscr{A}v(x_s,t_s)=\lim_{\Delta t\downarrow 0}\frac{\mathbb{E}_{Z_s}\left[\Delta v\right]}{\Delta t} \\
    &\leq -\alpha v(x_s,t_s)+\frac{L_k}{\sqrt{m_{\mathrm{sc}}(t_f)}}\sqrt{\|\hat{\textit{\oe}}_s-\textit{\oe}_s \|^2+\|\hat{x}_s-x_s\|^2}
    \label{L_def}
\end{align}
where $Z_s = (x_s,\hat{x}_s,\textit{\oe}_s,\hat{\textit{\oe}}_s)$,
\begin{align}
    \mathbb{E}_{Z_s}\left[\cdot\right] = \mathbb{E}\left[\left.\cdot\right|x(t_s)=x_s,\hat{x}(t_s)=\hat{x}_s,\textit{\oe}(t_s)=\textit{\oe}_s,\hat{\textit{\oe}}(t_s)=\hat{\textit{\oe}}_s\right],\nonumber
\end{align}
and the relation $m_{\mathrm{sc}}(t) \geq m_{\mathrm{sc}}(t_f)$ by the Tsiolkovsky rocket equation is also used to obtain the inequality.
\end{proof}
\begin{proof}[\rm \textbf{Step 2: Compute the expected position error bound}]\renewcommand{\qedsymbol}{}
\sloppy Dynkin's formula~\cite[p. 10]{sto_stability_book} (along with the hierarchical and combinational properties of contraction in~\cite{Ref:contraction1},~Theorem~2.7 of~\cite{tutorial},~and~Sec.~III-D~of~\cite{Pham2009}) gives the following with probability at least $1-\varepsilon_{\mathrm{est}}$, due to the smoothness assumption of the expected norm of the state tracking error and~\eqref{navigation_exp_bound} of Assumption~\ref{assumption_navigation}:
\begin{align}
    \label{s_bound}
    &\mathbb{E}_{Z_s}\left[\|\Lambda(p(t)-p_d(t))+\dot{p}(t)-\dot{p}_d(t)\|\right] \\
    &\leq \tfrac{e^{-\alpha (t-t_s)}v(x_s,t_s)}{\sqrt{m_{\mathrm{sc}}(t_f)}}+\tfrac{e^{-\alpha t}L_k}{m_{\mathrm{sc}}(t_f)}\int_{t_s}^{t}e^{\alpha \tau}\varsigma^{\tau}(Z_s,t_s)d\tau,~\forall t\in[t_s,\bar{t}_e]~~~~~~\nonumber
\end{align}
where $\varsigma^{\tau}(Z_s,t_s)$ is as given in~\eqref{navigation_exp_bound} and $\bar{t}_e$ is defined as $\bar{t}_e = \min\{t_e,t_f\}$, with $t_e$ being the first exit time given by
\begin{align}
    \label{def_first_exit}
    &t_e=\inf_{t \geq t_s} t \\
    &\text{\st{} }(\textit{\oe}(t)\notin\mathscr{I})\cup(\hat{\textit{\oe}}(t)\notin\mathscr{I})\cup (x(t)\notin\mathscr{S})\cup(\hat{x}(t)\notin\mathscr{S})
\end{align}
\ie{}, the time that any of the states leave their respective set given that the event of~\eqref{event_local_est} in Assumption~\ref{assumption_navigation} has occurred, 
where $\mathscr{I} = \mathcal{C}_{\mathrm{iso}}(r_{\mathrm{iso}})$ and $\mathscr{S} = \mathcal{C}_{\mathrm{sc}}(r_{\mathrm{sc}})$ just for notational simplicity. Since we further have that $\mathscr{A}\|p-p_d\| = \frac{d}{dt}\|p-p_d\| \leq \|\tilde{\varrho}\|-\underline{\lambda}(p-p_d)$ for $\tilde{\varrho} = \Lambda(p-p_d)+\dot{p}-\dot{p}_d$ due to the hierarchical structure of $\tilde{\varrho}$, utilizing Dynkin's formula one more time along with the smoothness assumption of the expected norm of the state tracking error, and then substituting~\eqref{s_bound} result in
\begin{align}
    &\mathbb{E}_{Z_s}\left[\|p(t)-p_d(t)\|\right] \leq e^{-\underline{\lambda} (t-t_s)}\|p_s-p_d(0)\| \\
    &+e^{-\underline{\lambda}t}\int_{t_s}^t\tfrac{e^{(\underline{\lambda}-\alpha)\tau+\alpha t_s}v(x_s,t_s)}{\sqrt{m_{\mathrm{sc}}(t_f)}}+\tfrac{L_k\varpi^{\tau}(Z_s,t_s}{m_{\mathrm{sc}}(t_f)})d\tau,~\forall t\in[0,\bar{t}_e]
    \label{proof_last}
\end{align}
where $x_s =[p_s^{\top},\dot{p}_s^{\top}]^{\top}$.
\end{proof}
\begin{proof}[\rm \textbf{Step 3: Derive the exponential bound}]\renewcommand{\qedsymbol}{}
What is left to show is that we can achieve $t_e>t_f$ with a finite probability for the first exit time $t_e$ of~\eqref{def_first_exit}. For this purpose, we need the following lemma.
\begin{lemma}
\label{lemma_supermartingale}
Suppose that the events of Assumption~\ref{assumption_navigation} has occurred, \ie{}, the event of \eqref{event_local_est} has occurred and the $2$-norm estimation error satisfies $\sqrt{\|\hat{\textit{\oe}}(t)-\textit{\oe}(t) \|^2+\|\hat{x}(t)-x(t)\|^2} < c_e$ for $\forall t\in[t_s,t_f]$. If Assumption~\ref{assumption_control1} holds and if we have
\begin{align}
    \label{AV_appendix}
    \mathscr{A}v(x_s,t_s) &\leq -\alpha v(x_s,t_s)+\tfrac{L_k}{\sqrt{m_{\mathrm{sc}}(t_f)}}\sqrt{\|\hat{\textit{\oe}}_s-\textit{\oe}_s\|^2+\|\hat{x}_s-x_s\|^2}
\end{align}
for any $x_s,\hat{x}_s\in\mathcal{C}_{\mathrm{sc}}(r_{\mathrm{sc}})$, $\textit{\oe}_s,\hat{\textit{\oe}}_s\in\mathcal{C}_{\mathrm{iso}}(r_{\mathrm{iso}})$, and $t_s\in[0,t_f]$ as in~\eqref{L_def}, then we get the following probabilistic bound for $\exists\varepsilon_{\mathrm{exit}}\in\mathbb{R}_{\geq 0}$:
\begin{align}
    &\mathbb{P}_{Z_s}\left[\bigcap_{t\in[t_s,t_f]}x(t)\in\mathcal{C}_{\mathrm{sc}}(\bar{r}_{\mathrm{sc}})\right] \geq 1-\varepsilon_{\mathrm{exit}} \\
    &= \begin{cases}
    \left(1-\frac{E_s}{\bar{v}}\right)e^{-\frac{H(t_f)}{\bar{v}}} &\text{if }\bar{v} \geq \frac{\sup_{t\in[t_s,t_f]}h(t)}{\bar{\alpha}} \\
    1-\frac{\bar{\alpha} E_s+(e^{\bar{H}(t_f)}-1)\sup_{t\in[t_s,t_f]}h(t)}{\bar{\alpha} \bar{v}e^{\bar{H}(t_f)}} &\text{otherwise}
    \end{cases}
    \label{X_supermartingale}
\end{align}
where $Z_s = (\textit{\oe}_s,\hat{\textit{\oe}}_s,x_s,\hat{x}_s)$ denotes the states at time $t=t_s$, which satisfy $\textit{\oe}_s \in \mathcal{C}_{\mathrm{iso}}(\bar{r}_{\mathrm{iso}})$, $\hat{\textit{\oe}}_s \in \mathcal{C}_{\mathrm{iso}}(\bar{R}_{\mathrm{iso}})$, $x_s \in \mathcal{C}_{\mathrm{sc}}(\bar{r}_{\mathrm{sc}})$, and $\hat{x}_s \in \mathcal{C}_{\mathrm{sc}}(\bar{R}_{\mathrm{sc}})$ for $\bar{r}_{\mathrm{iso}}=r_{\mathrm{iso}}-c_e \geq 0$, $\bar{R}_{\mathrm{iso}}=r_{\mathrm{iso}}-2c_e \geq 0$, $\bar{r}_{\mathrm{sc}} = r_{\mathrm{sc}}-c_e \geq 0$, and $\bar{R}_{\mathrm{iso}}=r_{\mathrm{sc}}-2c_e \geq 0$, $E_s=v(x_s,t_s)+\delta_p\|p_s-p_d(t_s)\|$, $H(t) = \int_{t_s}^{t}h(\tau)d\tau$, $\bar{H}(t) = 2H(t)\bar{\alpha}/\sup_{t\in[t_s,t_f]}h(t)$, and $h(t) = L_k\varsigma^t(Z_s,t_s)/\sqrt{m_{\mathrm{sc}}(t_f)}$. The suitable choices of $\bar{v},\bar{\alpha},\delta_p\in\mathbb{R}_{> 0}$ are to be defined in the proof.
\end{lemma}
\begin{proof}[Proof of Lemma~\ref{lemma_supermartingale}]
See Appendix.
\end{proof}
Now, let us select $c_e$ of Assumption~\ref{assumption_navigation} as $c_e = k_e\mathbb{E}\left[\varsigma^{t_s}(Z_0,0)\right]$, where $Z_0$ is the tuple of the true and estimated ISO and spacecraft state at time $t=0$ and $k_e\in\mathbb{R}_{>0}$. Using Markov's inequality along with the bounds~\eqref{event_local_est}~and~\eqref{navigation_exp_bound} of Assumption~\ref{assumption_navigation}, we get the following probability bound:
\begin{align}
    &\mathbb{P}\left[\sqrt{\|\hat{\textit{\oe}}_s-\textit{\oe}_s\|^2+\|\hat{x}_s-x_s\|^2} < k_e\mathbb{E}\left[\varsigma^{t_s}(Z_0,0)\right]\right] \\
    &\geq 1-\varepsilon_{\mathrm{est}}-\frac{1}{k_{e}}.
    \label{markov_err_prob}
\end{align}
Since we have $\hat{\textit{\oe}}_s\in\mathcal{C}_{\mathrm{iso}}(\bar{R}_{\mathrm{iso}})$ and $\hat{x}_s\in\mathcal{C}_{\mathrm{sc}}(\bar{R}_{\mathrm{sc}})$ by the theorem assumption, the occurrence of the event in~\eqref{markov_err_prob} implies $\textit{\oe}_s\in\mathcal{C}_{\mathrm{iso}}(\bar{r}_{\mathrm{iso}})$ and $x_s\in\mathcal{C}_{\mathrm{sc}}(\bar{r}_{\mathrm{sc}})$ for $\bar{r}_{\mathrm{iso}}=r_{\mathrm{iso}}-k_e\mathbb{E}\left[\varsigma^{t_s}(Z_0,0)\right] \geq 0$, $\bar{r}_{\mathrm{sc}}=r_{\mathrm{sc}}-k_e\mathbb{E}\left[\varsigma^{t_s}(Z_0,0)\right] \geq 0$, and then $\textit{\oe}(t)\in\mathcal{C}_{\mathrm{iso}}(\bar{r}_{\mathrm{iso}}),~\forall t\in[t_s,t_f]$ due to the forward invariance condition~\eqref{assump_deterministic_isoflow} in Assumption~\ref{assumption_control1}. Therefore, using $\varepsilon_{\mathrm{err}}$ of Assumption~\ref{assumption_navigation} along with the result of Lemma~\ref{lemma_supermartingale}, we have $\exists\varepsilon_{\mathrm{ctrl}}\in\mathbb{R}_{\geq0}$ \st{}
\begin{align}
    &\mathbb{P}\left[\left.\mathcal{A}\right|\hat{\textit{\oe}}(t_s)=\hat{\textit{\oe}}_s\cap\hat{x}(t_s)=\hat{x}_s\right] \geq 1-\varepsilon_{\mathrm{ctrl}} \\
    &= 1-\varepsilon_{\mathrm{exit}}-\varepsilon_{\mathrm{est}}-\varepsilon_{\mathrm{err}}-\frac{1}{k_e}
    \label{probability_last}
\end{align}
where $\mathcal{A} = \bigcap_{t\in[t_s,t_f]}\textit{\oe}(t)\in\mathcal{C}_{\mathrm{iso}}(\bar{r}_{\mathrm{iso}})\cap\hat{\textit{\oe}}(t)\in\mathcal{C}_{\mathrm{iso}}(r_{\mathrm{iso}})\cap x(t)\in\mathcal{C}_{\mathrm{sc}}(\bar{r}_{\mathrm{sc}})\cap\hat{x}(t)\in\mathcal{C}_{\mathrm{sc}}(r_{\mathrm{sc}})\cap\mathcal{E}$ with $\mathcal{E}$ being the event of~\eqref{event_local_est} in Assumption~\ref{assumption_navigation}. The desired relation~\eqref{position_error_bound_terminal} follows by evaluating the first term of the integral in~\eqref{proof_last}, and by observing that if $\mathcal{A}$ of~\eqref{probability_last} occurs, we have $t_e>t_f$ and $\sqrt{\|\hat{\textit{\oe}}_s-\textit{\oe}_s\|^2+\|\hat{x}_s-x_s\|^2} < k_e\mathbb{E}\left[\varsigma^{t_s}(Z_0,0)\right]$ due to~\eqref{markov_err_prob}.
\end{proof}

Note that if $\mathcal{E}_{\mathrm{iso}}=\mathcal{E}_{\mathrm{sc}}=\mathbb{R}^n$ in Assumption~\ref{assumption_navigation} and $k$ is globally Lipschitz in Assumption~\ref{assumption_control1}, the estimation bound~\eqref{navigation_exp_bound} and the Lipschitz bound~\eqref{Lipschitzk} always hold without the second condition of Assumption~\ref{assumption_navigation}. This indicates that \eqref{position_error_bound_terminal} holds as long as $(\textit{\oe}_s,x_s)\in\mathcal{D}_{\mathrm{est}}$ for given $\hat{\textit{\oe}}_s,\hat{x}_s \in \mathbb{R}^n$, which occurs with probability at least $1-k_e^{-1}$ due to~\eqref{markov_err_prob}.
\end{proof}
\begin{figure}[htbp]
    \centering
    \includegraphics[width=83mm]{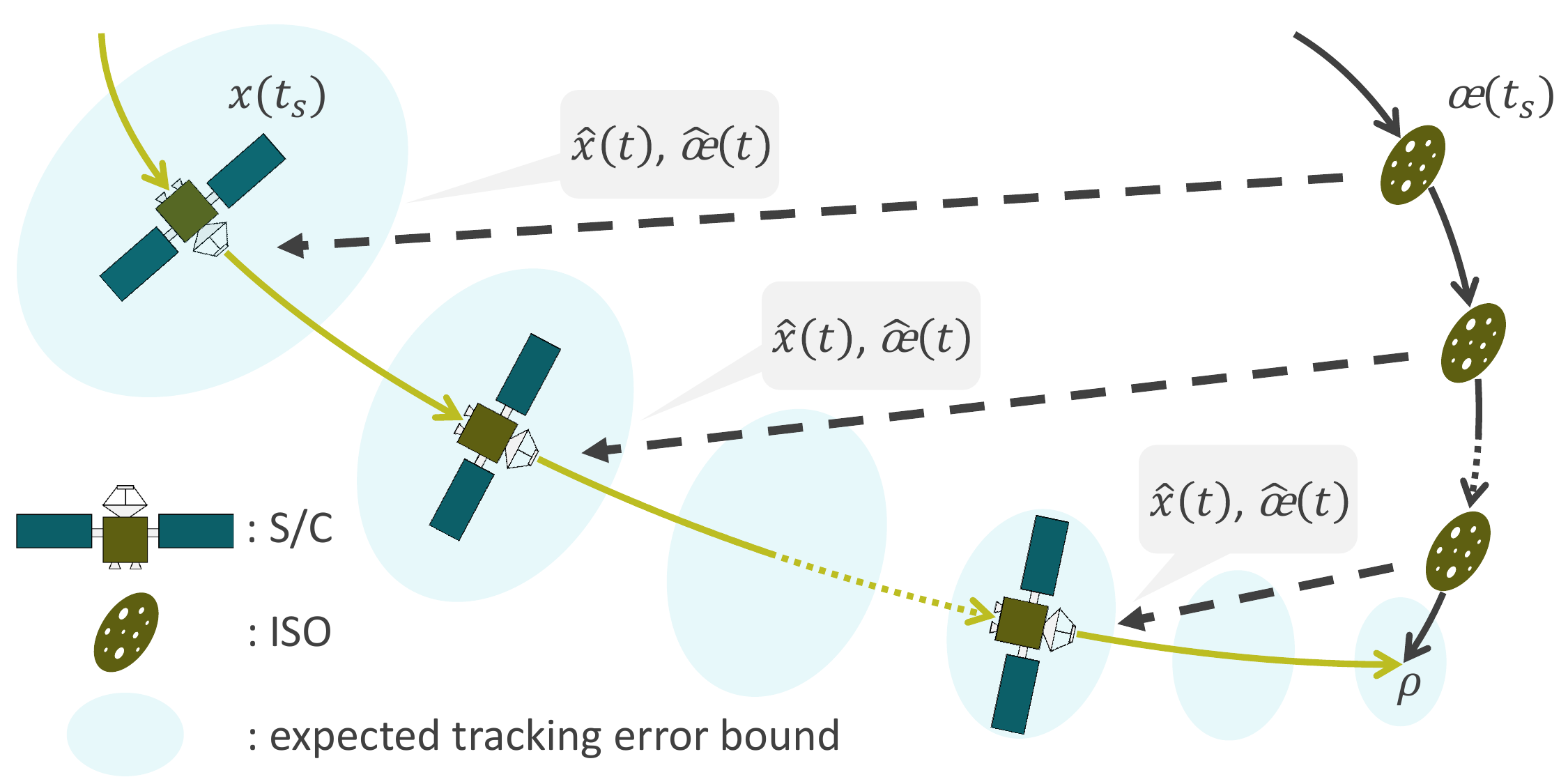}
    \caption{Illustration of the expected position tracking error bound~(\ref{position_error_bound_terminal}) of Theorem~\ref{Thm:control_robustness}, where $\bm{\varsigma^t(Z_s,t_s)}$ is assumed to be a non-increasing function in time.}
    \label{fig_minnorm_control}
\end{figure}
As derived in Theorem~\ref{Thm:control_robustness}, the SN-DNN min-norm control policy~\eqref{min_norm_controller} of Lemma~\ref{lemma_feasible} enhances the SN-DNN terminal guidance policy of Lemma~\ref{Lemma:SNDNNlearning} by providing the explicit spacecraft delivery error bound~\eqref{position_error_bound_terminal}, which holds even under the presence of the state uncertainty. This bound is valuable in modulating the learning and control parameters to achieve a verifiable performance guarantee consistent with a mission-specific performance requirement as to be seen in Sec.~\ref{sec_mpc_and_min_norm}.

As illustrated in Fig.~\ref{fig_minnorm_control}, the expected state tracking error bound~\eqref{position_error_bound_terminal} of Theorem~\ref{Thm:control_robustness} decreases exponentially in time if $\varsigma^t(Z_s,t_s)$ is a non-increasing function in $t$. The following example shows how we compute the bound~\eqref{position_error_bound_terminal} in practice.
\begin{example}
\label{example_exponential}
Suppose that the estimation error is upper-bounded by a function that exponentially decreases in time, \ie{}, we have $\varsigma^t(Z_s,t_s) = e^{-\beta (t-t_s)}\sqrt{\|\hat{\textit{\oe}}_s-\textit{\oe}_s \|^2+\|\hat{x}_s-x_s\|^2}+c$ for~\eqref{navigation_exp_bound} in Assumption~\ref{assumption_navigation}, where $c\in\mathbb{R}_{\geq 0}$ and $\beta\in\mathbb{R}_{> 0}$. Assuming that $\alpha\neq\beta$, $\beta\neq\underline{\lambda}$, and $\underline{\lambda}\neq\alpha$ for simplicity, we get
\begin{align}
    &\text{ {\small RHS~of~\eqref{position_error_bound_terminal} $\leq$} {\tiny$e^{-\underline{\lambda} \bar{t}_f}(\|\hat{p}_s-p_d(t_s)\|+c_e)+\tfrac{e^{-\underline{\lambda} \bar{t}_f}-e^{-\alpha \bar{t}_f}}{\alpha-\underline{\lambda}}\tfrac{(\bar{\lambda}+1)(\|\hat{x}_s-x_d(t_s)\|+c_e)}{\sqrt{m_{\mathrm{sc}}(t_f)}}$}} \nonumber \\
    &\text{{\tiny$+\tfrac{L_k}{m_{\mathrm{sc}}(t_f)}\left(\tfrac{c_e}{\alpha-\beta}\left(\tfrac{e^{-\underline{\lambda}\bar{t}_f}-e^{-\beta\bar{t}_f}}{\beta-\underline{\lambda}}-\tfrac{e^{-\underline{\lambda}\bar{t}_f}-e^{-\alpha\bar{t}_f}}{\alpha-\underline{\lambda}}\right)+\tfrac{c}{\alpha}\left(\tfrac{1-e^{-\underline{\lambda}\bar{t}_f}}{\underline{\lambda}}-\tfrac{e^{-\underline{\lambda}\bar{t}_f}-e^{-\alpha\bar{t}_f}}{\alpha-\underline{\lambda}}\right)\right) $}}~~~~~~~~~\label{ex_expestimation}
\end{align}
where $\bar{t}_f = t_f-t_s$. Note that the bound~\eqref{ex_expestimation} can be computed explicitly for given $\hat{x}_s$ and $\hat{\textit{\oe}}_s$ at time $t=t_s$ as long as $\hat{\textit{\oe}}_s \in \mathcal{C}_{\mathrm{iso}}(\bar{R}_{\mathrm{iso}})$ and $\hat{x}_s \in \mathcal{C}_{\mathrm{sc}}(\bar{R}_{\mathrm{sc}})$. Its dominant term in~\eqref{ex_expestimation} for large $\bar{t}_f$ is ${L_kc}/({m_{\mathrm{sc}}(t_f)\alpha\underline{\lambda}})$ due to
\begin{align}
    \lim_{t_f\to\infty}\text{RHS~of~\eqref{ex_expestimation}} = \frac{L_kc}{m_{\mathrm{sc}}(t_f)\alpha\underline{\lambda}}.
\end{align}
\end{example}
\begin{remark}
Even if the smoothness assumption of Theorem~\ref{Thm:control_robustness} is violated, a result analogous to~\eqref{position_error_bound_terminal} can still be achieved by bounding the function $\zeta^t$ by a piecewise constant function. This enables the application of a modified version of Gronwall's lemma~\cite{gronwall} along with Dynkin's formula~\cite[p. 10]{sto_stability_book}.
\end{remark}

The bound~\ref{position_error_bound_terminal} of Theorem~\ref{Thm:control_robustness} can be used as a tool to determine whether we should use the SN-DNN terminal guidance policy of Lemma~\ref{Lemma:SNDNNlearning} or the SN-DNN min-norm control policy~\eqref{min_norm_controller} of Lemma~\ref{lemma_feasible}, based on the trade-off between them as to be seen in the following.
\begin{figure*}[htbp]
    \centering
    \includegraphics[width=125mm]{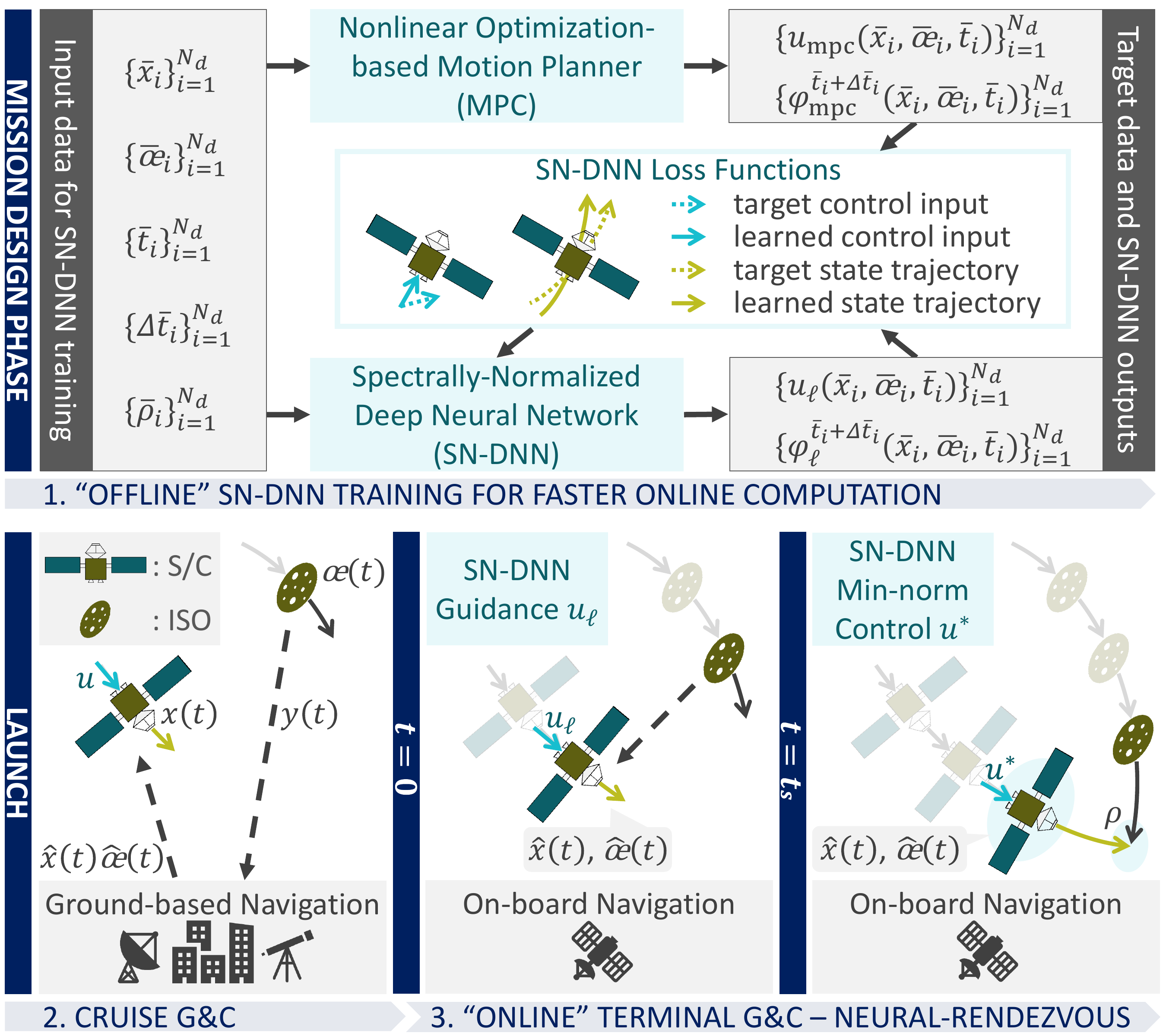}
    \caption{Mission timeline with the notations given in Fig.~\ref{fig_nr_outline}, Lemma~\ref{Lemma:SNDNNlearning}, and Theorem~\ref{Thm:control_robustness}. Note that the SN-DNN is to be trained offline.}
    \label{fig_nr_diagram}
\end{figure*}
\begin{algorithm}[htbp]
\footnotesize
\SetKwInOut{Input}{Inputs}\SetKwInOut{Output}{Outputs}
\Input{$u_{\ell}$ of Lemma~\ref{Lemma:SNDNNlearning} and $u^*_{\ell}$ of Lemma~\ref{lemma_feasible}}
\Output{Control input $u$ of~\eqref{original_dyn} for $t\in[0,t_f]$}
\BlankLine
$t_0 \gets \textit{current time}$ \\
$\Delta t \gets \textit{control time interval}$ \\
$t_k,~t_d,~t_{\mathrm{int}} \gets \textit{current time}-t_0$ \\
$\textit{flagA},~\textit{flagB} \gets 0$ \\
\While{$t_k < t_f$}{
    Obtain $\hat{\textit{\oe}}(t_k)$ and $\hat{x}(t_k)$ using navigation technique \\
    \eIf{$\textit{flagA} = 0$}
        {$u \gets u_{\ell}(\hat{x}(t_k),\hat{\textit{\oe}}(t_k),t_k)$}
        {\eIf{$\textit{flagB} = 1$}
            {$u \gets u^*_{\ell}(\hat{x}(t_k),\hat{\textit{\oe}}(t_k),t_k)$}
            {Compute~$\textit{RHS~of~\eqref{position_error_bound_terminal}}$ in Theorem~\ref{Thm:control_robustness} with $t_s = t_k$ \\
            \eIf{$\textit{RHS~of~\eqref{position_error_bound_terminal}} > threshold$}
                {$u \gets u_{\ell}(\hat{x}(t_k),\hat{\textit{\oe}}(t_k),t_k)$}
                {$u \gets u^*_{\ell}(\hat{x}(t_k),\hat{\textit{\oe}}(t_k),t_k)$ \\
                $\textit{flagB} \gets 1$
            }
        }
    }
    Apply $u$ to~\eqref{original_dyn} for $t\in[t_k,t_k+\Delta t]$ \\
    \While{$\textit{current time}-t_0 < t_k+\Delta t$}
        {\If{$\textit{flagB} = 0$}
            {Integrate dynamics to get $x_d$ and $\textit{\oe}_d$ of~\eqref{target_trajectory} form $t_{\mathrm{int}}$
        }
    }
    \eIf{$\textit{integration is complete}$}
        {Update $x_d$ and $\textit{\oe}_d$ \\
        $t_d,~t_{\mathrm{int}} \gets t_k+\Delta t$ \\
        $\textit{flagA} \gets 1$}
        {$t_{\mathrm{int}} \gets \textit{time when S/C stopped integration}$
    }
    $t_k \gets t_k+\Delta t$
}
\caption{\footnotesize Neural-Rendezvous}
\label{AlgISO}
\end{algorithm}
\subsection{Neural-Rendezvous}
\label{sec_mpc_and_min_norm}
We summarize the pros and cons of the aforementioned terminal G\&C techniques.
\setlength{\leftmargini}{17pt}     
\begin{itemize}
	\setlength{\itemsep}{1pt}      
	\setlength{\parskip}{0pt}     
    \item The SN-DNN terminal guidance policy of Lemma~\ref{Lemma:SNDNNlearning} can be implemented in real time and possesses an optimality gap as in~\eqref{sn_learning_error}, resulting in the near-optimal guarantee in terms of dynamic regret as discussed below~\eqref{mpc_control}~and~Lemma~\ref{Lemma:SNDNNlearning}. However, obtaining any quantitative bound on the spacecraft delivery error with respect to the desired relative position, to either flyby or impact the ISO, is difficult in general~\cite{tutorial}.
    \item In contrast, the SN-DNN min-norm control policy~\eqref{min_norm_controller} of Lemma~\ref{lemma_feasible}, which solves the problem~\ref{Problem1} of Sec.~\ref{sec_problem}, can also be implemented in real time and provides an explicit upper-bound on the spacecraft delivery error that decreases in time as proven in Theorem~\eqref{Thm:control_robustness}. However, the desired trajectory it tracks is subject to the large state uncertainty initially for small $t_s$, resulting in a large optimality gap.
\end{itemize}

Based on these observations, it is ideal to utilize the SN-DNN terminal guidance policy without any feedback control initially for large state uncertainty, to avoid having a large optimality gap, then activate the SN-DNN min-norm control once the verifiable spacecraft delivery error of~\eqref{position_error_bound_terminal} of Theorem~\eqref{Thm:control_robustness} becomes smaller than a mission-specific threshold value for the desired trajectory~\eqref{target_trajectory}, which is to be updated at $t_d\in[0,t_f]$ along the way as discussed in Remark~\ref{remark_ts}. The pseudo-code for the proposed learning-based approach for encountering the ISO is given in Algorithm~\ref{AlgISO}, and its mission timeline is shown in Fig.~\ref{fig_nr_diagram}.

Finally, let us again clarify the role of offline learning in Neural-Rendezvous, with the size of the neural network selected to be small enough for real-time implementation. It is to deal with the limited computational resources of the spacecraft in solving any nonlinear optimization with a non-negligible online computational load, including the one of MPC with receding horizons. This also enables utilizing Neural-Rendezvous in more challenging and highly nonlinear G\&C scenarios as to be seen in Sec.~\ref{sec_experiment}, where solving even a single optimization online could become unrealistic.
\begin{remark}
\label{remark_discretization}
We use $t_{\mathrm{int}}$ in Algorithm~\ref{AlgISO} to account for the fact that computing $x_d$ could take more than $\Delta t$ for small $\Delta t$ and $t_k$ as pointed out in~Remark~\ref{remark_ts}, and thus the spacecraft is sometimes required to compute it over multiple time steps. It can be seen that the SN-DNN terminal guidance policy $u_{\ell}$ is also useful when the spacecraft does not possess $x_d$ yet, \ie{}, when $\textit{flagA} = 0$. Also, the impact of discretization introduced in Algorithm~\ref{AlgISO} will be demonstrated in Sec.~\ref{sec_simulation_result}, where the detailed discussion of its connection to continuous-time stochastic systems can be found in~\cite{mypaperTAC}.
\end{remark}
\subsection{Extensions}
\label{sec_extensions}
In this section, we present several extensions of the proposed G\&C techniques for encountering ISOs, which can be used to further improve certain aspects of their performance.
\subsubsection{General Lyapunov Functions and Other Types of Disturbances}
\label{sec_extension_lyap}
For general nonlinear systems, we can construct a feedback control policy similar to Neural-Rendezvous as long as there exists a Lyapunov function. It is worth emphasizing that our proposed feedback control of Lemma~\ref{lemma_feasible} is also categorized as a Lyapunov-based approach, and thus we can show that it is robust against not only the state uncertainty of~\eqref{original_dyn}, but also deterministic and stochastic disturbances resulting from \eg{}, process noise, control execution error, parametric uncertainty, and unknown parts of dynamics. The proofs can be found in~\cite{Ref:contraction1,tutorial}, which also partially summarizes existing optimal and numerically efficient ways to find an optimal Lyapunov function in general.
\subsubsection{Stochastic MPC with Terminal Chance Constraints}
\label{sec_smpc}
We could utilize the expectation bound~\eqref{position_error_bound_terminal} to formulate a terminal chance-constrained stochastic optimal control problem, with probabilistic guarantees on reaching the terminal set that defines the encounter specifications with the ISO. Unlike the formulation given in~\eqref{mpc_eq}, this approach can explicitly account for the stochastic disturbance resulting from the ISO state uncertainty, leading to a more sophisticated offline solution computed by, \eg{}, the generalized polynomial chaos-based sequential convex programming method~\cite{yash_tro}. 

\subsubsection{Multi-Agent Systems in Cluttered Environments}
The optimization formulation with chance constraints in Sec.~\ref{sec_smpc} is also useful in extending our proposed approach to a multi-agent setting with obstacles, where each spacecraft is required to achieve its mission objectives in a collision-free manner. It allows expressing stochastic guidance problems as deterministic counterparts~\cite{yash_tro} so we could exploit existing methods for designing distributed, robust, and safe control policies for deterministic multi-agent systems, which can be computed in real time~\cite{glas}. This idea will be partially explored in Sec.~\ref{sec_uav_swarm}.
\subsubsection{Online Learning}
The proposed learning-based algorithm is based solely on offline learning and online guarantees of robustness and stability, but there could be situations where the parametric or non-parametric uncertainty of underlying dynamical systems is too large to be treated robustly. As shown in~\cite{tutorial}, robust control techniques, including our proposed approach in this paper, can always be augmented with adaptive control techniques with formal stability and with static or dynamic regret bounds for online nonlinear control problems~\cite{doi:10.1126/scirobotics.abm6597}.
\section{Simulation}
\label{sec_simulation}
Our proposed framework, Neural-Rendezvous, is demonstrated using the data set that contains ISO candidates for possible exploration~\cite{declan_nav} to validate if it indeed solves Problem~\ref{Problem1} introduced earlier in Sec.~\ref{sec_problem}. {\color{caltechgreen}\href{https://pytorch.org/}{PyTorch}} is used for designing and training neural networks and NASA's Navigation and Ancillary Information Facility ({\color{caltechgreen}\href{https://naif.jpl.nasa.gov/naif/}{NAIF}}) is used to obtain relevant planetary data. A YouTube video that visualizes these simulation results can be found at {\color{caltechgreen}\href{https://youtu.be/AhDPE-R5GZ4}{https://youtu.be/AhDPE-R5GZ4}}.
\begin{remark}
The main scope of our paper is to develop a control theoretical approach to derive a general mathematical guarantee \textit{given} a learning method. Therefore, instead of comparing our method with a countless number of works with learning and control with limited mathematical understanding, our simulations focus on existing G\&C approaches with strong mathematical guarantees. This is consistent with the safety-centered nature of space missions at NASA JPL, where we are interested in G\&C algorithms that can be theoretically shown to meet strict operational requirements.
\end{remark}
\subsection{Simulation Setup}
All the G\&C frameworks in this section are implemented with the control time interval $1$ s unless specified, and their computational time is measured using the MacBook Pro laptop (\SI{2.2}{\giga\hertz}~Intel Core i7, \SI{16}{\giga\byte}~\SI{1600}{\mega\hertz} DDR3 RAM). The terminal time $t_f$ of~\eqref{mpc_dynamics} for terminal guidance is selected to be $t_f = 86400$~(\si{\second}), and the wet mass of the spacecraft at the beginning of terminal guidance is assumed to be \SI{150}{\kilo\gram}. Also, we consider the SN-DNN min-norm control~\eqref{min_norm_controller} of Theorem~\ref{Thm:control_robustness} designed with $\Lambda = 1.3\times 10^{-3}$ and $\alpha = 8.9\times 10^{-7}$. The maximum control input is assumed to be $u_{\max} = 3$~(\si{\newton}) in each direction with the total admissible delta-V ($2$-norm S/C velocity increase) being \SI{0.6}{\kilo\metre\per\second}.
\subsection{Dynamical System-Based SN-DNN Training}
This section delineates how we train the dynamical system-based SN-DNN of Sec.~\ref{sec_guidance} for constructing the leaning-based terminal guidance algorithm.
\subsubsection{State Uncertainty Assumption}
\label{sec_estimation_assumption}
For the sake of simplicity, we assume that the spacecraft has access to the estimated ISO and its relative state generated by the respective normal distribution $\mathcal{N}(\mu,\Sigma)$, with $\mu$ being the true state and $\Sigma$ being the navigation error covariance, where $\trace(\Sigma)$ exponentially decaying in $t$ as in Example~\ref{example_exponential}. In particular, we assume that the standard deviations of the ISO and spacecraft absolute along-track position, cross-track position, along-track velocity, and cross-track velocity, expressed in the ECLIPJ2000 frame as in JPL’s {\color{caltechgreen}\href{https://naif.jpl.nasa.gov/pub/naif/toolkit_docs/C/req/frames.html}{SPICE toolkit}}, are $10^4$~\si{\kilo\metre}, $10^2$~\si{\kilo\metre}, $10^{-2}$~\si{\kilo\metre\per\second}, and $10^{-2}$~\si{\kilo\metre\per\second} initially at time $t=0$ (s), and decays to $\mathcal{O}(10^{1})$~\si{\kilo\metre}, $\mathcal{O}(10^{0})$~\si{\kilo\metre}, $\mathcal{O}(10^{-4})$~\si{\kilo\metre\per\second}, and $\mathcal{O}(10^{-4})$~\si{\kilo\metre\per\second} finally at time $t=t_f=86400$~(\si{\second}) in the along-track and cross-track direction, respectively, which can be achieved by using, \eg{}, the extended Kalman filter with full state measurements and estimation gains given by $R = \mathrm{I}_{n\times n}$ and $Q = \mathrm{I}_{n\times n}\times10^{-10}$.
\begin{remark}
As discussed in Remark~\ref{remark_nav}, G\&C are the focus of our study and navigation is beyond our scope, we have simply assumed the ISO state measurement uncertainty given above partially following the discussion of~\cite{declan_nav}. The assumption can be easily modified accordingly to the state estimation schemes to be used in each aerospace and robotic problem of interest. See Sec.~\ref{sec_autonav} for more discussion.
\end{remark}
\subsubsection{Training Data Generation}
\label{sec_data_generation}
We generate $499$ candidate ISO and spacecraft ideal trajectories based on the ISO population analyzed in~\cite{declan_nav}, and utilized the first $399$ ISOs for training the SN-DNN and the other $100$ for testing its performance later in this section. We then obtain $10000$ time and ISO index pairs $(t_i,I_i)$ uniformly and randomly from $[0,t_f)\times[1,399]\cup\mathbb{N}$ and perturbed the ISO and spacecraft ideal state with the uncertainty given in Sec.~\ref{sec_estimation_assumption} to produce the training samples $(\bar{x}_{i},\bar{\textit{\oe}}_{i},\bar{t}_i)$ of~(\ref{node_empirical_loss}) for the control input loss (\ie{}, the first term of~\eqref{node_loss}). The training data samples for the state trajectory loss (\ie{}, the second term of~\eqref{node_loss}) are obtained in the same way using the pairs generated uniformly and randomly from $[0,3600]\times[1,399]\cup\mathbb{N}$, with $\Delta \bar{t}_i$ of~(\ref{node_empirical_loss}) fixed to $\Delta \bar{t}_i = 10$~(\si{\second}). The desired relative positions $\bar{\rho}_i$ in~\eqref{node_empirical_loss} are also sampled uniformly and randomly from the surface of a ball with radius~\SI{100}{\kilo\metre}.

The desired control inputs $u_{\mathrm{mpc}}(\bar{x}_{i},\bar{\textit{\oe}}_{i},\bar{t}_i,\bar{\rho}_i)$ and desired state trajectories $\varphi_{\mathrm{mpc}}^{\bar{t}_i+\Delta\bar{t}_i}(\bar{x}_{i},\bar{\textit{\oe}}_{i},\bar{t}_i,\bar{\rho}_i)$ of~(\ref{node_empirical_loss}) are then sampled by solving~\eqref{mpc_control} using the sequential convex programming approach and by numerically integrating~\eqref{target_integration} using the fourth-order Runge–Kutta method, respectively, where the terminal position error is treated as a constraint $\|p_{\xi}(t_f)-\rho\| = 0$ in~\eqref{mpc_dynamics}, the cost function of~\eqref{mpc_eq} is defined with $c_0=0$, $c_1=1$, and $P\left(u(t),\xi(t)\right) = \|u(t)\|^2$ as in Remark~\ref{Remark_mpc_prob}, and the control input constraint $u(t)\in\mathcal{U}(t) = \{u\in\mathbb{R}^m||u_i| \leq u_{\max}\}$ $u_i$ is used with $u_i$ being the $i$th element of $u$ and $u_{\max} = 3$~(\si{\newton}). Note that the problem is discretized with the time step \SI{1}{\second}, consistently with the control time interval.
\subsubsection{Training Data Normalization}
Instead of naively training the SN-DNN with the raw data generated in Sec.~\ref{sec_data_generation}, we transform the SN-DNN input data $(\bar{x}_{i},\bar{\textit{\oe}}_{i},\bar{t}_i,\bar{\rho}_i)$ as follows, thereby accelerating the speed of learning process and improving neural network generalization performance:
\begin{align}
    \label{transform_SNDNN}
    \textit{\footnotesize SN-DNN input} = \left(\bar{p}_{i}-\bar{\rho}_i,\tfrac{\bar{p}_{i}-\bar{\rho}_i}{t_f-\bar{t}_i}+\dot{\bar{p}}_{i},\bar{\rho}_i,t_f-\bar{t}_i,\bar{\omega}_{z,i},G(\bar{p}_{i},\bar{\textit{\oe}}_{i})\right)~~~~~~~~~
\end{align}
where $\bar{x}_{i} = [\bar{p}_{i}^{\top},\dot{\bar{p}}_{i}^{\top}]^{\top}$, $G(p,\textit{\oe})$ is given in~\eqref{actualfnB}, and $\bar{\omega}_{z,i}$ is the $i$th training sample of the orbital element $\omega_z$ given in~\cite{doi:10.2514/1.37261}, which is the dominant element of the matrix function $C(\textit{\oe})$ of~\eqref{actualfnB}. We further normalize the input~\eqref{transform_SNDNN} and output $u_{\ell}(x,\textit{\oe},t,\rho;\theta_{\mathrm{nn}})$ of the SN-DNN by dividing them by their maximum absolute values in their respective training data.
\subsubsection{SN-DNN Configuration and Training}
\begin{figure}[htbp]
    \centering
    \includegraphics[width=83mm]{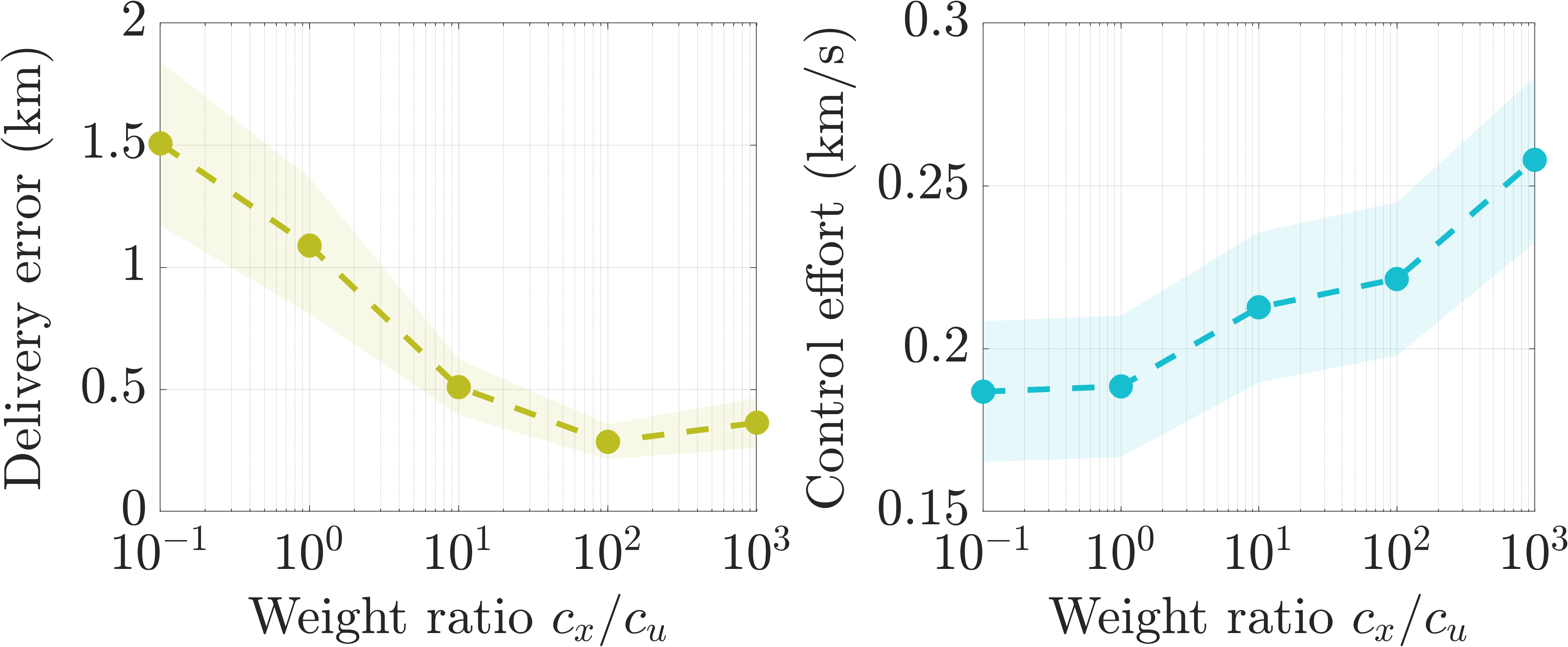}
    \caption{Control performances versus weight ratio of the SN-DNN loss function. The shaded area denotes the standard deviations (left: $\bm{{\pm2.5\times10^{-1}\sigma}}$ and right: $\bm{\pm5\times10^{-2}\sigma}$).}
    \label{fig_SNDNN}
\end{figure}

We select the number of hidden layers and neurons of the SN-DNN as $6$ and $64$, with the spectral normalization constant ($C_{nn}$ of Definition~6.2 in~\cite{tutorial}) being $C_{nn} = 25$. The activation function is selected to be $\tanh$, which is also used in the last layer not to violate the input constraint $u(t)\in\mathcal{U}(t) = \{u\in\mathbb{R}^m||u_i| \leq u_{\max}\}$ $u_i$ by design. Figure~\ref{fig_SNDNN} shows the terminal spacecraft position error (delivery error) and control effort (total delta-V) of the SN-DNN terminal guidance policy trained for $10000$ epochs using several different weights $c_u,c_x\in\mathbb{R}_{\geq 0}$ of the loss function~\eqref{node_loss} in Sec.~\ref{sec_guidance}, where its weight matrices are given as $C_x = c_x\diag(\mathrm{I}_{3\times3},10^7\times\mathrm{I}_{3\times3})$ and $C_u = c_u\mathrm{I}_{3\times3}$. The results are averaged over $50$ simulations for the ISOs in the test set without any state uncertainty. Consistently with the definition of~\eqref{node_loss}, this figure indicates that
\setlength{\leftmargini}{17pt}     
\begin{itemize}
	\setlength{\itemsep}{1pt}      
	\setlength{\parskip}{0pt}    
    \item as $c_x/c_u$ gets smaller, the loss function~\eqref{node_loss} penalizes the imitation loss of the control input more heavily than that of the state trajectory, and thus the spacecraft yields smaller control effort but with larger delivery error, and
    \item as $c_x/c_u$ gets larger, the loss function~\eqref{node_loss} penalizes the imitation loss of the state trajectory more heavily than that of the control input, and thus the spacecraft yields smaller delivery error but with larger control effort.
\end{itemize}
The weight ratio $c_x/c_u$ of the SN-DNN to be implemented in the next section is selected as $c_x/c_u=10^2$, which achieves the smallest delivery error with the smallest standard deviation of all the weight ratios in Fig.~\ref{fig_SNDNN}, while having the control effort smaller than the admissible delta-V of~\SI{0.6}{\kilo\metre\per\second}. Note that we could further optimize the ratio with the delta-V constraint based on this trade-off discussed above, but this is left as future work. The SN-DNN is then trained using SGD for $10000$ epochs with $10000$ training data points obtained as in Sec.~\ref{sec_data_generation}. 
\begin{remark}
The trend of Fig.~\ref{fig_SNDNN} is also due to the fact that the delivery is treated as a hard constraint by MPC in this paper as can be seen in~\eqref{mpc_eq}. In general, the best choice of the ratio $c_x/c_u$ will depend on how the motion planning is formulated and solved to sample training data (see, \eg{},~\cite{scpproof} for indirect approaches and~\cite{ddp_constraint} for direct approaches).
\end{remark}

\subsection{Neural-Rendezvous Performance}
\label{sec_simulation_result}
\begin{figure}[htbp]
    \centering
    \includegraphics[width=83mm]{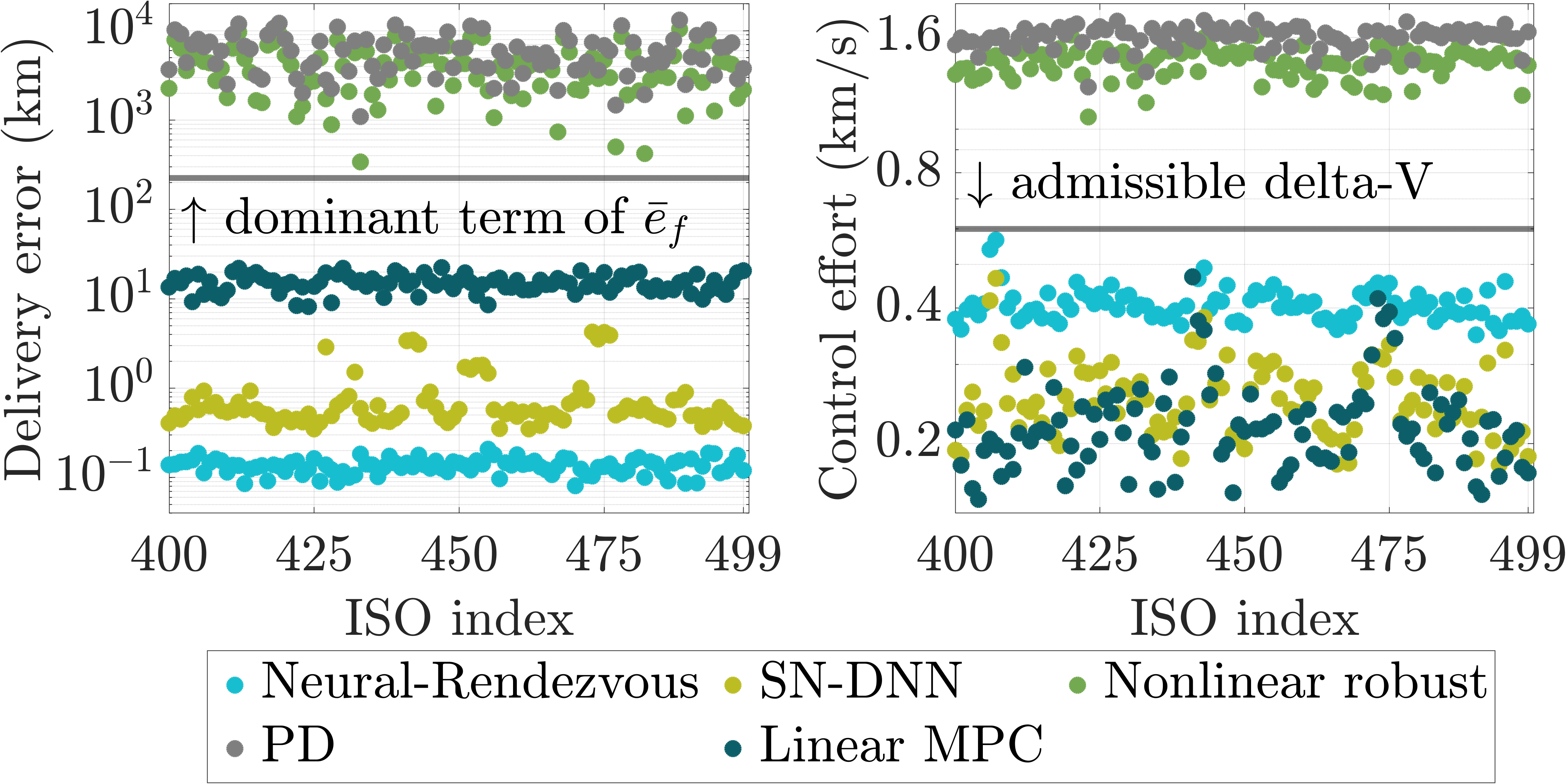}
    \caption{Control performances versus ISOs in the test set, where $\bm{\bar{e}_f}$ is the right-hand side of (\ref{position_error_bound_terminal}) (averaged over 10 simulations performed for each ISO).}
    \label{fig_xerr_dv}
\end{figure}

Figure~\ref{fig_xerr_dv} shows the spacecraft delivery error and control effort of Neural-Rendezvous of Algorithm~\ref{AlgISO}, SN-DNN terminal guidance of Sec.~\ref{sec_guidance}, PD guidance and control (\ie{}, PD control constructed to track a pre-computed and fixed desired trajectory, which is a JPL baseline), robust nonlinear tracking control of~\cite[pp. 397-402]{Ref_Slotine} (\ie{}, nonlinear sliding mode-type control for general Lagrangian systems to track a pre-computed and fixed desired trajectory), and MPC with linearized dynamics~\cite{mpc1}, where the SN-DNN min-norm control~\eqref{min_norm_controller} of Theorem~\ref{Thm:control_robustness} is activated at time $t = t_s = 26400$~(\si{\second}). It can be seen that Neural-Rendezvous achieves~$\leq 0.2$~\si{\kilo\metre} delivery error for \SI{99}{\percent} of the ISOs in the test set, even under the presence of the large ISO state uncertainty given in Sec.~\ref{sec_estimation_assumption}. Also, its error is indeed less than the dominant term of the expectation bound on the tracking error~\eqref{position_error_bound_terminal} of Theorem~\ref{Thm:control_robustness}, which is computed assuming the estimation error is upper-bounded by a function that exponentially decreases in time as in Example~\ref{example_exponential}. Furthermore, the SN-DNN terminal guidance can also achieve~$\leq 1$~\si{\kilo\metre} delivery error for \SI{86}{\percent} of the ISOs, and as expected from the optimality gap given in \eqref{sn_learning_error} of Lemma~\ref{Lemma:SNDNNlearning}, the control effort of Neural-Rendezvous is larger than that of the SN-DNN guidance and MPC with linearized dynamics, but it is still less than the admissible delta-V of \SI{0.6}{\kilo\metre\per\second} for all the ISOs in the test set.

\begin{figure}[htbp]
    \centering
    \includegraphics[width=83mm]{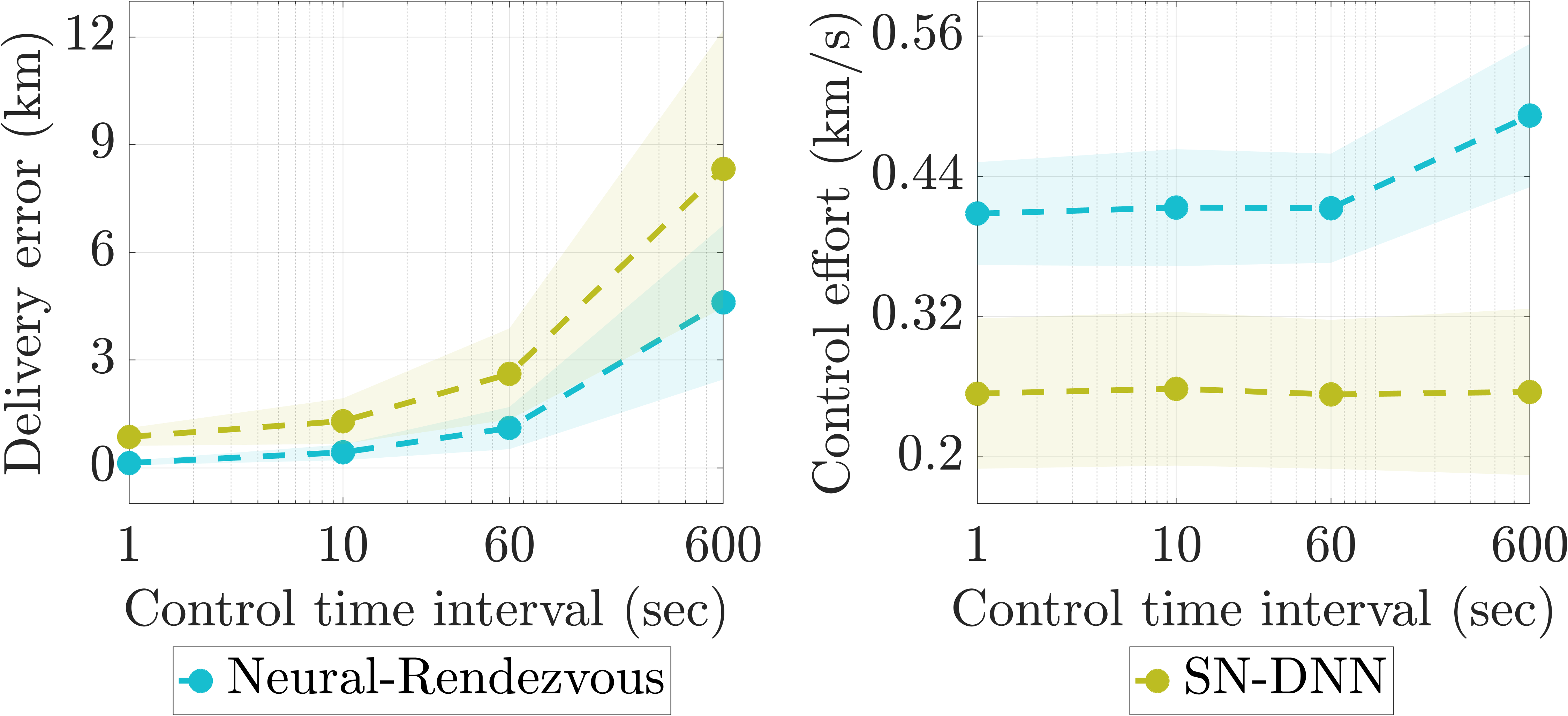}
    \caption{Control performances versus control time interval (averaged over 10 simulations for all ISOs in the test set). The shaded area denotes the standard deviations ($\bm{\pm1\sigma}$).}
    \label{fig_xerr_dt}
\end{figure}

Figure~\ref{fig_xerr_dt} then shows the spacecraft delivery error and control effort of Neural-Rendezvous of Algorithm~\ref{AlgISO} and SN-DNN terminal guidance of Sec.~\ref{sec_guidance}, averaged over $100$~ISOs in the test set, versus the control time interval. Although both of these methods involve discretization when implementing them in practice as pointed out in Remark~\ref{remark_discretization}, it can be seen that Neural-Rendezvous enables having the delivery error smaller than \SI{5}{\kilo\metre} with its standard deviation always smaller than that of the SN-DNN guidance, even for the control interval \SI{600}{\second}~(\SI{10}{\minute}). Since the MPC problem is solved by discretizing it with the time step $1$ s as explained in Sec.~\ref{sec_data_generation}, and the SN-DNN min-norm control~\eqref{min_norm_controller} of Theorem~\ref{Thm:control_robustness} is designed for continuous dynamics, Neural-Rendezvous yields less optimal control inputs for larger control time intervals, resulting in larger delivery error and control effort as expected from~\cite{mypaperTAC}.

\begin{figure}[htbp]
    \centering
    \includegraphics[width=83mm]{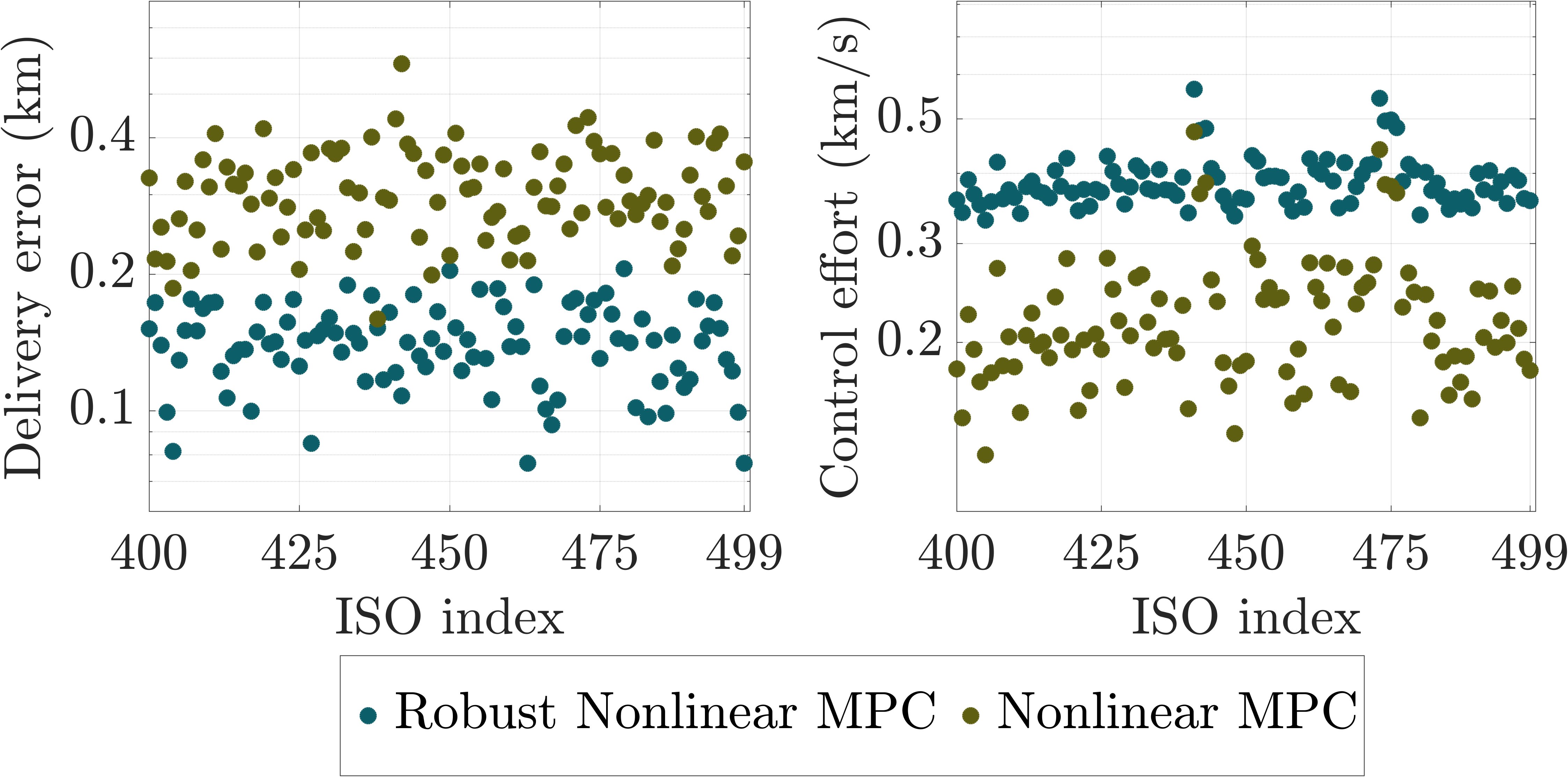}
    \caption{Control performances versus ISOs in the test set (averaged over 10 simulations performed for each ISO).}
    \label{fig_mpc_comp1}
\end{figure}

\begin{figure}[htbp]
    \centering
    \includegraphics[width=83mm]{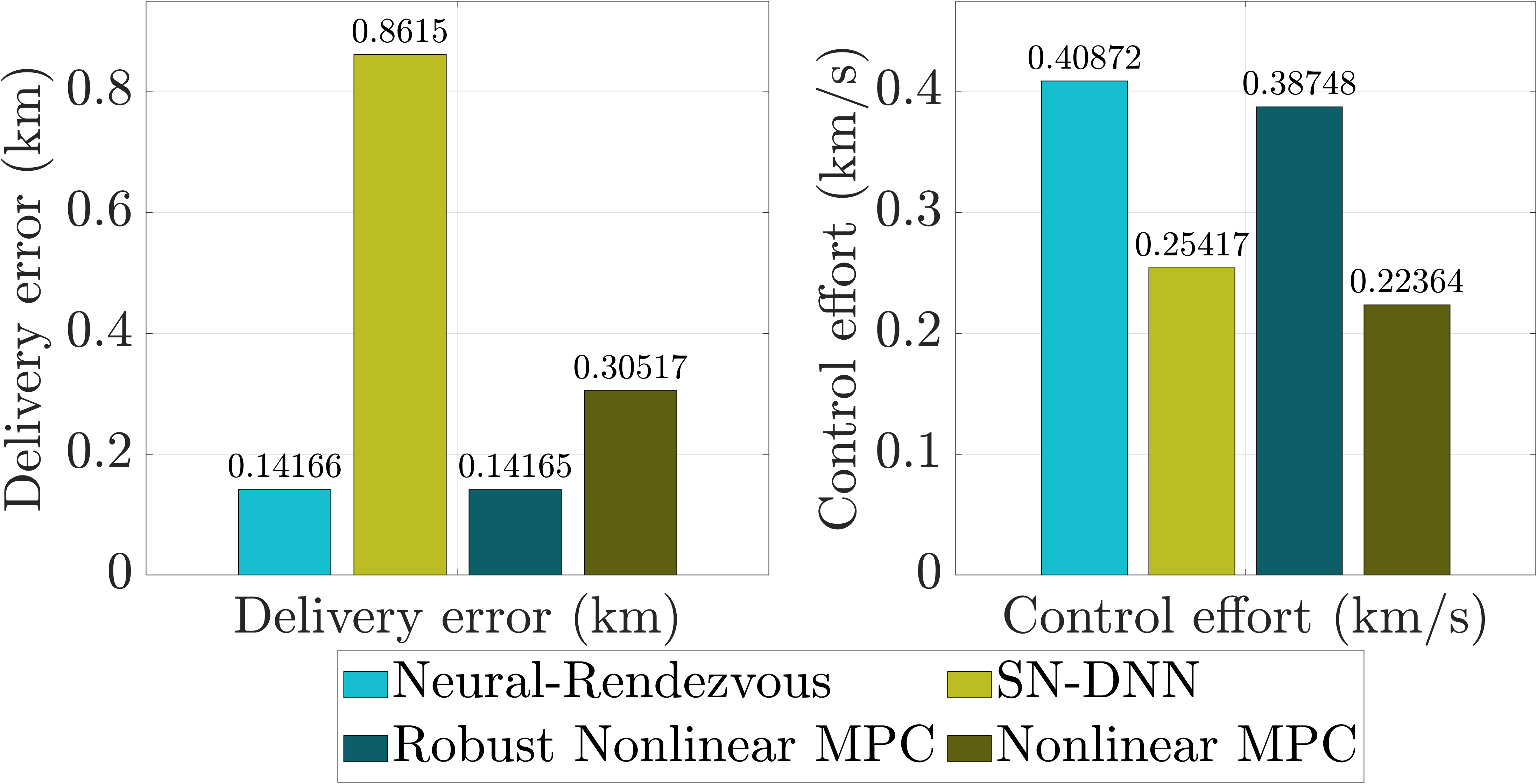}
    \caption{Performance comparison between our learning-based approaches and MPC. Each result shown above is the average of all ISOs in Fig.~\ref{fig_xerr_dv} and Fig.~\ref{fig_mpc_comp1}.}
    \label{fig_mpc_comp2}
\end{figure}

The performance of the nonlinear MPC (the optimization-based solution we aim to reproduce with machine learning) in this mission is shown in Figure~\ref{fig_mpc_comp1}, where the robust nonlinear MPC is the nonlinear MPC with the min-norm feedback control of Theorem~\ref{Thm:control_robustness}. The performance comparison between our learning-based control approaches and the MPC is summarized in Fig.~\ref{fig_mpc_comp2}. Comparing the SN-DNN guidance with the nonlinear MPC, we can see that that optimality gap is $1.37\times 10$\% on average for the control effort, while the SN-DNN delivery error is $1.82\times 10^2$\% larger than that of the MPC due to the lack of the delivery error guarantee, unlike Neural-Rendezvous. In contrast, the Neural-Rendezvous delivery error is only $5.84\times10^{-3}$\% larger than that of the robust nonlinear MPC at the slight expense of the control effort $5.48$\% larger than that of the robust nonlinear MPC. This is thanks to the formal probabilistic bound obtained in Theorem~\ref{Thm:control_robustness} equipped on top of the SN-DNN guidance, as expected.

\begin{figure}[htbp]
    \centering
    \href{https://youtu.be/AhDPE-R5GZ4?t=210}{\includegraphics[width=83mm]{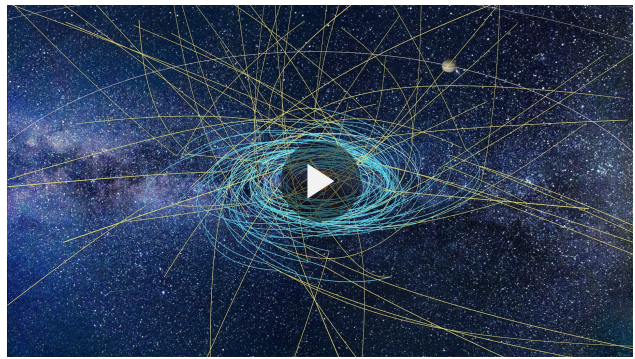}}
    \caption{Visualized Neural-Rendezvous trajectories for ISO exploration, where yellow curves represent ISO trajectories and blue curves represent spacecraft trajectories. More details can be found \href{hhttps://youtu.be/AhDPE-R5GZ4?t=210}{{\color{caltechgreen}here}}.}
    \label{fig_youtube}
\end{figure}

Finally, as shown in Table~\ref{tab_computational_time}, we can see that all the methods presented in this section including our proposed approaches, except for the nonlinear MPC or nonlinear robust MPC (global solution), can be computed in~$\leq 1$~(\si{\second}) and thus can be implemented in real time. The observations so far imply that the proposed approach indeed provides one of the promising solutions to Problem~\ref{Problem1} of Sec.~\ref{sec_problem}. The simulation results summarized in this section are visualized at {\color{caltechgreen}\href{https://youtu.be/AhDPE-R5GZ4}{https://youtu.be/AhDPE-R5GZ4}} as illustrated in Fig.~\ref{fig_youtube}.
\begin{table}[htbp]
\caption{Computational time of each method for the ISO encounter averaged over 100 evaluations, where the global solution is computed by the nonlinear MPC/nonlinear robust MPC.\label{tab_computational_time}}
\vspace{-1em}
\footnotesize
\begin{center}
\renewcommand{\arraystretch}{1.4}
\begin{tabular}{ l c }
\hline
\hline
 &Average computational time for one step~(\si{\second}) \\
\hline
Neural-Rendezvous & $8.0\times 10^{-4}$ \\
SN-DNN & $5.7\times 10^{-5}$ \\
Nonlinear robust control & $1.2\times 10^{-4}$ \\
PD control & $4.3\times 10^{-5}$ \\
Linear approx. MPC & $1.5\times 10^{-4}$ \\
\hline
Global Solution & $2.0\times 10^{3}$ \\
\hline
\hline
\end{tabular}
\end{center}
\vspace{-1em}
\end{table}
\subsection{Numerical Simulations at NASA JPL}
\label{sec_autonav}
The comparison of the performance of Neural-Rendezvous with the JPL state-of-practice autonomous navigation systems for small bodies, including ISOs, is discussed in~\cite{our_iso_mission}. Neural-Rendezvous is demonstrated to outperform them under realistic ISO state uncertainty in terms of the spacecraft delivery error, under mild GNC assumptions. Note that the uncertainty history is derived from an autonomous optical navigation orbit determination filter, as is used in the AutoNav system~\cite{autonav}, where a Monte-Carlo analysis simulating AutoNav performance is run to provide state estimation error versus time results.
\section{Empirical Validation}
\label{sec_experiment}
Applying deep learning-based algorithms to real-world systems always involves unmodeled uncertainties not just in its state, but also in the dynamics, environment, and control actuation, which could all lead to destabilizing behaviors different from what is learned and observed in numerical simulations. In fact, using the incremental stability-based analysis in the proof of Theorem~\ref{Thm:control_robustness}, we can further show that Neural-Rendezvous guarantees robustness against bounded and stochastic external disturbances and uncertainties in the dynamics and control actuation, including the SN-DNN learning error, even in a partially unknown environment~\cite{tutorial}. Such a strong mathematical guarantee helps us reproduce the simulation results seamlessly in real-world hardware experiments.
\subsection{Spacecraft Simulators}
\label{sec_scsimlator}
\begin{figure}[htbp]
    \centering
    \includegraphics[width=83mm]{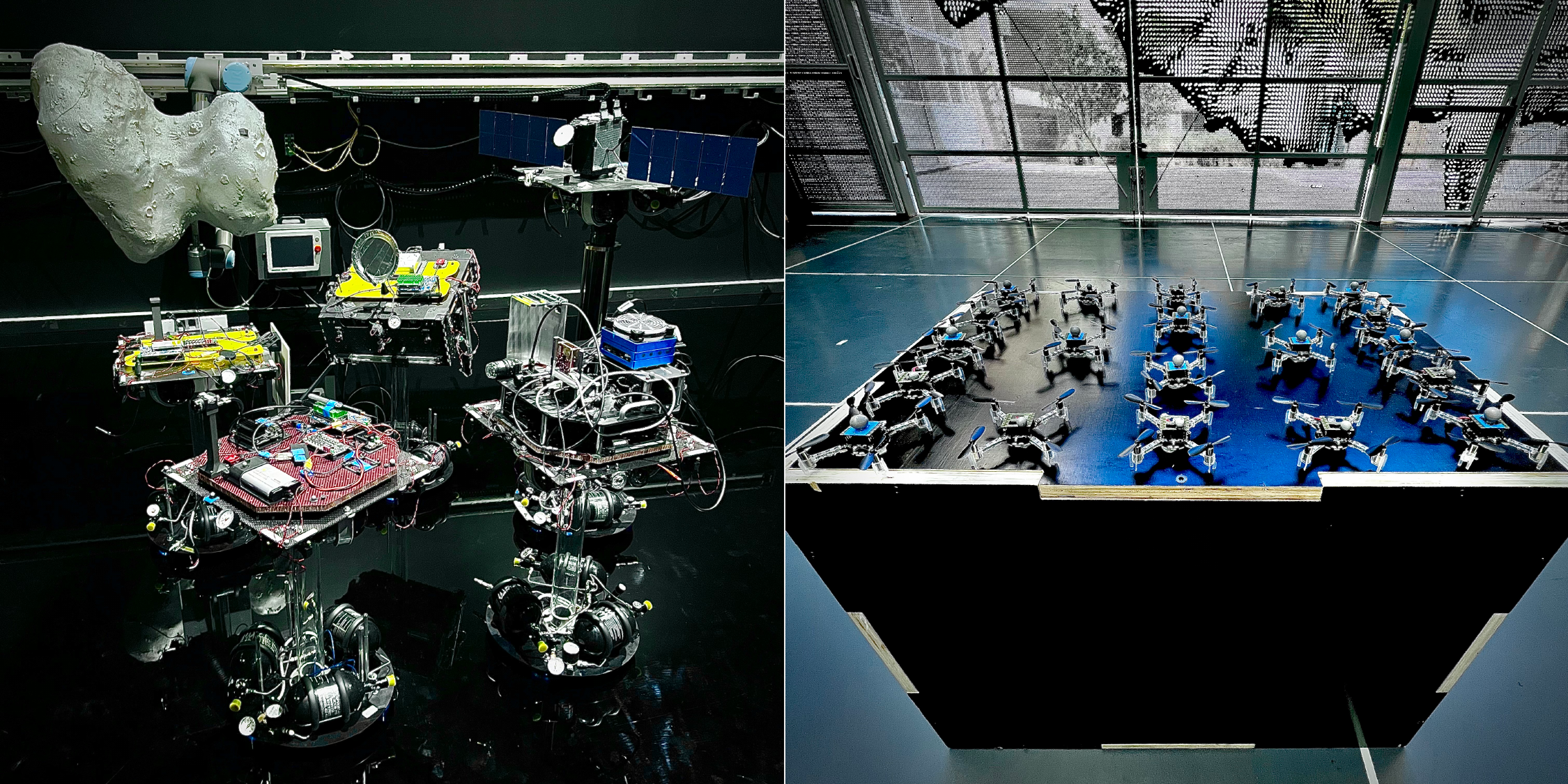}
    \caption{Left: Multi-Spacecraft Testbed for Autonomy Research (M-STAR). Right: Crazyflie, a versatile open-source flying development platform that only weighs 27~\si{\gram}.}
    \label{fig_mstar_cf}
\end{figure}

We first test the performance using our spacecraft simulator called M-STAR~\cite{SCsimulator} and epoxy flat floors for spacecraft motion simulation with the Vicon motion capture system ({\color{caltechgreen}\href{https://www.vicon.com/}{https://www.vicon.com/}}). We have $14$ motion capture cameras on the ceiling of this facility, with an IMU mounted on each spacecraft simulator for estimating the pose. The flatness of the epoxy floor is maintained within 0.001 inches for frictionless translation of the spacecraft dynamics using $3$ flat air-bearing pads, so we can properly demonstrate its motion in deep space. Each simulator has $8$ thrusters in addition to the $3$ air-bearing pads mounted at the bottom, which are to be used for controlling its position and attitude as in actual spacecraft. The left-hand side of Fig.~\ref{fig_mstar_cf} shows the spacecraft simulators and the ISO model used for this empirical validation.
\subsubsection{Relating M-STAR Dynamics to Spacecraft Dynamics Relative to ISO}
\label{sec_scale_down}
Although the M-STAR actuation works as in spacecraft in deep space due to the epoxy flat floor and thruster-based control, its dynamics given in~\cite{SCsimulator} is different from the one we consider in this paper~\eqref{original_dyn}. Also, due to the spatial limitation of the facility, we need to scale down the position, velocity, and control input of~\eqref{original_dyn} to the ones of M-STAR (see Table~\ref{tab_scales}).
\begin{table}[htbp]
\caption{Scales of the state and control in each dynamics.\label{tab_scales}}
\vspace{-1em}
\footnotesize
\begin{center}
\renewcommand{\arraystretch}{1.4}
\begin{tabular}{ l c c }
\hline
\hline
 & S/C w.r.t. to ISO & M-STAR \\
 \hline
Position~(\si{\kilo\metre}) & $\mathcal{O}(10^6)$ & $\mathcal{O}(10^{-3})$ \\
Velocity~(\si{\kilo\metre\per\second}) & $\mathcal{O}(10^1)$ & $\mathcal{O}(10^{-4})$ \\
Control~(\si{\kilo\metre\per\second\squared}) & $\mathcal{O}(10^{-5})$ & $\mathcal{O}(10^{-5})$ \\
\hline
\hline
\end{tabular}
\end{center}
\vspace{-1em}
\end{table}

To this end, we consider the following state $x_{\rm sim}$ and time $t_{\rm sim}$:
\begin{align}
    x_{\rm sim} = \begin{bsmallmatrix}
        p_{\rm sim} \\ v_{\rm sim}
    \end{bsmallmatrix} = 
    \begin{bsmallmatrix}
    \frac{p(t)-(p(0)+\dot{p}(0)t)}{c_p} \\ \frac{\dot{p}(t)-\dot{p}(0)}{c_v}
    \end{bsmallmatrix} ,~t_{\rm sim} = \tfrac{t}{c_t}
\end{align}
where $t$ is the actual time of~\eqref{original_dyn}, $p$ is the position of the spacecraft relative to the ISO as in~\eqref{actualfnB}, and $c_p$, $c_v$, and $c_t$ are constant scaling parameters. Taking the time derivative of $x_{\rm sim}$ with respect to $t_{\rm sim}$, we get
\begin{align}
    \frac{dx_{\rm sim}}{dt_{\rm sim}} = c_t
    \begin{bsmallmatrix}
    \dot{p}_{\rm sim} \\
    \dot{v}_{\rm sim}]
    \end{bsmallmatrix} =
    \begin{bsmallmatrix}
    \frac{c_vc_t}{c_p}v_{\rm sim} \\ \frac{c_t(-C(\textit{\oe})\dot{p}-G(p,\textit{\oe})+u)}{c_vm_{\mathrm{sc}}(t)}
    \end{bsmallmatrix}
\end{align}
where $\textit{\oe}$ is the ISO state, $u$ is the spacecraft control input, and $C$ and $G$ are the matrix functions defined in~\eqref{original_dyn} and~\eqref{actualfnB}. To be dynamically consistent, we select $c_p$, $c_v$, and $c_t$ to satisfy ${c_vc_t}/{c_p} = 1$. If we allocate the control input of the $8$ thrusters of M-STAR to have its $3$-dimensional acceleration vector $u_{\rm sim}$ as follows:
\begin{align}
    u_{\rm sim} = \frac{c_t}{c_vm_{\mathrm{sc}}(t)}(-C(\textit{\oe})\dot{p}-G(p,\textit{\oe})+u),
\end{align}
then we can convert the control input $u$ obtained by Neural-Rendezvous to the spacecraft simulator control input $u_{\rm sim}$, to be applied to the scaled-down M-STAR dynamics. Note that these procedures rewrite the dynamics as the double integrator dynamics, which can be easily demonstrated using the generalized pseudo-inverse control allocation scheme proposed in~\cite{SCsimulator}.
\subsubsection{Experimental Setup and Results}
\begin{figure}[htbp]
    \centering
    \href{https://drive.google.com/file/d/1CPVuKlRGM6u0tBCVWnwEZhH_JCgWhib0/view}{\includegraphics[width=83mm]{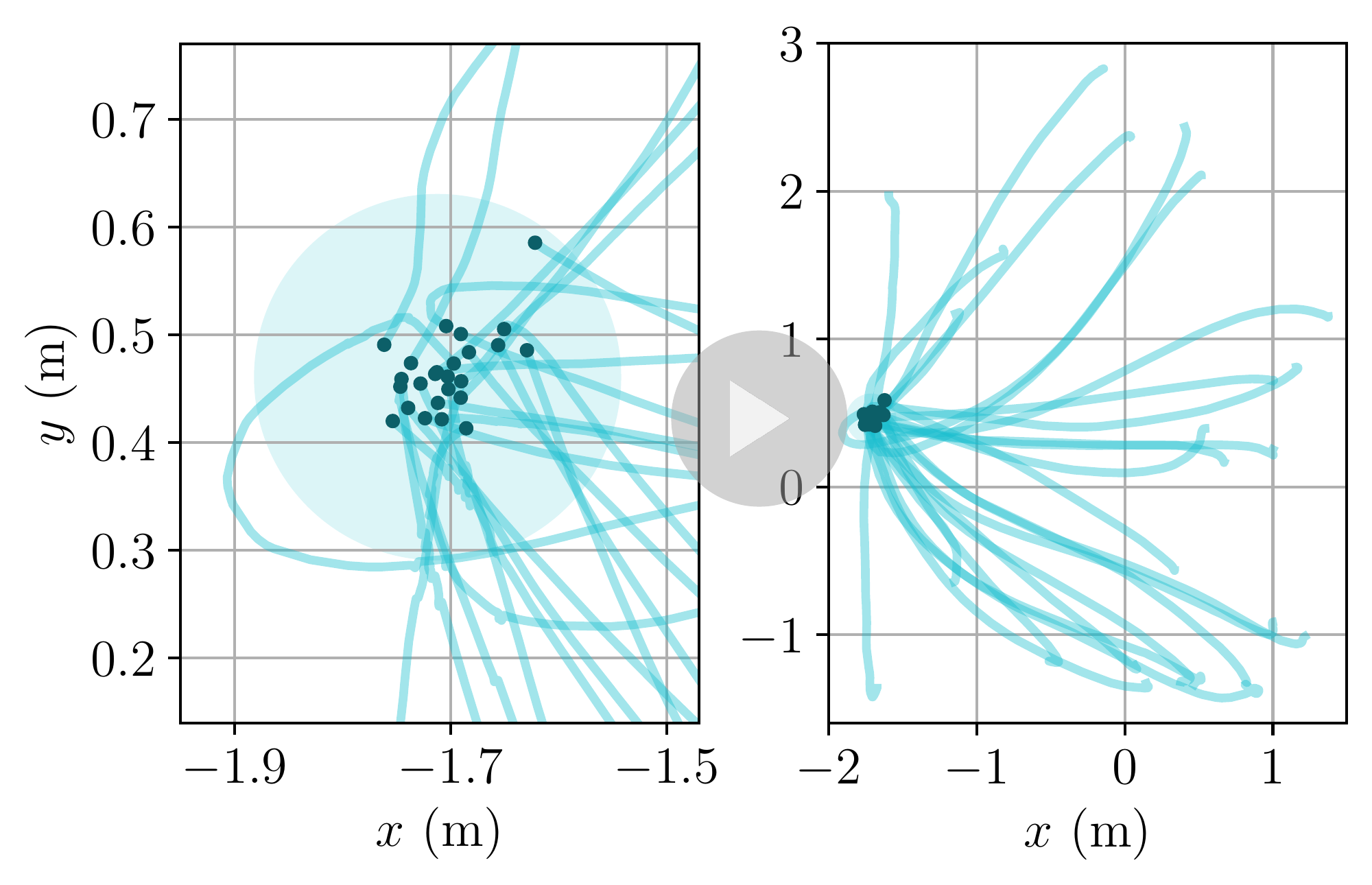}}
    \caption{Demonstrated Neural-Rendezvous trajectories of M-STAR, where the shaded blue circle indicates the expected delivery error bound (the left figure is a magnified view).}
    \label{fig_mstar_trj}
\end{figure}
We select NVIDIA Jetson TX2 as the main onboard computer to run the GNC and perception algorithms, where the software architecture is built on the Robotic Operating System (ROS) framework. The $8$ thrusters are controlled at $2$~\si{\hertz} control frequency (control input every $0.5$ seconds) with the motion capture camera system running at $100$~\si{\hertz}, which encourages the use of computationally efficient, deep learning-based control schemes in place of optimization-based control schemes. The state uncertainty in the M-STAR dynamics is added externally assuming the same uncertainty as in Sec.~\ref{sec_simulation}, scaled down by the method detailed in Sec.~\ref{sec_scale_down}. We select $25$ spacecraft relative state trajectories out of the ones given in Sec.~\ref{sec_simulation}, which meet the spatial constraint of our spacecraft simulator facility ($6.27$~\si{\metre} in $x$ direction and $8.05$~\si{\metre} in $y$ direction, see Table~\ref{tab_scales}) after the scaling-down procedures in Sec.~\ref{sec_scale_down}. Note that our spacecraft simulator facility demonstrates $2$-dimensional motions although the spacecraft in the actual mission moves in $3$-dimensional space.

\begin{figure}[htbp]
    \centering
    \href{https://youtu.be/D0HxsNh_rHM}{\includegraphics[width=83mm]{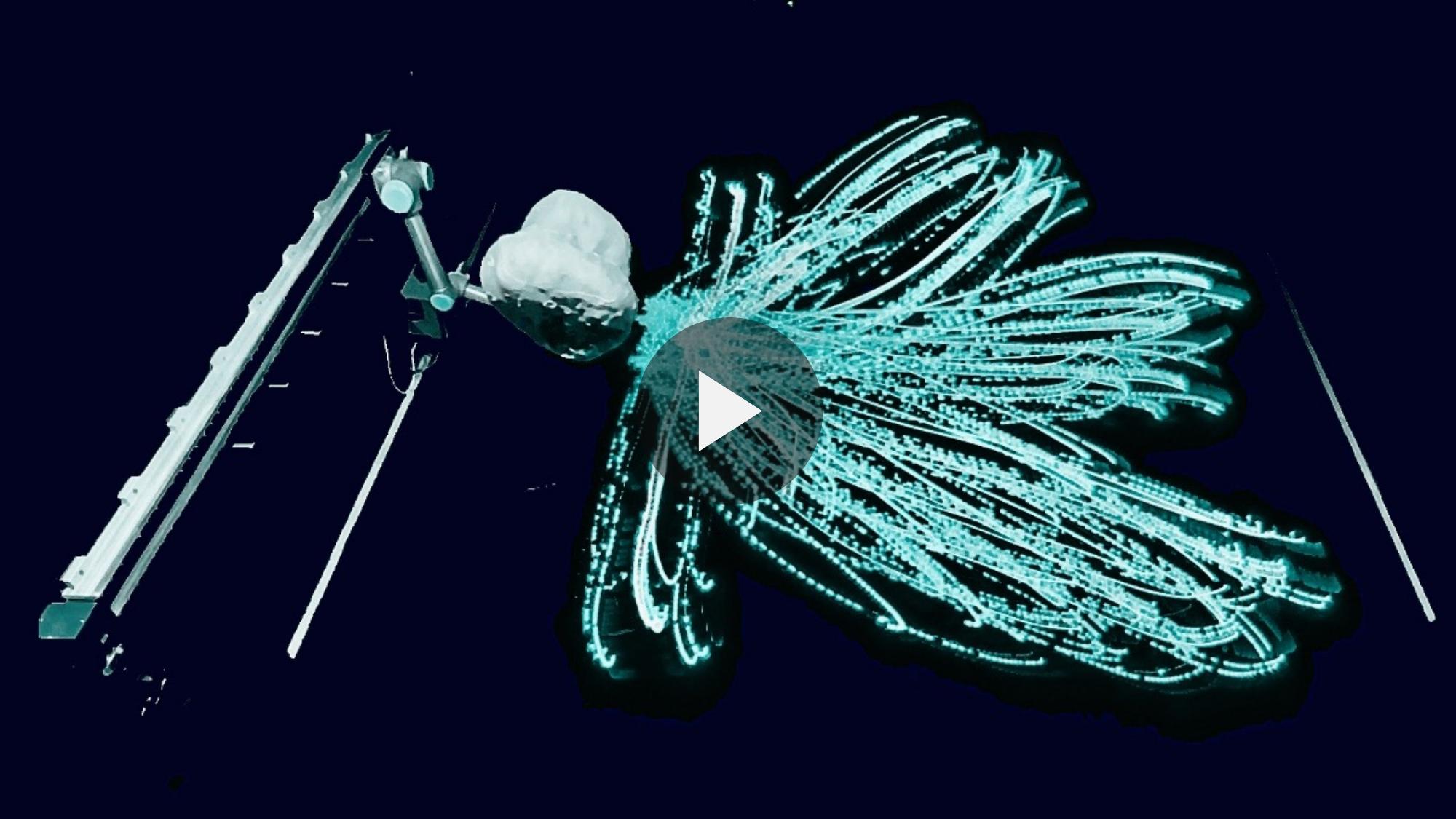}}
    \caption{Demonstrated Neural-Rendezvous trajectories of M-STAR observed in the actual environment, illustrated using their light trails. This movie can be found \href{https://youtu.be/D0HxsNh_rHM}{{\color{caltechgreen}here}}.}
    \label{fig_mstar_trj_real}
\end{figure}

The demonstrated trajectories of the M-STAR are shown in Fig.~\ref{fig_mstar_trj} and the trajectories observed in the actual environment are shown in Fig.~\ref{fig_mstar_trj_real}. The scaled-down trajectories of Fig.~\ref{fig_mstar_trj} and Fig.~\ref{fig_mstar_trj_real} correspond to the ones of the ISO candidates 400, 405, 406, 407, 408, 422, 425, 426, 430, 436, 437, 438, 439, 460, 461, 462, 463, 479, 480, 481, 490, 491, 492, 493, 494, simulated in Sec.~\ref{sec_simulation}. As can be seen from the figures and the \href{https://youtu.be/D0HxsNh_rHM}{{\color{caltechgreen}movie}}, Neural-Rendezvous indeed satisfies the probabilistic bound computed by Theorem~\ref{Thm:control_robustness}, which is indicated by the yellow ball in Fig.~\ref{fig_mstar_trj} and scaled down to the bound of the M-STAR dynamics using the method of Sec.~\ref{sec_scale_down}.
\subsection{High-Conflict and Distributed UAV Swarm Reconfiguration}
\label{sec_uav_swarm}
As we briefly mentioned in the introduction, Neural-Rendezvous and its guarantees are general enough to be used for other G\&C problems with completely different dynamics and objectives. To demonstrate this point, this section considers the problem of high-conflict reconfiguration of a UAV swarm, where the problem's complexity arises from the dynamics' nonlinearity, the aerodynamic nonlinear interaction between each UAV that acts as external disturbance, and highly nonlinear and distributed optimization required to efficiently avoid collisions considering the motions of multiple UAVs at each time instance, even with their limited communication radius. Another implicit focus of this validation is simply to demonstrate the performance of Neural-Rendezvous in the $3$-dimensional space in addition to the $2$-dimensional setup in Sec.~\ref{sec_scsimlator}. We use $20$ crazyflies shown on the right-hand side of Fig.~\ref{fig_mstar_cf}, which are designed to be a versatile open-source flying development platform to perform various types of aerial robotics research. The product description can be found here {\color{caltechgreen}\href{https://www.bitcraze.io/products/crazyflie-2-1/}{https://www.bitcraze.io/products/crazyflie-2-1/}}.
\subsubsection{Problem Statement}
Given multiple crazyflies, our objective is to intelligently and distributedly control each UAV to move to randomized and high-conflict target positions from randomized initial positions, optimally in a distributed manner. We solve nonlinear motion planning problems to sample target guidance and control inputs that minimize total control effort during the entire flight, using the sequential convex programming approach. The training is performed in a distributed manner as in~\cite{glas}, so we can use local observations to account for centralized global solutions even with the decentralized implementation of the G\&C algorithm. The distributed communication radius is set to $2$~\si{\metre} and the initial and target positions are randomly sampled from a $3$-dimensional cuboid ($x\in[-1.25,1.25]$, $y\in[0.60,1.90]$, and $z\in[0.7,2.7]$, all in meters) under the conditions that (i) the distances between the initial and target position of each UAV are at least $2.00$~\si{\metre} apart and (ii) the distances between each initial/target position and the other initial/target positions are at least $0.60$~\si{\metre} apart. The collision avoidance constraint is implemented in the optimization using a tall ellipsoid with the lengths of the principal axes being $0.6$~\si{\metre}, $0.6$~\si{\metre}, and $1.2$~\si{\metre}, in $x$, $y$, and $z$ directions, respectively, considering the downwash effect. The SN-DNN is trained as in Neural-Rendezvous with the robust learning approach proposed in~\cite{lagros}, which provides machine learning-based nonlinear motion planners with formal robustness and stability guarantees, even under the presence of learning errors and external disturbances.

\begin{figure*}[htbp]
    \centering
    \href{https://drive.google.com/file/d/1_s1ay7VNXJK3aimHfTeqArnuVtF7WVMG/view?usp=sharing}{\includegraphics[width=166mm]{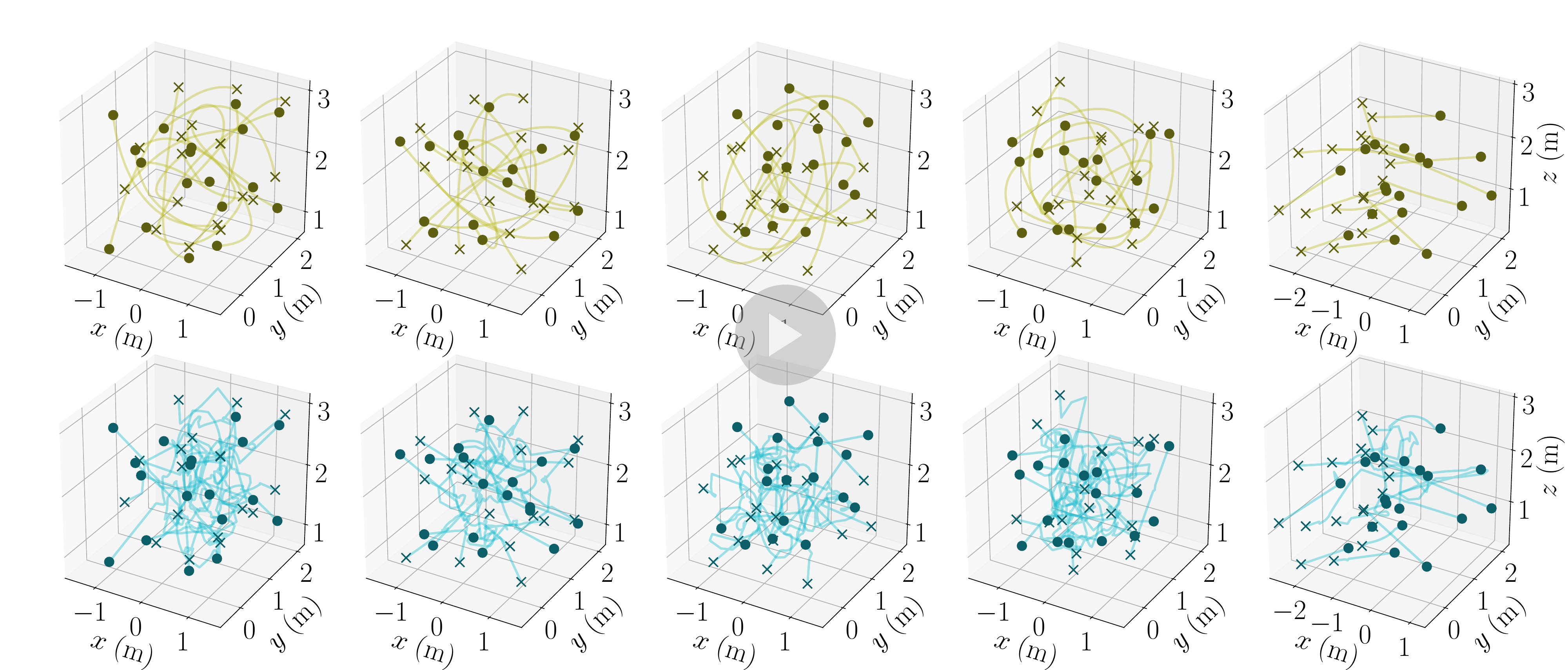}}
    \caption{First row: global solution trajectories of 18 crazyfies, found by solving the nonlinear motion planning. Second row: naive fictitious solution trajectories obtained with~\cite{7828016}.}
    \label{fig_cf_global}
\end{figure*}

Some example global solution trajectories are shown in the first row of Fig.~\ref{fig_cf_global}. They are obtained by solving the nonlinear motion planning problem by the sequential convex programming approach, which takes $4.4817\times10^3$~\si{\second} on average with the MacBook Pro laptop, \SI{2.2}{\giga\hertz}~Intel Core i7, \SI{16}{\giga\byte}~\SI{1600}{\mega\hertz} DDR3 RAM). The high-conflict nature of the UAV swarm reconfiguration is implied by the naive fictitious solution trajectories shown in the second row of Fig.~\ref{fig_cf_global}, which are computed just with the Buffered Voronoi Cells method~\cite{7828016} for collision avoidance, without solving the nonlinear motion planning.
\subsubsection{Experimental Setup and Results}
\begin{figure*}[ht]
    \centering
    \href{https://drive.google.com/file/d/1ZFxl6pF-kQC7jfzp7mV95ibaNhXxm2mz/view?usp=sharing}{\includegraphics[width=166mm]{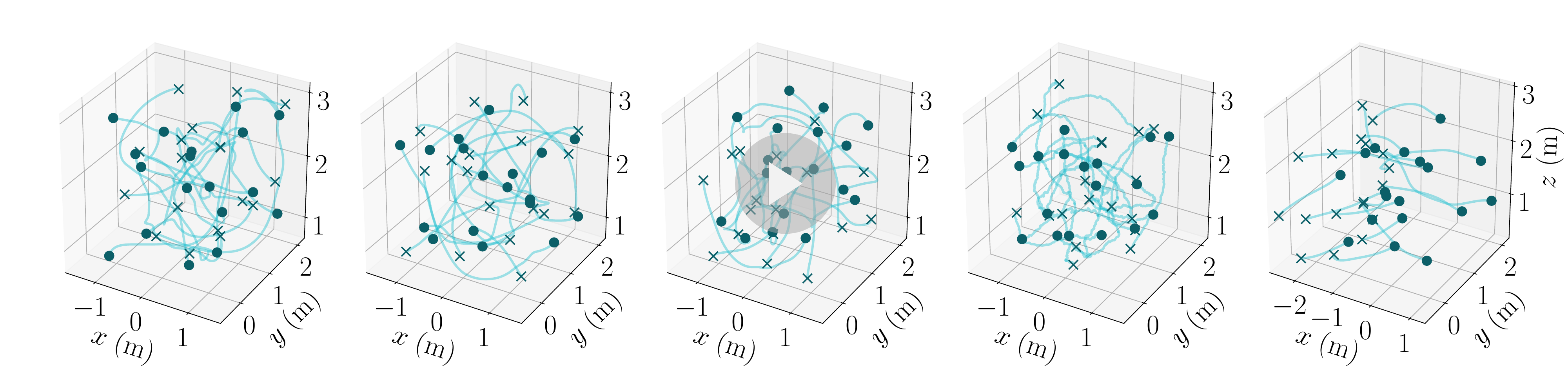}}
    \caption{Examples of demonstrated Neural-Rendezvous trajectories of 18 crazyfies in our experiment. The fourth figure represents those with artificial disturbance.}
    \label{fig_cf_trj}
\end{figure*}
The G\&C commands computed by Neural-Rendezvous are sent to crazyfies in real time over $8$ low-latency/long-range radio channels, with their control frequency being $100$~\si{\hertz}. The software architecture is constructed using {\color{caltechgreen}\href{https://crazyswarm.readthedocs.io/en/latest/}{crazyswarm}}, a ROS-based open-source platform for controlling UAV swarms~\cite{crazyswarm}. The pose of each UAV is estimated using the Vicon motion capture system, and we perform experiments both with and without artificial external and bounded disturbance acting on each of the UAV dynamics. When added, the disturbance $d$ with $\|d\|=0.125$ is used for demonstrating the robustness property of Neural-Rendezvous (see~\cite{lagros} for the definition of $d$). 

\begin{figure}[ht]
    \centering
    \href{https://youtu.be/u8ASO_r8rEA}{\includegraphics[width=83mm]{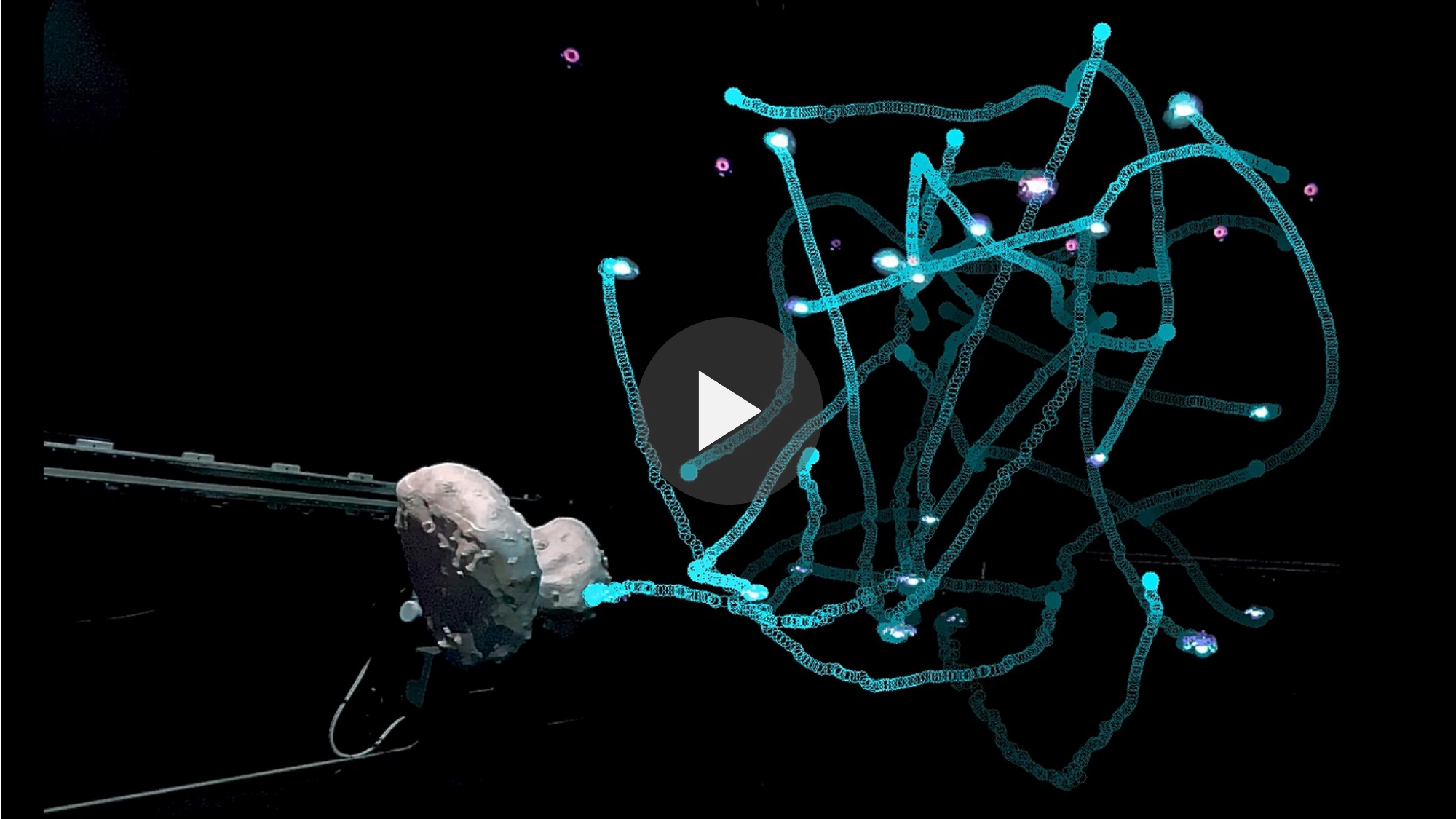}}
    \caption{Demonstrated Neural-Rendezvous trajectories of 20 crazyflies observed in the actual environment, illustrated using their light trails. This movie can be found \href{https://youtu.be/u8ASO_r8rEA}{{\color{caltechgreen}here}}.}
    \label{fig_cf_trj_real}
\end{figure}

The demonstrated trajectories of the crazyflies are shown in Fig.~\ref{fig_cf_trj} and the trajectories observed in the actual environment are shown in Fig.~\ref{fig_cf_trj_real}, where the first $3$ figures of Fig.~\ref{fig_cf_trj} represent the randomized experiments without artificial disturbance, the $4$th figure represents randomized simulation with artificial disturbance, and the $5$th figure represents the experiment with the ISO model depicted in Fig.~\ref{fig_cf_trj_real}. As implied in the figures and the \href{https://youtu.be/u8ASO_r8rEA}{{\color{caltechgreen}movie}}, Neural-Rendezvous successfully achieves the high-conflict UAV swarm reconfiguration in a distributed manner, robustly against the real-world disturbance solving the highly-nonlinear guidance and control problem. The total $2$-norm control effort for the reconfiguration is $1.7597$~\si{\metre\per\second} for Neural-Rendezvous and $1.2901$~\si{\metre\per\second} for the computationally-expensive global solution (which takes $4.4817\times10^3$~\si{\second} to be found on average when using the sequential convex programming approach on the MacBook Pro laptop, \SI{2.2}{\giga\hertz}~Intel Core i7, \SI{16}{\giga\byte}~\SI{1600}{\mega\hertz} DDR3 RAM). The optimality gap arises from the SN-DNN learning error, real-world disturbances, and distributed communication of the UAVs. The performance is demonstrated up to $20$ UAVs as is also visualized in this \href{https://youtu.be/u8ASO_r8rEA}{{\color{caltechgreen}movie}}.
\section{Conclusion}
\label{sec_conclusion}
This paper presents Neural-Rendezvous -- a deep learning-based terminal G\&C framework for achieving ISO encounter under large state uncertainty and high-velocity challenges, where we derive minimum-norm tracking control with an optimal MPC-based guidance policy imitated by the SN-DNN. As derived in Theorem~\ref{Thm:control_robustness} and illustrated in Fig.~\ref{fig_node_analogy}~--~\ref{fig_nr_diagram}, its major advantage is the formal optimality, stability, and robustness guarantee even with the use of machine learning, resulting in a spacecraft delivery error bound that decreases exponentially in expectation with a finite probability. The performance of Neural-Rendezvous is validated both in numerical simulations and hardware experiments, which also implies that it works not just for ISO exploration but for general nonlinear G\&C problems, solving them robustly against real-world disturbances and uncertainties. Having a verifiable performance guarantee is essential for using learning-based control in safety-critical robotic and aerospace missions, and our work provides a nonlinear control theoretical approach to formally meet this requirement.
\section*{Appendix}
\begin{proof}[Proof of Lemma~\ref{lemma_supermartingale} in Theorem~\ref{Thm:control_robustness}]
Let us define a non-negative and continuous function $E(x,t)$ as
\begin{align}
    \label{def_E}
    E(x,t) = v(x,t)+\delta_p\|p-p_d(t)\|
\end{align}
where $x=[p^{\top},\dot{p}^{\top}]^{\top}$ and $\delta_p\in\mathbb{R}_{> 0}$. Note that $E(x,t)$ of~\eqref{def_E} is $0$ only when $x=x_d(t)$. Since we have 
$\mathscr{A}\|p-p_d(t)\|\leq v(x,t)/\sqrt{m_{\mathrm{sc}}(t_f)}-\underline{\lambda}\|p-p_d(t)\|$, designing $\delta_p$ to have $\alpha-\delta_p/\sqrt{m_{\mathrm{sc}}(t_f)}>0$, the relation~\eqref{AV_appendix} gives
\begin{align}
    \label{def_AE}
    \mathscr{A}E(x_s,t_s) &\leq -\bar{\alpha}E(x_s,t_s)+\tfrac{L_k\sqrt{\|\hat{\textit{\oe}}_s-\textit{\oe}_s\|^2+\|\hat{x}_s-x_s\|^2}}{\sqrt{m_{\mathrm{sc}}(t_f)}}
\end{align}
for any $x_s,\hat{x}_s\in\mathcal{C}_{\mathrm{sc}}(r_{\mathrm{sc}})$, $\textit{\oe}_s,\hat{\textit{\oe}}_s\in\mathcal{C}_{\mathrm{iso}}(r_{\mathrm{iso}})$, and $t_s\in[0,t_f]$, where $\bar{\alpha} = \min\left\{\alpha-\delta_p/\sqrt{m_{\mathrm{sc}}(t_f)},\underline{\lambda}\right\} > 0$. Let us define another non-negative and continuous function $W(x,t)$ as
\begin{align}
    W(x,t) =E(x,t)e^{\gamma H(t)}+\frac{e^{\gamma H(t_f)}-e^{\gamma H(t)}}{\gamma}
\end{align}
where $\gamma$ is a positive constant that satisfies $2\bar{\alpha}\geq \gamma\sup_{t\in[0,t_f]}h(t)$. If $\bar{v},\bar{w}\in\mathbb{R}_{> 0}$ is selected to have $W(x,t)<\bar{w}\Rightarrow E(x,t)<\bar{v}\Rightarrow x\in\mathcal{C}_{\mathrm{sc}}(\bar{r}_{\mathrm{sc}})=\bigcup_{t\in[0,t_f]}\{s\in\mathbb{R}^n|\|s-x_d(t)\| < \bar{r}_{\mathrm{sc}}\} \subset \mathbb{R}^n$, which is always possible since $E(x,t)$ of~\eqref{def_E} is $0$ only when $x = x_d(t)$, then we can utilize~\eqref{def_AE} to compute $\mathscr{A}W$ as follows for any $x_s=[p_s^{\top},\dot{p}_s^{\top}]^{\top}\in\mathcal{C}_{\mathrm{sc}}(\bar{r}_{\mathrm{sc}})$ and $t_s\in[0,t_f]$ that satisfies $W(x_s,t_s)<\bar{w}$:
\begin{align}
    &\mathscr{A}W(x_s,t_s) = (\gamma h(t_s)E(x_s,t_s)+\mathscr{A}E(x_s,t_s))e^{\gamma H(t_s)} \\
    &-h(t_s)e^{\gamma H(t_s)} \leq \left(\mathscr{h}(\textit{\oe}_s,\hat{\textit{\oe}}_s,x_s,\hat{x}_s)-h(t_s)\right)e^{\gamma H(t_s)}
    \label{AW}
\end{align}
where $\mathscr{h}(\textit{\oe}_s,\hat{\textit{\oe}}_s,x_s,\hat{x}_s) = L_k\sqrt{(\|\hat{\textit{\oe}}_s-\textit{\oe}_s\|^2+\|\hat{x}_s-x_s\|^2)/m_{\mathrm{sc}}(t_f)}$ and the inequality follows from the relation $\bar{\alpha}\geq \gamma\sup_{t\in[0,t_f]}h(t)$. Applying Dynkin's formula~\cite[p. 10]{sto_stability_book} to~\eqref{AW}, it can be verified that $W(x(t),t)$ is a non-negative supermartingale due to the fact that $\mathbb{E}_{Z_s}\left[\mathscr{h}(\textit{\oe}(t),\hat{\textit{\oe}}(t),x(t),\hat{x}(t))\right] \leq L_k\varsigma^t(Z_s,t_s)/\sqrt{m_{\mathrm{sc}}(t_f)} = h(t)$ by definition of $\varsigma^t(Z_s,t_s)$ in~\eqref{navigation_exp_bound} of Assumption~\ref{assumption_navigation}, which gives the following due to Ville's maximal inequality~\cite[p. 26]{sto_stability_book}:
\begin{align}
    \label{Wsupermartingale}
    \mathbb{P}_{Z_s}\left[\sup_{t\in[t_s,t_f]}W(x(t),t) \geq \bar{w}\right] \leq \frac{E_s+\gamma^{-1}(e^{\gamma H(t_f)}-1)}{\bar{w}}
\end{align}
as long as we have $\textit{\oe}(t),\hat{\textit{\oe}}(t)\in\mathcal{C}_{\mathrm{iso}}(r_{\mathrm{iso}})$ and $\hat{x}(t)\in\mathcal{C}_{\mathrm{sc}}(r_{\mathrm{sc}})$ for $\forall t\in[t_s,t_f]$, where $Z_s = (\textit{\oe}_s,\hat{\textit{\oe}}_s,x_s,\hat{x}_s)$ is as defined in~\eqref{X_supermartingale}. For $Z_s$ satisfying $\textit{\oe}_s \in \mathcal{C}_{\mathrm{iso}}(\bar{r}_{\mathrm{iso}})$, $\hat{\textit{\oe}}_s \in \mathcal{C}_{\mathrm{iso}}(\bar{R}_{\mathrm{iso}})$, $x_s \in \mathcal{C}_{\mathrm{sc}}(\bar{r}_{\mathrm{sc}})$, and $\hat{x}_s \in \mathcal{C}_{\mathrm{sc}}(\bar{R}_{\mathrm{sc}})$, the occurrence of the events of Assumption~\ref{assumption_navigation} and the forward invariance condition of~\eqref{assump_deterministic_isoflow} of Assumption~\ref{assumption_control1} ensure that if we have $x(t)\in\mathcal{C}_{\mathrm{sc}}(\bar{r}_{\mathrm{sc}})$ for $\forall t\in[t_s,t_f]$, we get $\textit{\oe}(t)\in\mathcal{C}_{\mathrm{iso}}(\bar{r}_{\mathrm{iso}})$, $\hat{\textit{\oe}}(t)\in\mathcal{C}_{\mathrm{iso}}(r_{\mathrm{iso}})$, and $\hat{x}(t)\in\mathcal{C}_{\mathrm{sc}}(r_{\mathrm{sc}})$ for $\forall t\in[t_s,t_f]$. Therefore, selecting the constants $\gamma$ and $\bar{w}$ in the inequality~\eqref{Wsupermartingale} as in~\cite[pp. 82-83]{sto_stability_book} yields~\eqref{X_supermartingale} due to the relation $W(x,t)<\bar{w}\Rightarrow E(x,t)<\bar{v}\Rightarrow x\in\mathcal{C}_{\mathrm{iso}}(\bar{r}_{\mathrm{iso}})$.
\end{proof}
\section*{Funding Sources}
This work is funded by the Jet Propulsion Laboratory, California Institute of Technology.
\section*{Acknowledgments}
Part of the project was carried out at the Jet Propulsion Laboratory, California Institute of Technology, under a contract with the National Aeronautics and Space Administration.

We thank Stefano Campagnola (NASA JPL) for providing his useful simulation codes utilized in Sec.~\ref{sec_simulation}, thank Jimmy Ragan (Caltech) for his great help in setting up the spacecraft simulator experiments, thank Benjamin P. Riviere (Caltech) for sharing his valuable software resources in setting up the crazyswarm experiment, and thank Julie Castillo-Rogez (NASA JPL), Fred Y. Hadaegh (NASA JPL), Jean-Jacques E. Slotine (MIT), Karen Meech (University of Hawaii), and Robert Jedicke (University of Hawaii) for their insightful inputs and technical discussions.
\bibliography{main_jgcd}
\end{document}